\documentclass[twoside,12pt]{article}
\catcode`\@=11
\dimen\footins=24cm

\reversemarginpar
\ifcase\@ptsize
 \font\tenmsx=msam10
 \font\sevenmsx=msam7
 \font\fivemsx=msam5
 \font\tenmsy=msbm10
 \font\sevenmsy=msbm7
 \font\fivemsy=msbm5
\or
 \font\tenmsx=msam10 scaled \magstephalf
 \font\sevenmsx=msam8
 \font\fivemsx=msam6
 \font\tenmsy=msbm10 scaled \magstephalf
 \font\sevenmsy=msbm8
 \font\fivemsy=msbm6
\or
 \font\tenmsx=msam10 scaled \magstep1
 \font\sevenmsx=msam8
 \font\fivemsx=msam6
 \font\tenmsy=msbm10 scaled \magstep1
 \font\sevenmsy=msbm8
 \font\fivemsy=msbm6
\fi
 
\newfam\msxfam
\newfam\msyfam
\textfont\msxfam=\tenmsx  \scriptfont\msxfam=\sevenmsx
  \scriptscriptfont\msxfam=\fivemsx
\textfont\msyfam=\tenmsy  \scriptfont\msyfam=\sevenmsy
  \scriptscriptfont\msyfam=\fivemsy

\def\hexnumber@#1{\ifnum#1<10 \number#1\else
 \ifnum#1=10 A\else\ifnum#1=11 B\else\ifnum#1=12 C\else
 \ifnum#1=13 D\else\ifnum#1=14 E\else\ifnum#1=15 F\fi\fi\fi\fi\fi\fi\fi}

\def\msx@{\hexnumber@\msxfam}
\def\msy@{\hexnumber@\msyfam}
\mathchardef\boxdot="2\msx@00
\mathchardef\boxplus="2\msx@01
\mathchardef\boxtimes="2\msx@02
\mathchardef\square="0\msx@03
\mathchardef\blacksquare="0\msx@04
\mathchardef\centerdot="2\msx@05
\mathchardef\lozenge="0\msx@06
\mathchardef\blacklozenge="0\msx@07
\mathchardef\circlearrowright="3\msx@08
\mathchardef\circlearrowleft="3\msx@09
\mathchardef\rightleftharpoons="3\msx@0A
\mathchardef\leftrightharpoons="3\msx@0B
\mathchardef\boxminus="2\msx@0C
\mathchardef\Vdash="3\msx@0D
\mathchardef\Vvdash="3\msx@0E
\mathchardef\vDash="3\msx@0F
\mathchardef\twoheadrightarrow="3\msx@10
\mathchardef\twoheadleftarrow="3\msx@11
\mathchardef\leftleftarrows="3\msx@12
\mathchardef\rightrightarrows="3\msx@13
\mathchardef\upuparrows="3\msx@14
\mathchardef\downdownarrows="3\msx@15
\mathchardef\upharpoonright="3\msx@16

\mathchardef\downharpoonright="3\msx@17
\mathchardef\upharpoonleft="3\msx@18
\mathchardef\downharpoonleft="3\msx@19
\mathchardef\rightarrowtail="3\msx@1A
\mathchardef\leftarrowtail="3\msx@1B
\mathchardef\leftrightarrows="3\msx@1C
\mathchardef\rightleftarrows="3\msx@1D
\mathchardef\Lsh="3\msx@1E
\mathchardef\Rsh="3\msx@1F
\mathchardef\rightsquigarrow="3\msx@20
\mathchardef\leftrightsquigarrow="3\msx@21
\mathchardef\looparrowleft="3\msx@22
\mathchardef\looparrowright="3\msx@23
\mathchardef\circeq="3\msx@24
\mathchardef\succsim="3\msx@25
\mathchardef\gtrsim="3\msx@26
\mathchardef\gtrapprox="3\msx@27
\mathchardef\multimap="3\msx@28
\mathchardef\therefore="3\msx@29
\mathchardef\because="3\msx@2A
\mathchardef\doteqdot="3\msx@2B

\mathchardef\triangleq="3\msx@2C
\mathchardef\precsim="3\msx@2D
\mathchardef\lesssim="3\msx@2E
\mathchardef\lessapprox="3\msx@2F
\mathchardef\eqslantless="3\msx@30
\mathchardef\eqslantgtr="3\msx@31
\mathchardef\curlyeqprec="3\msx@32
\mathchardef\curlyeqsucc="3\msx@33
\mathchardef\preccurlyeq="3\msx@34
\mathchardef\leqq="3\msx@35
\mathchardef\leqslant="3\msx@36
\mathchardef\lessgtr="3\msx@37
\mathchardef\backprime="0\msx@38
\mathchardef\risingdotseq="3\msx@3A
\mathchardef\fallingdotseq="3\msx@3B
\mathchardef\succcurlyeq="3\msx@3C
\mathchardef\geqq="3\msx@3D
\mathchardef\geqslant="3\msx@3E
\mathchardef\gtrless="3\msx@3F
\mathchardef\sqsubset="3\msx@40
\mathchardef\sqsupset="3\msx@41
\mathchardef\trianglerighteq="3\msx@44
\mathchardef\trianglelefteq="3\msx@45
\mathchardef\bigstar="0\msx@46
\mathchardef\between="3\msx@47
\mathchardef\blacktriangledown="0\msx@48
\mathchardef\blacktriangleright="3\msx@49
\mathchardef\blacktriangleleft="3\msx@4A
\mathchardef\blacktriangle="0\msx@4E
\mathchardef\triangledown="0\msx@4F
\mathchardef\eqcirc="3\msx@50
\mathchardef\lesseqgtr="3\msx@51
\mathchardef\gtreqless="3\msx@52
\mathchardef\lesseqqgtr="3\msx@53
\mathchardef\gtreqqless="3\msx@54
\mathchardef\Rrightarrow="3\msx@56
\mathchardef\Lleftarrow="3\msx@57
\mathchardef\veebar="2\msx@59
\mathchardef\barwedge="2\msx@5A
\mathchardef\doublebarwedge="2\msx@5B
\mathchardef\angle="0\msx@5C
\mathchardef\measuredangle="0\msx@5D
\mathchardef\sphericalangle="0\msx@5E
\mathchardef\varpropto="3\msx@5F
\mathchardef\smallsmile="3\msx@60
\mathchardef\smallfrown="3\msx@61
\mathchardef\Subset="3\msx@62
\mathchardef\Supset="3\msx@63
\mathchardef\Cup="2\msx@64

\mathchardef\Cap="2\msx@65

\mathchardef\curlywedge="2\msx@66
\mathchardef\curlyvee="2\msx@67
\mathchardef\leftthreetimes="2\msx@68
\mathchardef\rightthreetimes="2\msx@69
\mathchardef\subseteqq="3\msx@6A
\mathchardef\supseteqq="3\msx@6B
\mathchardef\bumpeq="3\msx@6C
\mathchardef\Bumpeq="3\msx@6D
\mathchardef\lll="3\msx@6E

\mathchardef\ggg="3\msx@6F

\mathchardef\circledS="0\msx@73
\mathchardef\pitchfork="3\msx@74
\mathchardef\dotplus="2\msx@75
\mathchardef\backsim="3\msx@76
\mathchardef\backsimeq="3\msx@77
\mathchardef\complement="0\msx@7B
\mathchardef\intercal="2\msx@7C
\mathchardef\circledcirc="2\msx@7D
\mathchardef\circledast="2\msx@7E
\mathchardef\circleddash="2\msx@7F
\def\ulcorner{\delimiter"4\msx@70\msx@70 }
\def\urcorner{\delimiter"5\msx@71\msx@71 }
\def\llcorner{\delimiter"4\msx@78\msx@78 }
\def\lrcorner{\delimiter"5\msx@79\msx@79 }
\def\yen{\mathhexbox\msx@55 }
\def\checkmark{\mathhexbox\msx@58 }
\def\circledR{\mathhexbox\msx@72 }
\def\maltese{\mathhexbox\msx@7A }
\mathchardef\lvertneqq="3\msy@00
\mathchardef\gvertneqq="3\msy@01
\mathchardef\nleq="3\msy@02
\mathchardef\ngeq="3\msy@03
\mathchardef\nless="3\msy@04
\mathchardef\ngtr="3\msy@05
\mathchardef\nprec="3\msy@06
\mathchardef\nsucc="3\msy@07
\mathchardef\lneqq="3\msy@08
\mathchardef\gneqq="3\msy@09
\mathchardef\nleqslant="3\msy@0A
\mathchardef\ngeqslant="3\msy@0B
\mathchardef\lneq="3\msy@0C
\mathchardef\gneq="3\msy@0D
\mathchardef\npreceq="3\msy@0E
\mathchardef\nsucceq="3\msy@0F
\mathchardef\precnsim="3\msy@10
\mathchardef\succnsim="3\msy@11
\mathchardef\lnsim="3\msy@12
\mathchardef\gnsim="3\msy@13
\mathchardef\nleqq="3\msy@14
\mathchardef\ngeqq="3\msy@15
\mathchardef\precneqq="3\msy@16
\mathchardef\succneqq="3\msy@17
\mathchardef\precnapprox="3\msy@18
\mathchardef\succnapprox="3\msy@19
\mathchardef\lnapprox="3\msy@1A
\mathchardef\gnapprox="3\msy@1B
\mathchardef\nsim="3\msy@1C
\mathchardef\napprox="3\msy@1D
\mathchardef\nsubseteqq="3\msy@22
\mathchardef\nsupseteqq="3\msy@23
\mathchardef\subsetneqq="3\msy@24
\mathchardef\supsetneqq="3\msy@25
\mathchardef\subsetneq="3\msy@28
\mathchardef\supsetneq="3\msy@29
\mathchardef\nsubseteq="3\msy@2A
\mathchardef\nsupseteq="3\msy@2B
\mathchardef\nparallel="3\msy@2C
\mathchardef\nmid="3\msy@2D
\mathchardef\nshortmid="3\msy@2E
\mathchardef\nshortparallel="3\msy@2F
\mathchardef\nvdash="3\msy@30
\mathchardef\nVdash="3\msy@31
\mathchardef\nvDash="3\msy@32
\mathchardef\nVDash="3\msy@33
\mathchardef\ntrianglerighteq="3\msy@34
\mathchardef\ntrianglelefteq="3\msy@35
\mathchardef\ntriangleleft="3\msy@36
\mathchardef\ntriangleright="3\msy@37
\mathchardef\nleftarrow="3\msy@38
\mathchardef\nrightarrow="3\msy@39
\mathchardef\nLeftarrow="3\msy@3A
\mathchardef\nRightarrow="3\msy@3B
\mathchardef\nLeftrightarrow="3\msy@3C
\mathchardef\nleftrightarrow="3\msy@3D
\mathchardef\divideontimes="2\msy@3E
\mathchardef\varnothing="0\msy@3F
\mathchardef\nexists="0\msy@40
\mathchardef\mho="0\msy@66
\mathchardef\thorn="0\msy@67
\mathchardef\beth="0\msy@69
\mathchardef\gimel="0\msy@6A
\mathchardef\daleth="0\msy@6B
\mathchardef\lessdot="3\msy@6C
\mathchardef\gtrdot="3\msy@6D
\mathchardef\ltimes="2\msy@6E
\mathchardef\rtimes="2\msy@6F
\mathchardef\shortmid="3\msy@70
\mathchardef\shortparallel="3\msy@71
\mathchardef\smallsetminus="2\msy@72
\mathchardef\thicksim="3\msy@73
\mathchardef\thickapprox="3\msy@74
\mathchardef\approxeq="3\msy@75
\mathchardef\succapprox="3\msy@76
\mathchardef\precapprox="3\msy@77
\mathchardef\curvearrowleft="3\msy@78
\mathchardef\curvearrowright="3\msy@79
\mathchardef\digamma="0\msy@7A
\mathchardef\varkappa="0\msy@7B
\mathchardef\hslash="0\msy@7D
\mathchardef\hbar="0\msy@7E
\mathchardef\backepsilon="3\msy@7F
\def\Bbb{\ifmmode\let\next\Bbb@\else
 \def\next{\errmessage{Use \string\Bbb\space only in math mode}}\fi\next}
\def\Bbb@#1{{\Bbb@@{#1}}}
\def\Bbb@@#1{\fam\msyfam#1}

\catcode`\@=12
\usepackage{epic}
\usepackage{german}
\usepackage{times}
\newcommand\SEKIedition{%
Searchable Online Edition\\}
\long\def\math#1{\relax\ifmmode{#1}\else$#1$\fi}
\newcommand\yestop   {\vspace{\topsep}}

\newcommand{\ASFm}{ASF$^+$}

\newcommand{\tsym}[1]{\mbox{``{\tt #1}''}}

{\begin{array}[t]{@{}l@{}}}
{\end{array}}

\newenvironment{einruecken}[1]%
{\begin{list}{}{\leftmargin#1 \topsep1ex \parskip0cm} \item[]}%
{\end{list}}

\newenvironment{block}%
{\begin{einruecken}{3ex}}%
{\end{einruecken}}

\newsavebox{\zurueck}

\newenvironment{algorithmus}[5]%
{\yestop\vspace{3ex} \parskip0ex \parsep1ex \fboxsep0.2cm \sloppy \noindent
   {#1}: $\begin{array}[t]{l}
            #2 
         \\ \longrightarrow
            #3 
         \end{array}$
   \\[2ex]
   \centerline{\fbox{\parbox{14.8cm}{#4}}} \\[1ex]
   \sbox{\zurueck}{\sf #5}
   \begin{block} \sf}%
{\vspace{1ex}\end{block} \usebox{\zurueck} \vspace{3ex}\yestop}

\newcommand{\placepic}[1]{\vspace{2ex} \centerline{#1}}
\newcommand{\ind}{\mbox{\hspace*{3em}}}

\newcommand{\Bergstra}{[Bergstra\&al.89]}
\newcommand{\Eschbach}{[Eschbach94]}
\newcommand{\Lunde}{[Wirth\&Lunde94]}
\newcommand{\WirthA}{[Wirth\&Gramlich93]}
\newcommand{\WirthB}{[Wirth\&Gramlich94]}

\newcommand{\sign}{\mbox{\it sign}}
\newcommand{\modn}{\mbox{\it modn}}

\newcommand{\private}{{\tt private}}
\newcommand{\public}{{\tt public}}

\newcommand{\cunion}{\sqcup}
\newcommand{\Union}{\displaystyle\bigcup}
\newcommand{\Cunion}{\displaystyle\bigsqcup}
\newcommand{\nform}{{\it nform}}
\newcommand{\nforms}{{\it nforms}}
 
\newcommand{\ASFMODULE}{\mbox{\sf ASF-MODULE}}
\newcommand{\ASFSPEC}{\mbox{\sf ASF-SPEC}}
\newcommand{\PROVEDB}{\mbox{\sf PROVE-DB}}
\newcommand{\MODULENAME}{\mbox{\sf MODULE-NAME}}
\newcommand{\SHORTMODULENAME}{\mbox{\sf SHORT-MODULE-NAME}}
\newcommand{\INSTNAME}{\mbox{\sf INST-NAME}}
\newcommand{\USERNAME}{\mbox{\sf USER-NAME}}
\newcommand{\MODULEINSTNAME}{\mbox{\sf MODINST-NAME}}
\newcommand{\SHORTMODULEINSTNAME}{\mbox{\sf SHORT-MODINST-NAME}}
\newcommand{\SPECNAME}{\mbox{\sf SPEC-NAME}}
\newcommand{\SORTVECTOR}{\mbox{\sf SORT-VECTOR}}
\newcommand{\DISAMBSPECNAME}{\mbox{\sf DISAMB-SPEC-NAME}}
\newcommand{\VISIBILITYFUNC}{\mbox{\sf VISIBILITY-FUNC}}
\newcommand{\RENAMINGFUNC}{\mbox{\sf RENAMING-FUNC}}
\newcommand{\BINDINGBLOCK}{\mbox{\sf BINDING-BLOCK}}
\newcommand{\IMPORT}{\mbox{\sf IMPORT}}
\newcommand{\SIG}{\mbox{\sf SIG}}
\newcommand{\VARSORTFUNC}{\mbox{\sf VAR-SORT-FUNC}}
\newcommand{\CLAUSE}{\mbox{\sf CLAUSE}}
\newcommand{\EQUATION}{\mbox{\sf EQUATION}}
\newcommand{\MODULE}{\mbox{\sf MODULE}}
\newcommand{\ORIGIN}{\mbox{\sf ORIGIN}}
\newcommand{\ORIGINFUNC}{\mbox{\sf ORIGIN-FUNC}}
\newcommand{\DEPENDENCYFUNC}{\mbox{\sf DEPENDENCY-FUNC}}
\newcommand{\GFORM}{\mbox{\sf GF}}
\newcommand{\NFORM}{\mbox{\sf NF}}
\newcommand{\DISAMBRENAMINGFUNC}{\mbox{\sf DISAMB-RENAMING-FUNC}}
\newcommand{\lvs}{\tsym{label}, \tsym{variable}, \tsym{sort}}

\newcommand{\disambname}{\mbox{\it dis-name}}
\newcommand{\pot}{\mbox{$\cal P$}}
\newcommand{\dom}{\mbox{\sf Dom}}
\newcommand{\range}{\mbox{\sf Ran}}
\newcommand{\name}{{\it name\/}}
\newcommand{\uname}{{\it uname\/}}
\newcommand{\specname}{{\it specname\/}}

\newcommand{\sortvector}{{\it sortvector\/}}
\newcommand{\sortv}{{\it sortv\/}}
\newcommand{\sortn}{{\it sortn\/}}

\newcommand{\subimp}{_{\mbox{\tiny imp}}}

\newcommand{\subresult}{_{\mbox{\tiny result}}}
\newcommand{\subact}{_{\mbox{\tiny ACT}}}
\newcommand{\subactav}{_{\mbox{\tiny ACT-AV}}}
\newcommand{\subform}{_{\mbox{\tiny FORM}}}
\newcommand{\subconstform}{_{\mbox{\tiny const,FORM}}}
\newcommand{\subconstactav}{_{\mbox{\tiny const,ACT-AV}}}
\newcommand{\subnonconstform}{_{\mbox{\tiny non-const,FORM}}}
\newcommand{\subnonconstactav}{_{\mbox{\tiny non-const,ACT-AV}}}
\newcommand{\subnp}{_{\mbox{\scriptsize np}}}

\newcommand{\importinggf}{\mbox{\it importing-gf}}
\newcommand{\modname}{{\it modname\/}}
\newcommand{\modiname}{{\it modiname\/}}
\newcommand{\modinames}{{\it modinames\/}}
\newcommand{\defmodiname}{{\it modiname\/}}
\newcommand{\paradefmodiname}{{\it paradefmod\/}}
\newcommand{\paradefmodinst}{{\it paradefmodinst\/}}
\newcommand{\iname}{{\it iname\/}}

\newcommand{\oldinames}{{\it oldinames\/}}
\newcommand{\symboltype}{{\it symboltype\/}}
\newcommand{\symboltypes}{{\it symboltypes\/}}
\newcommand{\origin}{{\it origin\/}}
\newcommand{\originfunc}{{\it originfunc\/}}
\newcommand{\visibility}{{\it visibility\/}}
\newcommand{\sig}{{\it sig\/}}
\newcommand{\module}{{\it module\/}}
\newcommand{\mod}{{\it mod\/}}
\newcommand{\originf}{{\it originf\/}}
\newcommand{\depf}{{\it depf\/}}
\newcommand{\depfimpconst}{\mbox{\it depf-imp-const\/}}
\newcommand{\depfimpmod}{\mbox{\it depf-imp-mod\/}}
\newcommand{\depfactmod}{\mbox{\it depf-act-mod\/}}

\newcommand{\parameters}{{\it parameters\/}}
\newcommand{\provedb}{\mbox{\it prove-db\/}}
\newcommand{\renaming}{{\it renaming\/}}
\newcommand{\ren}{{\it ren\/}}

\newcommand{\binding}{{\it binding\/}}
\newcommand{\bindingblocks}{{\it bindingblocks\/}}
\newcommand{\visibilityfunc}{{\it visibilityf\/}}
\newcommand{\condition}{{\it condition\/}}
\newcommand{\conditions}{{\it conditions\/}}
\newcommand{\equations}{{\it equations\/}}
\newcommand{\toinstanciate}{{\it toinst\/}}

\newcommand{\varsortfunc}{{\it varsortfunc\/}}
\newcommand{\goal}{{\it goal\/}}
\newcommand{\goals}{{\it goals\/}}
\newcommand{\spec}{{\it spec\/}}

\newcommand{\import}{{\it import\/}}
\newcommand{\imports}{{\it imports\/}}
\newcommand{\const}{{\it const\/}}
\newcommand{\nonconst}{\mbox{\it non-const\/}}
\newcommand{\sort}{{\it sort}}
\newcommand{\sorts}{{\it sorts}}
\newcommand{\disfuncs}{{\it disfuncs}}
\newcommand{\cons}{\mbox{\it cons-decs}}
\newcommand{\ncons}{\mbox{\it ncons-decs}}
\newcommand{\sortpar}{{\it sortpar}}
\newcommand{\funcpar}{{\it funcpar}}
\newcommand{\subst}{\mbox{\it sub}}
\newcommand{\sortparrenaming}{\mbox{\it sortpar-renaming}}
\newcommand{\funcparrenaming}{\mbox{\it funcpar-renaming}}
\newcommand{\parrenaming}{\mbox{\it par-renaming}}
\newcommand{\normalforms}{{\it normalforms\/}}

\newcommand{\GetSpecName}{{\it GetSpecName\/}}
\newcommand{\GetParameterRenamings}{{\it GetParameterRenamings\/}}
\newcommand{\GetRenaming}{{\it GetRenaming\/}}
\newcommand{\MakeConsistent}{{\it MakeConsistent\/}}

\newcommand{\AdaptVisibility}{{\it AdaptVisibility\/}}
\newcommand{\CombineDependencies}{{\it CombineDependencies\/}}
\newcommand{\CombineImports}{{\it CombineImports\/}}
\newcommand{\CombineWithImports}{{\it Combine\-With\-Imports\/}}
\newcommand{\CombineWithActModule}{{\it CombineWithActModule\/}}

\newcommand{\Hide}{{\it Hide\/}}
\newcommand{\InstanciateModInstName}{{\it InstanciateModInstName\/}}
\newcommand{\Instanciate}{{\it Instanciate\/}}
\newcommand{\Rename}{{\it Rename\/}}
\newcommand{\modinst}{{\it modinst}}
\newcommand{\modinstances}{{\it modinstances\/}}

\newcommand{\CheckSemanticConditions}{{\it CheckSemanticConditions\/}}
\newcommand{\SeperateParaBlock}{{\it SeparateParaBlock\/}}

\newcommand{\Bind}{{\it Bind\/}}

\newcommand{\ModuleText}{{\it ModuleText\/}}
\newcommand{\MakeGF}{{\it MakeGF\/}}
\newcommand{\NForm}{{\it NF\/}}

\newcommand{\ExternModRep}{{\it ExternModRep\/}}
\newcommand{\NormalForm}{{\it NormalForm\/}}

\newcommand{\Ur}{\mbox{\it Ur}}

\newcommand{\ts}[1]{``{\tt #1}''}
\newcommand{\nts}[1]{{\sf $<$#1$>$}}
\newcommand{\opt               }[1]{\mbox{[\ #1\ ]}}
\newcommand{\reps              }[1]{\mbox{#1$^{+}$}}
\newcommand{\rep               }[1]{\mbox{#1*}}
\newcommand{\repswt            }[2]{\mbox{(#1 \ts{#2})$^{+}$}}
\newcommand{\repwt             }[2]{\mbox{(#1 \ts{#2})*}}
\newcommand{\tpar              }[1]{\ts{(}#1\ts{)}}

\newcommand{\Gspezification}{\nts{specification}}
\newcommand{\Gmodule}{\nts{module}}
\newcommand{\Gparameterblock}{\nts{parameter-block}}
\newcommand{\Gimport}{\nts{import}}
\newcommand{\Gextendedparameterblock}{\nts{ext-para-block}}
\newcommand{\Gnamewithrenaming}{\nts{name-with-ren}}
\newcommand{\Gextfunctionname}{\nts{ext-func-name}}
\newcommand{\Gimportblock}{\nts{import-block}}
\newcommand{\Gaddsignature}{\nts{add-signature}}
\newcommand{\Gparameterblocksignature}{\nts{para-block-sig}}
\newcommand{\Gsignature}{\nts{signature}}
\newcommand{\Gfunctiondec}{\nts{function-dec}}
\newcommand{\Gvariables}{\nts{variables}}
\newcommand{\Gvariabledec}{\nts{variable-dec}}
\newcommand{\Gequations}{\nts{equations}}
\newcommand{\Gequation}{\nts{equation}}
\newcommand{\Geq}{\nts{eq}}
\newcommand{\Ggoals}{\nts{goals}}
\newcommand{\Gclause}{\nts{clause}}
\newcommand{\Gshortmodulename}{\nts{short-module-name}}
\newcommand{\Gmodulename}{\nts{module-name}}
\newcommand{\Gsortname}{\nts{sort-name}}
\newcommand{\Gfunctionname}{\nts{function-name}}
\newcommand{\Gsortorfunctionname}{\nts{sort-or-func-name}}
\newcommand{\Gvariablename}{\nts{variable-name}}
\newcommand{\Ginstancename}{\nts{instance-name}}
\newcommand{\Gterm}{\nts{term}}
\newcommand{\Gprimary}{\nts{primary}}
\newcommand{\Gmacroequation}{\nts{macro-equation}}
\newcommand{\Glabel}{\nts{label}}
\begin{document}
\pagestyle{empty}
\newlength{\disp}
\setlength{\disp}{\textwidth}
\addtolength{\disp}{-2.0\parindent}
\pagestyle{empty}
\setcounter{page}{1}
\flushbottom
\noindent
\begin{minipage}{\textwidth}
\mbox{}

\vspace{3.5em}
\begin{center}
{\LARGE\bf
ASF$^+$ \\--- 
eine
ASF-"ahnliche
\\
Spezifikationssprache
\\\mbox{}
}
{
\\\mbox{}
\\\mbox{}
\\\mbox{}
\large
R\"{u}diger Lunde,
Claus-Peter Wirth
\\\mbox{}
\\
}
{\SEKIedition
December 22, 1994
\\\mbox{}
\\\mbox{}
\\\mbox{}
SEKI-WORKING-PAPER SWP--94--05 (SFB)\\\mbox{}\\
Fachbereich Informatik,\\
 Universit\"at Kaiserslautern, \\
 D--67663 Kaiserslautern
\\
\mbox{}\\
\mbox{}\\
\mbox{}\\

}
\end{center}
\end{minipage}

\noindent
\begin{center}
\begin{minipage}{\disp}
\footnotesize\renewcommand{\baselinestretch}{0.8}{\bf Zusammenfassung:}
Ohne auf wesentliche Aspekte der in {[Bergstra\&al.89]} 
vorgestellten algebraischen Spezifikationssprache
ASF zu verzichten, haben wir ASF um 
die folgenden Konzepte erweitert:
W"ahrend in ASF einmal exportierte Namen bis zur Spitze der 
Modulhierarchie
sichtbar bleiben m"ussen,
erm"oglicht 
ASF$^+$ 
ein
differenziertes
Verdecken von Signaturnamen.
Das fehlerhafte Vermischen unterschiedlicher Strukturen,
welches in ASF beim Import verschiedener Aktualisierungen desselben
parametrisierten Moduls auftritt,
wird in ASF$^+$ durch eine ad"aquatere Form der 
Parameterbindung vermieden.
Das neue Namensraum-Konzept von ASF$^+$ 
erlaubt es dem Spezifizierer, einerseits die Herkunft verdeckter Namen
direkt zu identifizieren und anderseits beim Import eines Moduls 
auszudr"ucken, 
ob dieses Modul nur benutzt oder in seinen wesentlichen
Eigenschaften ver"andert werden soll.
Im ersten Fall kann er auf eine einzige global zur Verf"ugung stehende 
Version zugreifen; im zweiten Fall mu"s er eine Kopie des Moduls
importieren.
Schlie"slich
erlaubt 
ASF$^+$ 
semantische Bedingungen
an Parameter und die Angabe von Beweiszielen.

\end{minipage}
\end{center}

\vspace{\fill}

\noindent
\footnoterule
\noindent
{\footnotesize 
Diese Arbeit ist aus einer von
Klaus Madlener
und
Claus-Peter Wirth
betreuten 
Projektarbeit
R"udiger Lundes hervorgegangen und wurde 
gef"ordert
von der Deutschen
Forschungsgemeinschaft, SFB 314 (D4-Projekt).

}

\pagebreak

\mbox{}
\vfill
\vfill
\vfill
\vfill

\noindent
\mbox{}\\
\begin{center}
\begin{minipage}{\disp}
\footnotesize\renewcommand{\baselinestretch}{0.8}{\bf Abstract:}
Maintaining the main aspects of the
algebraic specification language ASF as presented in {[Bergstra\&al.89]}
we have extend ASF with the following concepts:
While once exported names in ASF must stay visible up to the top the
module hierarchy,
ASF$^+$ 
permits
a more sophisticated hiding of signature names.
The erroneous merging of distinct structures that 
occurs when importing different actualizations of the same 
parameterized module in ASF
is avoided in ASF$^+$ by a more adequate form of
parameter binding.
The new ``Namensraum''-concept of ASF$^+$ 
permits the specifier on the one hand directly to
identify the origin of hidden names
and on the other 
to decide 
whether an imported
module is only to be accessed or whether 
an important property of it is to be modified.
In the first case he can access one single globally 
provided version; in the second he has to import a copy of the module.
Finally 
ASF$^+$ 
permits
semantic conditions on parameters and the 
specification of 
tasks for a theorem prover.

\end{minipage}
\end{center}

\vfill
\vfill

\cleardoublepage

\tableofcontents
\contentsline{section}{Literatur}{\pageref{bibliography}}
\vspace{\fill}

\cleardoublepage

\pagenumbering{arabic}
\setcounter{page}{1}
\pagestyle{myheadings}%

\section{Einleitung}

Mit steigender Leistungsf"ahigkeit moderner automatischer Beweissysteme
w"achst auch die Komplexit"at der mit ihnen zu bearbeitenden
Problemstellungen. Auf der Suche nach Konzepten zur logisch strukturierten
Formulierung derartiger Probleme haben sich in der Entwicklung von 
Spezifikationssprachen Modularisierungsans"atze herausgebildet.
Eine Spezifikation besteht danach aus mehreren Modulen, die mit Hilfe von
Importbefehlen aufeinander Bezug nehmen.
Besonders in umfangreichen Spezifikationen erweisen sich modulare Repr"asentationen
von Spezifikationen als vorteilhaft. Die Verst"andlichkeit wird durch die
Zerlegung in einzelne, durch exakt definierte Schnittstellen
(Importkonstrukte) miteinander verbundene Teilspezifikationen gesteigert.
Au"serdem k"onnen h"aufig verwendete Strukturen (beispielsweise die
Datenstruktur Boolean) in Bibliotheken abgelegt werden, was den
Spezifikationsaufwand reduziert.

Verschiedene M"oglichkeiten, Module miteinander zu kombinieren, werden in
dieser Arbeit diskutiert.
 % Eine interessante Frage dabei ist, in wie weit
 % sich aus Beweisen gewonnenes Wissen "uber ein Modul auf eine Spezifikation
 % "ubertragen l"a"st, die das Modul enth"alt.
Das Hauptinteresse gilt der Entwicklung einer
Sprache f"ur modulare Spezifikationen mit positiv/negativ bedingten
Gleichungen. Ausgehend von der in \Bergstra\ vorgestellten Sprache ASF, die
bereits "uber ein recht differenziertes Modularisierungskonzept verf"ugt,
wird eine Erweiterung \ASFm\ vorgestellt, welche die 
im ersten und vorvorletzten 
Punkt von ``1.4.1.\ Known defects and limitations of 
ASF'' in \Bergstra\ genannten M"angel von ASF behebt. 
\ASFm\ unterst"utzt:
\begin{itemize}
\item Import und Parametrisierung von Modulen
\item "Uberladen von Funktionsnamen
\item Infix-Operatoren
\item differenziertes Verdecken von Funktions- und Sortennamen
\item positiv/negativ bedingte Gleichungen
\item rudiment"are Verwaltung von Beweiszielen
\end{itemize}

Als Semantik wird, analog zu [Bergstra \&~al. 89], semi-formal eine
Normalisierungsprozedur
angegeben, welche die Modulhierachie einer komplexen Spezifikation in eine
flache Spezifikation (ohne Importe) umwandelt. Von zentraler Bedeutung 
ist in diesem Zusammenhang die Originfunktion, die jedem in der
Spezifikation auftretenden Namen einen Informationsblock zuweist. Dieser
enth"alt f"ur den Normalisierungsproze"s wichtige Informationen "uber den
Kontext der Namensdefinition, beispielsweise den Namen des Definitionsmoduls.
Neben der Originfunktion verwaltet die Normalisierungsprozedur aus \ASFm\
eine Dependenzfunktion. Sie spielt bei expliziten Umbenennungen und
Parameterbindungen eine wichtige Rolle und tr"agt der hierarischen Struktur
der Spezifikation Rechnung.
Neu in \ASFm\ ist auch, da"s bei der Kombination von Modulen das Umbenennen von
verdeckten Namen nicht ausschlie"slich durch Konfliktfreiheit definiert wird.
Jeder verdeckte Name beinhaltet in \ASFm\ unter anderem das K"urzel des
Herkunftsmoduls, was zum einen Konfliktfreiheit garantiert, zum andern auch
modulare Information sichtbar macht und damit der "Ubersicht dient.

\pagebreak %--------------------------------------------------------------------

\section{Das Konzept, erkl"art anhand von Beispielspezifikationen}

Um mit der Syntax von \ASFm\ vertraut zu werden und ein erstes intuitives
Verst"andnis der neuen Sprache zu gewinnen, bietet es sich an, zun"achst
einige Beispielspezifikationen zu betrachten. Die hier angegebenen
Module {\tt Booleans}, {\tt Naturals} und {\tt Sequences} entsprechen im
wesentlichen den gleichnamigen Modulen aus [Bergstra \&~al. 89], 
Kapitel 1.1.2., was einen direkten Vergleich erlaubt.

\subsection{Bottom-Up-Spezifikationen}

\begin{verbatim}
module Booleans
short  Bo
{      
   add signature
   {  public:
         sorts
            BOOL
         constructors
            true, false :             -> BOOL
         non-constructors
            and, or     : BOOL # BOOL -> BOOL

      private:
         non-constructors
            not         : BOOL        -> BOOL  }
   
   variables
   {  non-constructors
         x,y : -> BOOL  }
      
   equations
   {
      macro-equation and(x,y)
      {
         case
         {  ( x @ true ) : y
            ( x @ false ): false  }
      }
      
      macro-equation not(x)
      {
         case
         {  ( x @ true ) : false
            ( x @ false ): true  }
      }

      [e1] or(x, y) = not(and(not(x), not(y)))
   }
} /* Booleans */
\end{verbatim}

Jedes Modul einer Spezifikation beginnt mit dem Schl"usselwort {\tt module}, 
gefolgt vom Modulnamen, dem optionalen {\tt short}-Konstrukt und einem Block.
Das {\tt short}-Konstrukt stellt ein Modulnamenk"urzel zur Verf"ugung,
das beim Umbenennen verdeckter Namen Verwendung findet und zumindest bei langen
Modulnamen nicht fehlen sollte.
Fehlt die Angabe des Modulk"urzels, so wird der Modulname selbst ersatzweise
als sein eigenes K"urzel verwendet. Die K"urzel werden global zur Bezeichnung
der Module herangezogen und m"ussen daher innerhalb der Spezifikation
eindeutig sein.

Alle nicht importierten Teile der Signatur werden mit dem {\tt add signature}-Konstrukt
zur internen Signatur zusammengefa"st. Sie umfa"st einen nach au"sen
sichtbaren (\public) und einen nur innerhalb des Moduls
zug"anglichen (\private) Bereich,
in denen Sorten- und Funktionsnamen deklariert werden k"onnen. Da der
Spezifikationssemantik ein konstruktorbasierter Ansatz zu Grunde liegt
(vergleiche etwa \WirthA\ oder \WirthB), wird
zwischen {\tt constructors} und {\tt non-constructors} unterschieden.
Im Beispiel sind die Sorte
{\tt BOOL}, die Konstanten {\tt true}, {\tt false} und die 
(Pr"adikats-) Funktionen {\tt and}, {\tt or} nach au"sen sichtbar (k"onnen
also von anderen Modulen importiert werden). {\tt not} wird zu Illustrationszwecken
nicht exportiert, und kann  infolgedessen nur innerhalb des Moduls
referenziert werden.

Im Beispiel folgt eine Variablenvereinbarung, die jeder im Gleichungsblock
verwendeten Variable eine Sorte zuweist. Die Overloadingf"ahigkeit von \ASFm\
   (d.h. die M"og\-lich\-keit namensgleiche Funktionen mit verschiedenen
   Argumentsorten zu unterscheiden)
macht eine Deklaration aller Variablen zwingend notwendig. \ASFm\ unterscheidet zwischen
Konstruktor- und Non-Konstruktor-Variablen, die durch die Schl"usselw"orter
{\tt constructors} und {\tt non-con\-struc\-tors} gekennzeichnet werden. Defaultwert ist
{\tt constructors}. Werden nur Konstruktor-Variablen verwendet, so kann deshalb
das Schl"usselwort (wie in den folgenden Beispielen) entfallen. 

\ASFm\ unterst"utzt Spezifikationen mit positiv/negativ bedingten Gleichungen.
Sie k"onnen im Gleichungsblock entweder explizit angegeben werden (im Beispiel
die Zeile mit Marke {\tt e1}) oder mit Hilfe des {\tt macro-equation}-Konstrukts
erzeugt werden.
Das {\tt macro-equation}-Konstrukt geht aus dem {\tt macro-rule-Konstrukt} aus \Lunde\
hervor und unterscheidet sich nur durch die C-"ahnliche Syntax. Seine Semantik
ist durch Makro-Expansion in positiv/negativ bedingte Gleichungen gegeben.
Eine wichtige Rolle spielen sogenannte match-conditions (Symbolisiert durch
{\tt @}), mit deren Hilfe Gleichungen, deren linke Seiten mit dem
gleichen Funktionssymbol beginnen, zusammengefa"st werden k"onnen.
Im Beispiel f"uhrt die Makro-Expansion zu den vier Gleichungen
\begin{verbatim}
[me-and1] and(true, y)  = y
[me-and2] and(false, y) = false
[me-not1] not(true)     = false
[me-not2] not(false)    = true
\end{verbatim}
Bei umfangreichen Funktionsdefinitionen bietet die Darstellung als {\tt
macro-equation}
gro"se Vorteile, weil durch verschachtelte {\tt case}-Konstrukte zahlreiche
Wiederholungen von Bedingungen eingespart werden k"onnen.
F"ur die genaue Bedeutung der Makros {\tt @}, {\tt case}, {\tt if} und
{\tt else} sei auf \Lunde\ verwiesen.

Alle verwendeten Variablen und Marken werden semantisch wie
{\tt private}-deklarierte Signaturnamen behandelt und m"ussen nur innerhalb
des Moduls eindeutig sein.

\begin{verbatim}
module Naturals
short  Nat
{
   import Booleans {  public: BOOL, true, false  }
    
   add signature
   {
      public:
         sorts
            NAT
         constructors
            0     :           -> NAT
            s     : NAT       -> NAT
         non-constructors
            _ + _ : NAT # NAT -> NAT
            eq    : NAT # NAT -> BOOL
   }
   variables
   {  x,y,u : -> NAT  }
      
   equations
   {
      macro-equation (x + y)
      {
         case
         {  ( y @ 0 )    : x
            ( y @ s(u) ) : s(x + u)  }
      }

      macro-equation eq(x,y)
      {  if ( x = y ) true
            else      false  }
   }
} /* Naturals */
\end{verbatim}

Das Modul {\tt Naturals} importiert das Modul {\tt Booleans}. Der Block,
der dem Importbefehl folgt, tr"agt der Forderung nach einem flexiblen
Lokalit"atsprinzip Rechnung. Er sorgt daf"ur, da"s nur die im Block aufgef"uhrten
Namen im Modul zug"anglich sind. Im Beispiel sind die Sorte {\tt BOOL} und
die Konstanten {\tt true} und {\tt false} innerhalb des Moduls
{\tt Naturals} sichtbar und werden auch von ihm exportiert.
Die von {\tt Booleans} exportierten, aber im Importkonstrukt nicht aufgef"uhrten
Funktionen {\tt and} und {\tt or} und die nicht exportierte Funktion {\tt not}
k"onnen innerhalb von {\tt Naturals} nicht referenziert werden.
Ihre Namen gelten als verdeckt (hidden).

Unter den im {\tt add signature}-Konstrukt deklarierten Funktionssymbolen
befindet sich auch der Infix-Operator ``{\tt +}''. Seine Deklarationssyntax wurde,
wie auch die der Pr"afix-Operatoren, aus ASF "ubernommen.

\begin{verbatim}
module OrdNaturals
short  ONat
{
   import Booleans
   {  public: BOOL, true; private: or }
   
   import Naturals
   {  public: NAT, 0, s, eq, false  }
    
   add signature
   {  public:
         non-constructors
            greater, geq: NAT # NAT -> BOOL  }
   
   variables
   {  x,y,u,v : -> NAT  }
      
   equations
   {
      macro-equation greater(x,y)
      {
         case
         {  ( x @ 0 )              : false
            ( x @ s(u), y @ 0 )    : true
            ( x @ s(u), y @ s(v) ) : greater(u,v)  }
      }
      
      [e1] geq(x,y) = or(greater(x,y), eq(x,y))
   }
   
   goals
   {  [irref] greater(x, x)
         -->
      [trans] greater(x, u), greater(u, y)
         -->  greater(x, y)
      [total] 
         -->  greater(x, y), greater(y, x), x = y  }
      
} /* OrdNaturals */
\end{verbatim}

\vfill\pagebreak

{\tt OrdNaturals} spezifiziert eine irreflexive Ordnung {\tt greater} und
eine reflexive Ordnung {\tt geq} f"ur Elemente des Typs {\tt NAT}. Der doppelte
Import des Moduls {\tt Booleans} (direkt und indirekt "uber {\tt Naturals})
demonstriert, da"s die Sichtbarkeit von Namen eines importierten Moduls
im allgemeinen nicht von einem Importbefehl allein abh"angt. So w"are es
falsch, aus dem Fehlen des Namens {\tt false} im ersten Importblock
abzuleiten, da"s {\tt false} innerhalb von {\tt OrdNaturals} verdeckt sein mu"s.

Der {\tt goals}-Block am Ende von {\tt OrdNaturals} erm"oglicht es dem
Spezifizierer, Beweisziele anzugeben. Jede Beweisaufgabe besteht aus einer
in eckigen Klammern eingefa"sten Marke, gefolgt von einer Gentzenklausel.
Syntaktisch handelt es sich dabei um eine Folge von durch Kommas getrennte
Gleichungen, gefolgt von einem Pfeil und einer weiteren Folge von Gleichungen.
Semantisch ist die Gentzenformel
$e_1, \ldots, e_n$ {\tt -->} $e_{n+1}, \ldots ,e_{n+m}$ equivalent zu
$e_1 \wedge \ldots \wedge e_n \longrightarrow e_{n+1} \vee \ldots \vee e_{n+m}$.
Gleichungen der Form $P(x_1, \ldots, x_n)$ {\tt = true} k"onnen wie im Beispiel
durch $P(x_1, \ldots, x_n)$ abgek"urzt werden. Syntaktisch korrekt ist eine
solche abgek"urzte Gleichung jedoch nur dann, wenn {\tt true} innerhalb des
Moduls sichtbar und sortengleich mit der Zielsorte von $P$ ist.
In \ASFm\ werden alle Beweisziele exportiert. Auf Flags zur Beschr"ankung
der Sichtbarkeit, wie sie in ART \Eschbach\ Verwendung finden, wird verzichtet.
\ASFm\ versteht sich als Eingabeschnittstelle zu einem Beweiser, nicht als
Ausgabeschnittstelle. Deshalb wird auch auf solche Flags verzichtet, die Auskunft
dar"uber geben, welche der Klauseln als bewiesen gelten d"urfen und welche
nicht. Der Stempel ``proved'' ohne einen Verweis auf den Beweis, ist ohnehin
von zweifelhaftem Wert, zumal kaum "uberpr"uft werden kann, ob die
Spezifikation nach setzen des Flags vom Benutzer ver"andert wurde.
Es wird davon ausgegangen, da"s der Beweiser f"ur die bearbeitete Spezifikation
eine Datei anlegt, die Informationen "uber die Spezifikation enth"alt
(zum Beispiel den Namen des Top-Moduls und Datum+Zeit der letzten
Spezifikationsmodifikation) und neben allen bewiesenen Theoremen
auch Referenzen auf die Beweise beinhaltet.

\pagebreak %---------------------------------------------------------------------

\subsection{Parametrisierte Module}

Die bisher eingef"uhrten Konstrukte erscheinen ausreichend f"ur
Bottom-Up-Spezifika\-tionen. W"unschenswert sind jedoch auch Mechanismen,
die es gestatten, Freir"aume innerhalb eines Modules zu erhalten, die erst
sp"ater (z.~B.\ beim Import des Moduls in ein weiteres) mit konkretem
Inhalt gef"ullt werden m"ussen. Das Parameterkonzept von \ASFm\ gestattet es,
Sorten und Funktionen in ein parametrisiertes Modul nachtr"aglich durch 
Parameterbindung zu ``implantieren''.
Als Beispiel betrachten wir das Modul {\tt Sequences},
in dem Sequenzen von nicht n"aher spezifizierten Elementen definiert werden.
Als Konstruktoren dienen {\tt nil} (erzeugt die leere Sequenz) und {\tt cons}
(f"ugt ein Element an eine Sequenz an).

\begin{verbatim}
module Sequences <(ITEMpar)>
short  Seq
{     
   add signature
   {
      parameters:
      (  sorts
            ITEMpar  )
      public:
         sorts
            SEQ
         constructors
            nil      :               -> SEQ
            cons     : ITEMpar # SEQ -> SEQ
   }
} /* Sequences */
\end{verbatim}

In \ASFm\ m"ussen alle formalen Parameter (ob importiert, oder wie im
Beispiel im  {\tt add signature}-Konstrukt deklariert) an prominenter
Stelle direkt hinter dem Modulnamen in spitzen Klammern angegeben werden. 
Beim Auftreten mehrerer Parameter kann mit Hilfe der runden Klammern die
Zahl der m"oglichen Parameterbindungen eingeschr"ankt werden. Alle Parameter
eines durch runde Klammern eingefa"sten Tupels d"urfen nur an Namen
desselben Moduls gebunden werden.

Auch {\tt OrdSequences} (unten) spezifiziert Sequenzen "uber eine durch
Bindung des Parameters {\tt ITEMpar} zu pr"azisierende Sorte von Elementen.
In Frage kommen hier jedoch nur Sorten, f"ur die eine irreflexive Ordnung
spezifiziert wurde. Mit Hilfe dieser Ordnung wird eine lexikographische Ordnung
auf Sequenzen definiert.

\label{OrdSequences}
\begin{verbatim}
module OrdSequences <(ITEMpar, ordpar)>
short  OSeq
{
   import Booleans {public: BOOL, true, false}
   
   add signature
   {
      parameters:
         (  sorts
               ITEMpar
            non-constructors
               ordpar : ITEMpar # ITEMpar -> BOOL
            conditions
               [irref] ordpar(i1,i1) 
                  -->
               [trans] ordpar(i1,i2), ordpar(i2,i3)
                  -->  ordpar(i1,i3)
               [total]
                  -->  ordpar(i1,i2), ordpar(i2,i1), i1 = i2
         )
      public:
         sorts
            SEQ
         constructors
            nil      :               -> SEQ
            cons     : ITEMpar # SEQ -> SEQ
         non-constructors
            greater  : SEQ     # SEQ -> BOOL
   }
   variables
   {  i1, i2, i3         : -> ITEMpar
      seq1, seq2, s1, s2 : -> SEQ  }

   equations
   {
      macro-equation greater(seq1, seq2)
      {              /* lex-order on sequences */
         case
         {
            ( seq1 @ nil )                     : false
            ( seq1 @ cons(i1, s1), seq2 @ nil ): true
            ( seq1 @ cons(i1, s1), seq2 @ cons(i2, s2) ):
               if ( ordpar(i1, i2) )
                     true
                  else if ( i1 = i2 ) 
                             greater(s1, s2)
                          else
                             false
         }
      }
   }
} /* OrdSequences */
\end{verbatim}

\ASFm\ verzichtet im Gegensatz zu ASF auf die Einf"uhrung
eines formalen Parameters f"ur den Modulnamen, an den ein Parameter-Tupel
gebunden wird. Als Parameter werden in \ASFm\ statt dessen die Sorten- 
und Funktionsnamen innerhalb der Parameter-Tupel bezeichnet.
Die Gruppierung der Parameter in Bl"ocke (hier Tupel,
dargestellt durch runde Klammern) wird jedoch beibehalten,
weil sie sich bei der Formulierung semantischer Bedingungen als vorteilhaft
erweist.
Funktionsparameter k"onnen nicht "uberladen werden.

Unter ``semantischen Bedingungen'' verstehen wir in \ASFm\ Gentzenklauseln,
die in der Definition eines Parameter-Tupels im Parameterteil des
{\tt add signatur}-Konstrukts angegeben werden k"onnen
(hier {\tt irref}, {\tt trans} und {\tt total}).
Die Zul"assigkeit der Bindung eines Parameter-Tupels an Namen 
eines Moduls $M_{ACT}$ h"angt nun davon ab, ob die aus der Bindung hervorgehenden
Gentzenklauseln innerhalb von $M_{ACT}$ gelten oder nicht.
Da dieses Problem im allgemeinen unentscheidbar ist wird zus"atzlich
gefordert, da"s $M_{ACT}$ Beweisziele enth"alt, die sich nur durch die Marken-
und Variablennamen von den Bedingungsklauselinstanzen unterscheiden und
f"ur die bereits Beweise existieren.
Mit Bedingungen verkn"upfte Parameter-Tupel k"onnen nur an Namen
solcher Module gebunden werden, die keine ungebundenen Parameter mehr enthalten,
weil f"ur Module mit freien Parametern (bisher) keine Semantik
innerhalb des ASF-Ansatzes existiert.

Mit dem Konzept der semantischen Bedingungen werden vor allem zwei Ziele
verfolgt: Einerseits werden semantisch unsinnige Parameterbindungen
schon in der Akzeptanzphase der Spezifikation erkannt, au"serdem k"onnen
diese Bedingungen in parametrisierten Beweisen als Lemmas von Bedeutung
sein, weil sie die ``wesentlichen'' Eigenschaften der Parameter enthalten.

Beim Import eines parametrisierten Moduls sind alle Parametertupel hinter
dem Modulnamen in eckigen Klammern aufzuf"uhren:
\begin{verbatim}
   import OrdSequences <(ITEMpar, ordpar)>
   {  public: SEQ, nil, cons  }
\end{verbatim}
Da Parameter nicht verdeckt werden k"onnen, entspricht diese
Syntax dem Grundsatz, da"s alle innerhalb eines Moduls
sichtbaren Namen dort auch angegeben werden m"ussen.

\subsection{Das Namensraumkonzept}

Eine Grundidee des flexiblen Verdeckungsmechanismus aus \ASFm\ ist die
eindeutige Zuordnung von
Namen zu Namensr"aumen. Im wesentlichen beschreibt der Namensraum das Modul,
in dem der Name zum ersten Mal in Erscheinung tritt (im Folgenden als
Definitionsmodul bezeichnet). Im Beispiel {\tt Naturals} geh"oren
unter anderem {\tt NAT} und {\tt 0} dem Namensraum {\tt Naturals} und
{\tt BOOL} und {\tt true} dem Namensraum {\tt Booleans} an. Der Name {\tt x}
kommt in beiden Namensr"aumen als Variable in unterschiedlicher Bedeutung
vor. Ein Namensraum umfa"st also alle innerhalb eines Moduls eingef"uhrten
Namen (einschlie"slich Marken) abz"uglich der importierten.
Die Namen eines Moduls geh"oren im allgemeinen also verschiedenen Namensr"aumen
an. Wir bezeichnen den Namensraum, dem die im Modul definierten Namen angeh"oren,
als den moduleigenen Namensraum (er erbt auch den Namen des Moduls), alle
anderen hei"sen importierte Namensr"aume. Namen aus verschiedenen Importbefehlen
k"onnen nur dann miteinander identifiziert werden, wenn sie dem gleichen
Namensraum angeh"oren, was bei mehrfachem Import desselben Moduls der Fall
sein kann.
Was geschieht aber, wenn Namen Namensr"aumen von Modulen angeh"oren,
die durch Renaming oder Parameterbindung beim Import ``manipuliert'' wurden?
\ASFm\ l"ost das Problem durch Schaffung neuer Namensrauminstanzen, die
Kopien der urspr"unglichen Namensr"aume repr"asentieren. Das Kopieren einzelner
Namen aus ASF wird durch gruppenweises Kopieren ersetzt, deren kleinste
Einheiten die Namensr"aume bilden. Die schwerwiegenden Gr"unde f"ur diese
konzeptionelle Entscheidung werden im Kapitel \ref{Der kopierende Import} diskutiert.

\subsection{Explizites Renaming}

Unter explizitem Renaming verstehen wir in \ASFm\ das Umbenennen von
Signatur- und Parameternamen aus importierten Modulen mit Hilfe des
{\tt renamed to}-Konstrukts.
\begin{verbatim}
module Integers
short  Int
{
   import Naturals[Int1]
   {  public: NAT renamed to INT, 0, s, +, eq  }
   
   add signature {  public: constructors p : INT -> INT }
   
   variables {  x, y : -> INT  }
   
   equations
   {  [e1]  s(p(x))  = x
      [e2]  p(s(x))  = x
      [e3]  p(x) + y = p(x + y)  }
      
} /* Integers */
\end{verbatim}

{\tt Integers} spezifiziert den Datentyp der ganzen Zahlen unter Verwendung
der nat"urlichen Zahlen.
In \ASFm\ wird erwartet, da"s jeder Importbefehl, in dem ein explizites
Renaming oder eine Parameterbindung vorgenommen wird, eine
innerhalb der Spezifikation eindeutige (m"oglichst kurze) Instanzbezeichnung
(im Beispiel {\tt Int1}) beinhaltet. Sie wird gebraucht, um Namen unterschiedlich
instanziierter Namensr"aume zu unterscheiden.

Im letzten Beispiel geh"oren u.~a.\ 
{\tt INT}, {\tt 0} und {\tt s} dem neuen Namensraum {\tt Naturals[Int1]} an. 
{\tt Naturals[Int1]} ist dabei eine Instanz (bzw. Kopie) des Namensraumes
{\tt Naturals}, die durch das explizite Renaming im
Importbefehl geschaffen wurde. Nat"urlich k"onnen auch instanziierte Namesr"aume
bei einem weiteren Import manipuliert werden:
\begin{verbatim}
   import Integers[Int2]{  public : INT renamed to INTnew  }
\end{verbatim}

{\tt INTnew} geh"ort, wie auch beispielsweise
der hier nicht mehr sichtbare Name {\tt 0}, nun dem Namensraum
{\tt Naturals[Int1,Int2]} an.

Die hierachische Struktur einer Spezifikation
bedingt Abh"angigkeiten zwischen Namensr"aumen. Im Beispiel f"uhrt die Umbenennung
von {\tt INT} des Namensraumes {\tt Naturals[Int1]} nach {\tt INTnew} auch zu
einer "Anderung der Konstruktordeklaration
f"ur {\tt p} des Namensraumes {\tt Integers} (Definitions- und Wertebereich
werden ge"andert),
der Namensraum {\tt Booleans} bleibt dagegen unbeeinflu"st. \ASFm\ tr"agt diesem
Umstand Rechnung, indem der Konstruktor {\tt p} und die Variablen {\tt x} und
{\tt y} aus {\tt Integers} dem neuen Namensraum {\tt Integers[Int2]} zugeordnet
werden. {\tt BOOL} geh"ort nach wie vor dem Namensraum {\tt Booleans} an.
Allgemein h"angt ein moduleigener Namensraum von allen importierten
Namensr"aumen ab, was bei der Modifikation von Namen aus indirekt importierten
Modulen zur Instanziierung mehrerer Namensr"aume f"uhrt. 

Jede Instanzbezeichnung darf innerhalb einer Spezifikation nur ein einziges mal 
verwendet werden. Da zwischen Modulk"urzeln und Instanzbezeichnungen keine
Verwechselungsgefahr besteht, bietet es sich an, das Modulk"urzel als
Instanzbezeichnung wiederzuverwenden, sofern im Modul nur ein instanziierender
Import vorgenommen wird.

\subsection{Parameterbindungen}

\begin{verbatim}
module OrdNatSequences
short  ONSeq
{
   import OrdSequences[ONSeq] <(ITEMpar bound to NAT,
                                ordpar  bound to greater) of OrdNaturals >
   {  public: SEQ renamed to NSEQ, 
              nil renamed to Nnil,
              cons, greater,
              BOOL, true, false  }

   import OrdNaturals
   {  public: NAT, greater, 0, s  }
}
\end{verbatim}

Analog zu ASF werden Parameter blockweise an ein Modul gebunden. 
Semantisch gesehen bedeutet die Bindung von Parametern eines Moduls $M_{FORM}$ 
(im Beispiel {\tt OrdSeqences}) an Namen eines Moduls $M_{ACT}$ (im Beispiel 
{\tt OrdNaturals}) einerseits, da"s Parameternamen aus $M_{FORM}$ durch
Namen aus $M_{ACT}$ ersetzt werden. Letztere k"onnen entweder
exportierbare Signaturnamen oder Parameter sein. Da sie jedoch nur im
Kontext des Moduls $M_{ACT}$ eine Bedeutung besitzen, m"ussen andererseits
beide Module miteinander kombiniert werden.
Der Importblock, der der Parameterbindung folgt, bestimmt ausschlie"slich
die Sichtbarkeit der Signaturnamen des Moduls $M_{FORM}$. Explizites Renaming
ist zul"assig. Die Signaturnamen des Moduls $M_{ACT}$ (auch die aktuellen
Parameter selbst, sofern sie nicht wieder Parameter sind) gelten im
bindenden Modul (im Beispiel {\tt OrdNatSequences})
als verdeckt, es sei denn, ein weiterer (direkter) Import nimmt wie im Beispiel
Einflu"s auf die Sichtbarkeit einzelner Namen. Explizites Renaming ist in diesem
Falle jedoch kaum sinnvoll, weil sonst die Signaturnamen des zus"atzlich
importierten Moduls aufgrund der unterschiedlichen Namensrauminstanzen
nicht mit denen aus $M_{ACT}$ identifiziert werden.

Genau wie das explizite Renaming f"uhrt auch das Binden von Parametern zur
Instanziierung der direkt betroffenen und aller davon abh"angigen Namensr"aume.
Um auch ohne den direkten Import von $M_{ACT}$ eine vollst"andige Signatur
zu garantieren, sorgt die Semantik der Parameterbindung daf"ur, da"s neben
dem Modul $M_{FORM}$ auch automatisch $M_{ACT}$ (verdeckt) in das bindende Modul
importiert wird --- ein Vorgang, der im folgenden als {\em impliziter Import\/}
bezeichnet wird. 

{\tt greater} kann als Demonstrationsbeispiel f"ur eine "uberladene
Funktion gesehen werden.
In {\tt OrdNatSequences} referenziert der Name sowohl eine irreflexive Ordnung
auf den nat"urlichen Zahlen als auch auf Sequenzen.

Da jede Parameterbindung zu einer Instanziierung des Namensraumes der zu
bindenden Parameter und aller davon abh"angigen Namensr"aume bis hin zum
Modul $M_{FORM}$ f"uhrt (diese Namensr"aume fallen zusammen, falls wie in
unseren Beispielen $M_{FORM}$ die Parameter selbst definiert,
also nicht importiert), ist es
auch m"oglich, Module ``an sich selbst'' zu binden, ohne da"s es zu einer
unerw"unschten Vermischung der dort eingef"uhrten Strukturen kommt. 

{\tt SeqOfSeq} spezifiziert Sequenzen von Elementen,
die selbst Sequenzen sind. Um Namenskollisionen zwischen Signaturnamen
der Module $M_{FORM}$ und $M_{ACT}$ zu vermeiden ist eine die Umbenennung
aller Sorten und Konstanten, deren Sichtbarkeit im bindenden Modul erw"unscht
ist (im Beispiel {\tt SEQ} und {\tt nil}) zwingend notwendig.
Die sowohl aus $M_{FORM}$ als auch aus $M_{ACT}$ importierten Konstruktoren
{\tt cons} unterscheiden sich in ihren Argumentsorten und d"urfen daher
"uberladen werden.

\begin{verbatim}
module SeqOfSeq <(ITEMpar)>
short  SOS
{
   import Sequences[SOS] <( ITEMpar bound to SEQ )
                             of Sequences <(ITEMpar)> >
   {  public: SEQ renamed to SEQ1,
              nil renamed to nil1,
              cons  }
   
   import Sequences <(ITEMpar)>
   {  public: SEQ, nil, cons  }
}
\end{verbatim}

\vfill\pagebreak

\section{Strukturdiagramme}
Die modulare Struktur von \ASFm-Spezifikationen kann mit Hilfe von
Strukturdiagrammen
veranschaulicht werden. Alle Namen innerhalb eines importfreien Moduls geh"oren
demselben (moduleigenen) Namensraum an. Er wird durch ein Rechteck, genannt
Namensraumbox dargestellt, in dem zentriert unter der Oberkante die
Namensraumbezeichnung (= Modulname) steht.

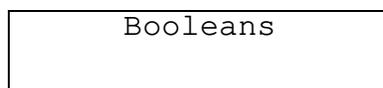
\placepic{
\begin{picture}(160,50)(0,-10)
\drawline(160,0)(160,35)(0,35)
        (0,0)(160,0)
\put(80,25){\makebox(0,0)[b]{\tt Booleans}}
\end{picture}}

Enth"alt das darzustellende Modul Importbefehle, so kann der Import durch
ineinander verschachtelte Boxen dargestellt werden.
Sie symbolisieren die hierarchische Struktur der Namensr"aume, die im
Modul, dessen moduleigener Namensraum durch die "au"serste Box gegeben
ist, eine Rolle spielen. Ein Namensraum ist von
allen Namensr"aumen abh"angig, die seine Box umschlie"st.

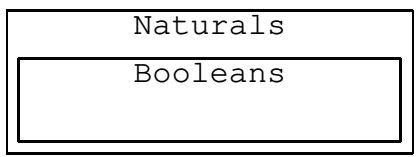
\placepic{
\begin{picture}(170,75)(0,-10)
\drawline(165,5)(165,40)(5,40)
        (5,5)(165,5)
\put(85,30){\makebox(0,0)[b]{\tt Booleans}}
\drawline(170,0)(170,60)(0,60)
        (0,0)(170,0)
\put(85,50){\makebox(0,0)[b]{\tt Naturals}}
\end{picture}}

Im {\tt add signatur}--Konstrukt eines Moduls enthaltene Parametertupel werden
oberhalb der Boxen f"ur importierte Namensr"aume in Sechsecken aufgef"uhrt.

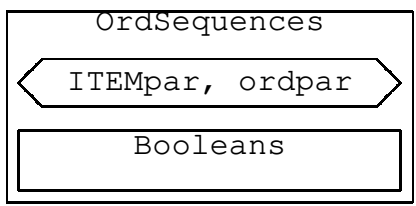
\placepic{
\begin{picture}(170,95)(0,-10)
\drawline(5,50)(15,40)(155,40)
        (165,50)(155,60)(15,60)
        (5,50)(5,50)
\drawline(165,5)(165,30)(5,30)
        (5,5)(165,5)
\drawline(170,0)(170,80)(0,80)
        (0,0)(170,0)
\put(85,45){\makebox(0,0)[b]{\tt ITEMpar, ordpar}}
\put(85,70){\makebox(0,0)[b]{\tt OrdSequences}}
\put(85,20){\makebox(0,0)[b]{\tt Booleans}}
\end{picture}}

Werden beim Import Namen eines Moduls ge"andert oder Parameter gebunden,
f"uhrt das in \ASFm\ dazu, da"s alle direkt betroffenen, sowie die davon
abh"angigen Namensr"aume mit der Instanzbezeichnung des Importbefehls
instanziiert werden. Eine fehlerhafte Identifikation von Namen aus diesen
manipulierten R"aumen mit den ``Originalen'' ist dadurch ausgeschlossen.

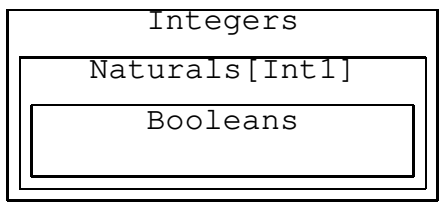
\placepic{
\begin{picture}(180,96)(0,-10)
\drawline(170,10)(170,40)(10,40)
        (10,10)(170,10)
\drawline(175,5)(175,60)(5,60)
        (5,5)(175,5)
\drawline(180,0)(180,80)(0,80)
        (0,0)(180,0)
\put(90,30){\makebox(0,0)[b]{\tt Booleans}}
\put(90,70){\makebox(0,0)[b]{\tt Integers}}
\put(90,50){\makebox(0,0)[b]{\tt Naturals[Int1]}}
\end{picture}}

\vfill\pagebreak

Das Binden von Parametern eines Moduls $M_{FORM}$ (im Bsp. {\tt OrdSequence}) an
Namen eines weiteren Moduls $M_{ACT}$ (im Beispiel {\tt OrdNaturals}) wird durch
einen Pfeil angedeutet. Die Richtung des Pfeils verdeutlicht die Abh"angigkeit
zwischen den Namensr"aumen der Module $M_{ACT}$ und $M_{FORM}$. Der instanziierte
Namensraum der zu bindenden Parameter sowie alle von
ihm abh"angigen Namensr"aume h"angen von jedem in $M_{ACT}$ enthaltenen
Namensraum ab.

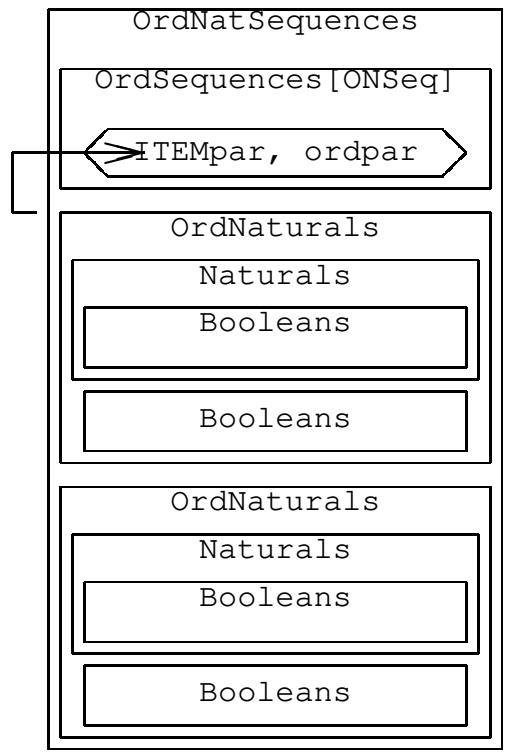
\placepic{
\begin{picture}(205,327)(0,-10)
\drawline(190,45)(190,70)(30,70)
        (30,45)(190,45)
\drawline(195,40)(195,90)(25,90)
        (25,40)(195,40)
\drawline(190,10)(190,35)(30,35)
        (30,10)(190,10)
\drawline(200,5)(200,110)(20,110)
        (20,5)(200,5)
\put(110,100){\makebox(0,0)[b]{\tt OrdNaturals}}
\put(110,80){\makebox(0,0)[b]{\tt Naturals}}
\put(110,60){\makebox(0,0)[b]{\tt Booleans}}
\put(110,20){\makebox(0,0)[b]{\tt Booleans}}
\drawline(190,160)(190,185)(30,185)
        (30,160)(190,160)
\drawline(195,155)(195,205)(25,205)
        (25,155)(195,155)
\drawline(190,125)(190,150)(30,150)
        (30,125)(190,125)
\drawline(200,120)(200,225)(20,225)
        (20,120)(200,120)
\put(110,175){\makebox(0,0)[b]{\tt Booleans}}
\put(110,195){\makebox(0,0)[b]{\tt Naturals}}
\put(110,215){\makebox(0,0)[b]{\tt OrdNaturals}}
\put(110,135){\makebox(0,0)[b]{\tt Booleans}}
\drawline(40,240)(180,240)(190,250)
        (180,260)(40,260)(30,250)(40,240)
\drawline(200,235)(200,285)(20,285)
        (20,235)(200,235)
\drawline(205,0)(205,310)(15,310)
        (15,0)(205,0)
\thicklines
\drawline(39.000,246.000)(55.000,250.000)(39.000,254.000)
\drawline(55,250)(0,250)(0,225)(10,225)
\put(110,245){\makebox(0,0)[b]{\tt ITEMpar, ordpar}}
\put(110,275){\makebox(0,0)[b]{\tt OrdSequences[ONSeq]}}
\put(110,300){\makebox(0,0)[b]{\tt OrdNatSequences}}
\end{picture}}

Da neben dem impliziten, durch die Parameterbindung verursachten Import
von $M_{ACT}$ ein zus"atzlicher (direkter) Import erforderlich ist,
um Signaturnamen aus $M_{ACT}$ f"ur das bindende Modul sichtbar zu machen
f"uhren wir eine kompaktere Darstellung ein, in der wir den impliziten und
den direkten Import (falls vorhanden) zu einer Box zusammenfassen.

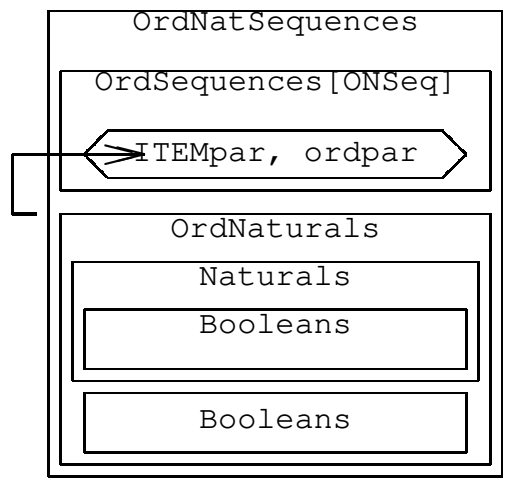
\placepic{
\begin{picture}(205,212)(0,-10)
\drawline(190,45)(190,70)(30,70)
        (30,45)(190,45)
\drawline(195,40)(195,90)(25,90)
        (25,40)(195,40)
\drawline(190,10)(190,35)(30,35)
        (30,10)(190,10)
\drawline(200,5)(200,110)(20,110)
        (20,5)(200,5)
\put(110,60){\makebox(0,0)[b]{\tt Booleans}}
\put(110,80){\makebox(0,0)[b]{\tt Naturals}}
\put(110,100){\makebox(0,0)[b]{\tt OrdNaturals}}
\put(110,20){\makebox(0,0)[b]{\tt Booleans}}
\drawline(40,125)(180,125)(190,135)
        (180,145)(40,145)(30,135)(40,125)
\drawline(200,120)(200,170)(20,170)
        (20,120)(200,120)
\thicklines
\drawline(39.000,131.000)(55.000,135.000)(39.000,139.000)
\drawline(55,135)(0,135)(0,110)(10,110)
\thinlines
\drawline(205,0)(205,195)(15,195)
        (15,0)(205,0)
\put(110,130){\makebox(0,0)[b]{\tt ITEMpar, ordpar}}
\put(110,160){\makebox(0,0)[b]{\tt OrdSequences[ONSeq]}}
\put(110,185){\makebox(0,0)[b]{\tt OrdNatSequences}}
\end{picture}}

\pagebreak %----------------------------------------------------------------

Erweitert werden k"onnen die \ASFm-Strukturdiagramme durch Hinzunahme der Signatur.
Jedes Modul zerf"allt zun"achst in zwei Bereiche. Links stehen die sichtbaren,
rechts die verdeckten Signaturnamen. Der linke Bereich der sichtbaren Namen
zerf"allt seinerseits in zwei Sichtbarkeitsstufen:
Neben den \public--deklarierten Namen, die
vom betreffenden Modul exportiert werden k"onnen (auf die also importierende Module
zugreifen k"onnen), gibt es noch die \private--deklarierten Namen, welche
nur innerhalb des Moduls sichtbar sind und auch nur dort referenziert
werden k"onnen.
Insgesamt existieren also die drei Bereiche ``public'', ``private'' und ``hidden'',
die durch zwei gepunktete senkrechte Trennungslinien dargestellt werden
k"onnen.

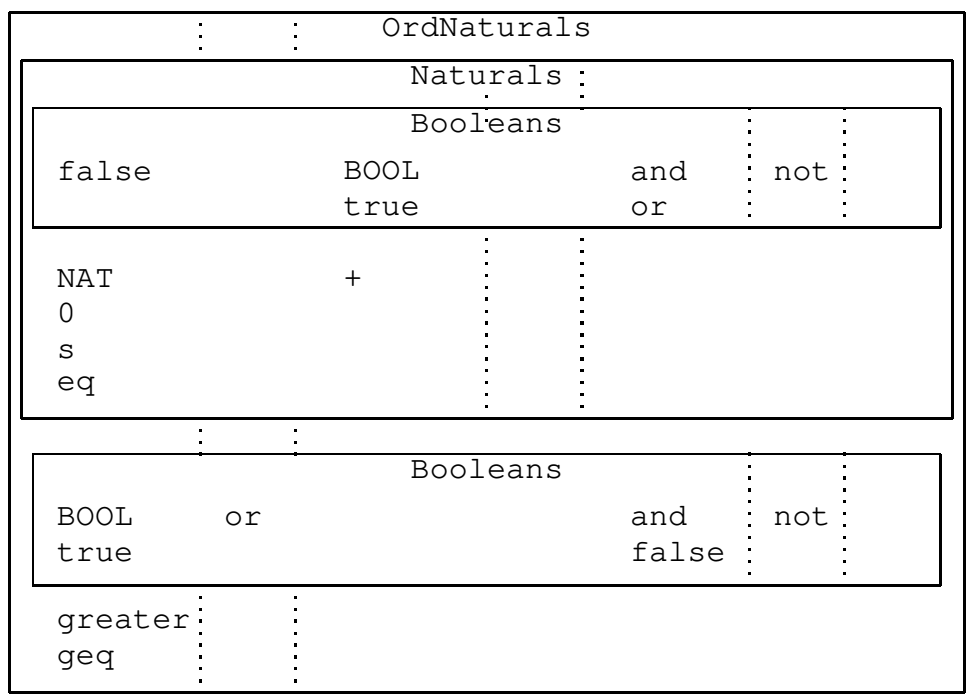
\placepic{
\begin{picture}(400,300)(0,-10)
\put(20,170){\makebox(0,0)[lb]{{\tt NAT}}}
\put(20,155){\makebox(0,0)[lb]{{\tt 0}}}
\put(20,140){\makebox(0,0)[lb]{{\tt s}}}
\put(20,125){\makebox(0,0)[lb]{{\tt eq}}}
\put(20,25){\makebox(0,0)[lb]{{\tt greater}}}
\put(20,10){\makebox(0,0)[lb]{{\tt geq}}}
\put(260,215){\makebox(0,0)[lb]{{\tt and}}}
\put(260,200){\makebox(0,0)[lb]{{\tt or}}}
\put(260,70){\makebox(0,0)[lb]{{\tt and}}}
\put(260,55){\makebox(0,0)[lb]{{\tt false}}}
\put(140,215){\makebox(0,0)[lb]{{\tt BOOL}}}
\put(140,200){\makebox(0,0)[lb]{{\tt true}}}
\drawline(390,195)(390,245)(10,245)
        (10,195)(390,195)
\drawline(395,115)(395,265)(5,265)
        (5,115)(395,115)
\drawline(390,45)(390,100)(10,100)
        (10,45)(390,45)
\drawline(400,0)(0,0)(0,285)
        (400,285)(400,0)
\thicklines
\dottedline{5}(310,245)(310,195)
\dottedline{5}(350,245)(350,195)
\dottedline{5}(350,100)(350,45)
\dottedline{5}(310,100)(310,45)
\dottedline{5}(240,195)(240,115)
\dottedline{5}(200,195)(200,115)
\dottedline{5}(200,240)(200,250)
\dottedline{5}(240,265)(240,245)
\dottedline{5}(120,265)(120,285)
\dottedline{5}(80,265)(80,285)
\dottedline{5}(120,100)(120,115)
\dottedline{5}(80,100)(80,115)
\dottedline{5}(120,45)(120,0)
\dottedline{5}(80,45)(80,0)
\put(200,275){\makebox(0,0)[b]{{\tt OrdNaturals}}}
\put(200,255){\makebox(0,0)[b]{{\tt Naturals}}}
\put(200,235){\makebox(0,0)[b]{{\tt Booleans}}}
\put(20,215){\makebox(0,0)[lb]{{\tt false}}}
\put(320,215){\makebox(0,0)[lb]{{\tt not}}}
\put(200,90){\makebox(0,0)[b]{{\tt Booleans}}}
\put(20,70){\makebox(0,0)[lb]{{\tt BOOL}}}
\put(20,55){\makebox(0,0)[lb]{{\tt true}}}
\put(320,70){\makebox(0,0)[lb]{{\tt not}}}
\put(140,170){\makebox(0,0)[lb]{{\tt +}}}
\put(90,70){\makebox(0,0)[lb]{{\tt or}}}
\end{picture}
}

Im Beispiel sind innerhalb von {\tt Naturals} die Namen {\tt NAT}, {\tt 0},
{\tt s}, {\tt eq}, {\tt +} und die importierten Namen {\tt false}, {\tt BOOL},
{\tt true} sichtbar. Nach dem Import in {\tt OrdNaturals} bleiben davon
zun"achst lediglich die Namen {\tt NAT}, {\tt 0},
{\tt s}, {\tt eq} und {\tt false} "ubrig. {\tt +}, {\tt BOOL}, und {\tt true}
werden hier hingegen nicht sichtbar. Der zweite
Import des Moduls {\tt Booleans} sorgt daf"ur, da"s auch f"ur {\tt OrdNaturals}
{\tt BOOL} und {\tt true} sichtbar sind. Hauptzweck dieses Imports ist es
jedoch, die Referenzierbarkeit von {\tt or} f"ur {\tt OrdNaturals} zu erreichen,
was beim indirekten Import "uber {\tt Naturals} nicht m"oglich war.
Am Beispiel wird deutlich, da"s bei mehrfachem Import desselben Moduls
ein Name in unterschiedlichen Sichtbarkeitsstufen auftreten kann und wird.
Die Sichtbarkeit im importierenden Modul richtet sich bei \ASFm\ in diesem
Fall nach der gr"o"sten importierten Sichtbarkeit (Auftreten am weitesten
links im Strukturdiagramm). Gleichnamige Parameter, gleichnamige Sorten
sowie gleichnamige Funktionen mit gleichen Argumentsorten, die innerhalb
eines Moduls sichtbar sind und unterschiedlichen Namensr"aumen angeh"oren,
stellen einen Namenskonflikt, also einen Spezifikationsfehler, dar.

\section{Semantik hierarchischer Konzepte}
\label{hierachische Konzepte}
F"ur hierarchische Konzepte algebraischer Spezifikationssprachen sind
grunds"atzlich zwei Semantikans"atze denkbar:
\begin{itemize}
\item Jedes Modul erh"alt eine Semantik. Die Semantik einer
   hierarchisch modularisierten Spezifikation errechnet sich aus den
   einzelnen Modulsemantiken.
\item Nur f"ur elementare (flache) Spezifikationen wird eine algebraische
   Semantik definiert. Hierarchischen Spezifikationen wird mit Hilfe eines 
   Normalform-Algorithmus eine elementare Spezifikation zugewiesen, deren
   Semantik die Semantik der hierarchischen Spezifikation definiert.
   Die Bedeutung der Importkonstrukte ist hier eine auf der Syntax von
   Spezifikationsmodulen und nicht auf deren Semantiken definierte Funktion.
\end{itemize}

Obwohl hinsichtlich der Modularisierung von Beweisen die erste Variante 
interessante Perspektiven bietet, f"allt unsere Wahl aufgrund der hohen
Komplexit"at und der vielen offenen Fragen in bezug auf praktische
Ad"aquatheit einer geeigneten Modulsemantik auf die zweite.
Ein Vorteil dieser auch bei ASF angewandten Vorgehensweise ist die gute
Operationalisierbarkeit.
Von zentraler Bedeutung ist die Normalisierungsprozedur, da mit ihr (indirekt)
die Semantik der einzelnen Importbefehle festgelegt wird. Im folgenden sollen
grunds"atzliche M"oglichkeiten beleuchtet, Schwachstellen der ASF-Semantik
erl"autert und Alternativen aufgezeigt werden.

\subsection{Der ``benutzende'' Import}
Lassen wir zun"achst das Verdeckungskonzept au"ser acht und verzichten
au"serdem auf die M"oglichkeit, Funktionen zu "uberladen.
Dann kann man sich die Bedeutung eines renamingfreien Importbefehls ohne
Parameterbindungen in erster N"aherung als eine ``komponentenweise''
Vereinigung des importierten Moduls mit dem importierenden
Modul vorstellen. Die Sortennamenmenge des resultierenden Moduls ergibt sich
als Vereinigungsmenge der Sortennamen des importierten und des importierenden
Moduls. Gleiches gilt f"ur Konstruktor- und Non-Konstruktor-Funktionsdeklarationen,
Parametertupel, Variablendeklarationen, Gleichungen, Beweisziele und,
mit Ausnahme des gerade ausgewerteten Importbefehls (der nun gel"oscht
werden kann), auch f"ur die Importbefehle. Der Modulname des resultierenden
Moduls ist durch die komponentenweise Vereinigung nicht festgelegt.
Die Normalform einer mit Hilfe solcher Importbefehle hierarchisch
strukturierten Spezifikation berechnet sich dann als komponentenweise
Vereinigung aller direkt und indirekt importierten Module mit dem Top-Modul.
Die Reihenfolge, mit der die Importbefehle eleminiert werden, spielt dabei
f"ur das resultierende Normalformmodul keine Rolle. Ein Spe\-zifi\-ka\-tions\-fehler
liegt vor, wenn bei der Vereinigung ein inkorrektes Modul erzeugt wird.

\pagebreak

Alternativ k"onnen die Importbefehle eines Moduls auch in zwei Schritten
eleminiert werden: Zun"achst werden die importierten Module
untereinander und danach das Zwischenresultat mit dem importierenden Modul
``vereinigt''. Dieses Vorgehen liefert bei der bisher betrachteten
eingeschr"ankten Form von Importbefehlen das gleiche Resultat.
Die Vereinigung der Importbefehlmenge mu"s in diesem Fall sinngem"a"s modifiziert
werden: Bei der komponentenweisen Vereinigung im ersten Schritt
(wir schreiben $\bigsqcup$) m"ussen alle Importbefehle der vereinigten
(importierten) Module im Zwischenresultat ber"ucksichtigt werden, w"ahrend im
zweiten Schritt (hier schreiben wir $\sqcup$) nur die Importbefehle
des Zwischenresultats (und nicht die des importierenden Moduls)
in das Resultat "ubernommen werden d"urfen.

Dies erlaubt nun die folgende Operationalisierung der Vereinigungssemantik,
welche den Vorteil hat, da"s alle Zwischenergebnisse Normalformen sind, was
bei der Behandlung von verdeckten Namen von Vorteil ist.
\begin{itemize}
\item Die Normalform eines importfreien Moduls ist das Modul selbst.
\item Die Normalform eines Moduls $M$, welches $M_1, \ldots, M_n$ importiert
  ergibt sich aus der komponentenweisen Vereinigung der Normalformen von $M_1,
  \ldots, M_n$ und $M$.
\end{itemize}

\noindent Wir schreiben:

\NForm($M$)\ :=\ $\left\{ 
   \begin{array}{ll} 
      M
      & \mbox{falls $M$ importfrei}

\\    M \cunion\ \Cunion_{i=1}^n \NForm(M_i)
      & \mbox{falls $M_1, \ldots, M_n$ von $M$ importiert werden}.
   \end{array}\right.$

Die Vereinigungssemantik ist invariant gegen"uber mehrfachem Import des
gleichen Moduls, auch ist die Reihenfolge der Importe ohne Bedeutung.
Entscheidend bleibt lediglich, welche Module importiert werden. Diese
Eigenschaften sind typisch f"ur eine bestimmte Art von Importen, die wir
``benutzende Importe'' nennen.

Die Einfachheit der Vereinigungssemantik wird jedoch mit einem schweren
Defekt erkauft. Sie identifiziert Sorten und Funktionsdeklarationen aus
verschiedenen Herkunftsmodulen im Falle zuf"alliger syntaktischer Gleichheit,
auch wenn sie nichts miteinander zu tun haben. Nur wenige Konflikte
zwischen Modulen, die den gleichen Namen in unterschiedlicher
Bedeutung benutzen, werden erkannt.

Eine modifizierte Version der ``Vereinigungssemantik'' sollte also pr"ufen,
ob es innerhalb der Spezifikation einen Namen gibt, der in zwei Modulsignaturen
unterschiedlich definiert wird. In diesem Fall (wir gehen von sichtbaren,
nicht "uberladbaren Namen aus) liegt ein Namenskonflikt vor. Diese
Modifikation kann auf die rekursive Variante nicht ohne weiteres "ubertragen
werden, da den Signaturnamen der Normalformen der zu importierenden Module
nicht direkt angesehen werden kann, welchem Modul sie ihre Entstehung
verdanken.

\pagebreak %-------------------------------

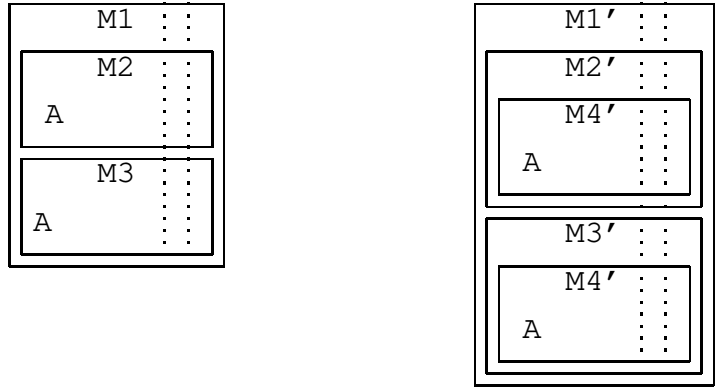
\placepic{
\begin{picture}(295,175)(0,-10)
\drawline(85,100)(85,140)(5,140)
        (5,100)(85,100)
\drawline(85,55)(85,95)(5,95)
        (5,55)(85,55)
\drawline(90,50)(90,160)(0,160)
        (0,50)(90,50)
\drawline(290,75)(290,140)(200,140)
        (200,75)(290,75)
\drawline(285,80)(285,120)(205,120)
        (205,80)(285,80)
\drawline(295,0)(295,160)(195,160)
        (195,0)(295,0)
\drawline(290,5)(290,70)(200,70)
        (200,5)(290,5)
\drawline(285,10)(285,50)(205,50)
        (205,10)(285,10)
\thicklines
\dottedline{5}(75,50)(75,160)
\dottedline{5}(65,50)(65,160)
\dottedline{5}(275,0)(275,160)
\dottedline{5}(265,0)(265,160)
\put(45,150){\makebox(0,0)[b]{\tt M1}}
\put(45,130){\makebox(0,0)[b]{\tt M2}}
\put(45,85){\makebox(0,0)[b]{\tt M3}}
\put(245,150){\makebox(0,0)[b]{\tt M1'}}
\put(245,130){\makebox(0,0)[b]{\tt M2'}}
\put(245,60){\makebox(0,0)[b]{\tt M3'}}
\put(245,110){\makebox(0,0)[b]{\tt M4'}}
\put(245,40){\makebox(0,0)[b]{\tt M4'}}
\put(15,110){\makebox(0,0)[lb]{\tt A}}
\put(10,65){\makebox(0,0)[lb]{\tt A}}
\put(215,90){\makebox(0,0)[lb]{\tt A}}
\put(215,20){\makebox(0,0)[lb]{\tt A}}
\end{picture}}

In obigem Beispiel tritt der Sortenname {\tt A} sowohl links in den Signaturen der
Normalformen von $M_2$ und $M_3$ (sie sind bereits in Normalform) als auch
rechts in den
Signaturen der Normalformen von $M_2'$ und $M_3'$ auf. W"ahrend dies in $M_1$
zu einem Namenskonflikt f"uhrt, k"onnen in $M_1'$ beide Namen identifiziert
werden, da sie aus der gleichen Definition in $M_4'$ hervorgegangen sind.
Den Normalformmodulen ist das jedoch nicht mehr zu entnehmen.
ASF l"ost das Problem durch Einf"uhrung einer Originfunktion. Sie weist
jedem Signaturnamen einen Informationsblock (Origin) zu, der u.~a.\ den Namen
des Moduls enth"alt, welches f"ur die Definition des Namens verantwortlich
ist. Tritt in zwei zu importierenden Normalformmodulen der gleiche
Signaturname auf, kann anhand der zugeordneten Origins entschieden werden,
ob es sich um einen Namenskonflikt handelt oder nicht.

Ein modifizierter Normalformalgorithmus k"onnte folgenderma"sen aussehen:
Sei $\{M_i \quad |\quad 1 \leq i \leq n\}$ die Menge der vom Spezifizierer
erzeugten Module einer Spezifikation, $\modn_i$ der Name des Moduls $M_i$
und $\{\sign_{i,j} \quad | \quad 1 \leq j \leq m_i\}$ die Menge aller
Signaturnamen des Moduls $M_i$.
Wir definieren zu jedem Modul eine Originfunktion

$\begin{array}{lclcl}
\Ur_i &: &\{\sign_{i,j}\quad|\quad 1 \leq j \leq m_i\} &\longrightarrow
  &\{\modn_i\} \\
\Ur_i &:= &\{\sign_{i,j}\quad|\quad 1 \leq j \leq m_i\} &\times
  &\{\modn_i\}.
\end{array}$

\yestop
\noindent
Die Normalform eines Moduls $M_i$ errechnet sich rekursiv wie folgt:

$\NForm(M_i)\ :=\  \left\{
   \begin{array}{ll}
      (M_i, \Ur_i)
      & \mbox{falls $M_i$ importfrei}
\\
      (M_i \cunion \Cunion_{k=1}^{p_i} M'_{i'_k},\ 
                        \Ur_i \cup \bigcup_{k=1}^{p_i} o'_{i'_k})
      & \begin{array}[t]{@{}l@{}}
           \mbox{falls $M_i$ die Module $M_{i'_k}$ importiert und}
\\
           \mbox{$(M'_{i'_k}, o'_{i'_k}) = \NForm(M_{i'_k})$ 
                 gilt $(1 \leq k \leq {p_i})$.}
        \end{array}
   \end{array}
\right.$

\noindent
Ein Namenskonflikt liegt genau dann vor, wenn
$(\displaystyle \Ur_i \cup \bigcup_{k=1}^{p_i} o'_{i'_k})$
keine Funktion ist.

Der angegebene Algorithmus liefert genau
die Semantik renamingfreier Importe ohne Parameterbindung f"ur sichtbare
nicht "uberladene Namen aus ASF bzw.\ \ASFm.

\vfill

\pagebreak

\subsection{Der ``kopierende'' Import}
\label{Der kopierende Import}
Besonders in gro"sen Spezifikationen wird es h"aufig zu Namenskonflikten kommen,
weil die Zahl der Namen mit jedem neuen Modul w"achst. W"urde das Aufl"osen
solcher Konflikte das Edieren der verantwortlichen Module erzwingen, z"oge
das gleichzeitig Namens"anderungen in allen Modulen nach sich, die auf
das edierte Modul zugreifen. Der Spezifizierer h"atte bei der Erstellung
eines neuen Moduls darauf zu achten, da"s alle neu eingef"uhrten Namen in
keinem anderen bisher vorhandenen Modul verwendet werden, was der 
Konzeption des modularen Spezifizierens nicht entspricht. Deshalb
stellt ASF ein Renamingkonstrukt zur Verf"ugung, welches das Umbenennen von
Namen beim Import erm"oglicht. Leider f"uhrt jedoch die ASF-Bedeutung
dieses Konstrukts zum Vermischen unterschiedlicher Strukturen, wie das
folgende ASF-Beispiel zeigt:

\begin{verbatim}
module exA
begin
   exports
      begin sorts A
            functions
               mk_A : -> A
      end
end exA
\end{verbatim}

Die Anweisung ``\ {\tt imports exA \{  renamed by [mk\_A -> make\_A] \}}\ ''
bedeutet in ASF den Import eines Moduls namens {\tt exA}, das sich vom Original
{\tt exA} dadurch unterscheidet, da"s jedes Auftreten vom Signaturnamen
{\tt mk\_A} durch {\tt make\_A} ersetzt wurde. Das erscheint sinnvoll,
solange innerhalb einer Spezifikation nur mit einer Version des Moduls
gearbeitet wird. "Au"serst unsch"on erweist sich die Semantik jedoch beim Import
mehrerer Varianten eines Moduls:

\begin{verbatim}
Module Murks
begin
   imports exA,
           exA {  renamed by [mk_A -> make_A]  }
end Murks
\end{verbatim}

Die Semantik von ASF kann zwischen beiden Instanzen des importierten Moduls
{\tt exA} nicht unterscheiden, was dazu f"uhrt, da"s  {\tt Murks} "uber zwei
Konstruktoren f"ur die Sorte {\tt A} verf"ugt. Das namenweise Kopieren 
kann in gr"o"seren Spezifikationen leicht dazu f"uhren, da"s Namen, die
nicht direkt am expliziten Renaming beteiligt sind, f"alschlich miteinander
identifiziert werden.

\pagebreak

Zu derartig unmotivierten Namensidentifikationen kommt es in ASF auch beim
Import verschiedener, durch Parameterbindungen aktualisierter Versionen
des gleichen Moduls. Als Demonstrationsbeispiel untersuchen wir Sequenzen
"uber nat"urlichen Zahlen und Boole'schen Werten in ASF:

\begin{verbatim}
module Sequences
begin
   parameters
      Items begin
               sorts ITEM
            end Items
   exports
      begin
         sorts SEQ
         functions nil :            -> SEQ
                   cons: ITEM # SEQ -> SEQ
      end
end Sequences
\end{verbatim}

\yestop
\begin{verbatim}
module Auwei
begin
    imports Sequences
              {  Items bound by [ITEM -> NAT]  to Naturals  },
            Sequences
              {  Items bound by [ITEM -> BOOL] to Booleans  }
end Auwei
\end{verbatim}

\yestop\yestop
Die Module {\tt Naturals} und {\tt Booleans} seien sinngem"a"s (analog
zu den gleichnamigen \ASFm-Modulen) definiert. Das
Modul {\tt Auwei} importiert zwei verschiedene Arten von Sequenzen. Beide
Arten tragen jedoch den gleichen Sortennamen {\tt SEQ}, was eigentlich einen
Namenskonflikt erwarten lie"se. Statt dessen werden jedoch von ASF beide
Sorten miteinander identifiziert, was dazu f"uhrt, da"s
{\tt cons(s(0), cons(true, nil))} als wohlsortierter Term der Sorte {\tt SEQ}
akzeptiert wird. Dies entspricht sicherlich nicht den Vorstellungen des
Spezifizierers!

Lassen wir weiterhin verdeckte Namen und Overloading au"ser acht, dann
kann das Renaming aus ASF als Erweiterung der modifizierten Vereinigungssemantik
gesehen werden:

\pagebreak
\noindent
$\NForm(M_i) := \left\{
   \begin{array}{ll}
      (M_i, \Ur_i)
      & \mbox{falls $M_i$ importfrei}
\\
      (M_i \cunion \Cunion_{k=1}^{p_i} R_{i'_k}(M'_{i'_k}),\ 
                        \Ur_i \cup  \bigcup_{k=1}^{p_i} R_{i'_k}(o'_{i'_k}))
      & $\begin{tabular}[t]{@{}l@{}}
           falls $M_i$ die Module $M_{i'_k}$ impor- \\
           tiert und  $(M'_{i'_k}, o'_{i'_k}) = \NForm(M_{i'_k})$ \\
           gilt $(1 \leq k \leq {p_i})$.
         \end{tabular}$
   \end{array}
\right.$

Hier ist $R_{i'_k}$ eine Funktion, die Signaturnamen des zu importierenden Moduls
nach Ma"sgabe des Renamingkonstrukts (falls vorhanden) durch andere ersetzt,
und auf Module und Originfunktionen angewendet werden kann. Falls der
Importbefehl f"ur das Modul $M_{i'_k}$ kein Renamingkonstrukt enth"alt, ist
$R_{i'_k}$ die Identit"at.

Unver"andert bleiben in dieser Erweiterung (wie auch bei der hier nicht
formalisierten Erweiterung f"ur Parameterbindungen) die Modulnamen im
Wertebereich der Originfunktionen. So ist es zwar einerseits
m"oglich, vorhandene Module zu modifizieren, andererseits k"onnen diese
verschiedenen Aktualisierungen dann nicht unterschieden werden, was bei
Mehrfachimporten zu ungew"unschter Vermischung der Strukturen f"uhrt.

\ASFm\ geht hier einen anderen Weg. Die Modulnamen im Wertebereich der
Originfunktion werden als Namensraumbezeichnungen interpretiert. Manipulationen
wie explizites Renaming oder das Binden von Parametern stellen einen
schwerwiegenden Eingriff in die den beteiligten Namen zugeordneten Namensr"aume
dar. Um sicher zu stellen, da"s Namen aus den ver"anderten Namensr"aumen nicht
mit Namen des urspr"unglichen Namensraumes identifiziert werden, ordenet \ASFm\
den ver"anderten Namensr"aumen neue Bezeichnungen zu. Diese setzen sich aus
den alten Bezeichnungen und den Instanzbezeichnungen der instanziierenden
Importbefehle zusammen. Wir sagen: Die Namensr"aume werden instanziiert.

Ein interessanter Fall tritt ein, wenn durch Renaming oder Parameterbindung
ein Modul ver"andert wird, das selbst weitere Module importiert, dessen
(Signatur-) Namen also verschiedenen Namensr"aumen angeh"oren. Ein undifferenziertes
Instanziieren aller Namensr"aume w"urde zu zahlreichen "uberfl"ussigen Namenskonflikten
f"uhren. Beispielsweise beeinflu"st
das Binden des Parametertupels von {\tt OrdSequences} (siehe Seite
\pageref{OrdSequences}) beim Import 
das indirekt importierte Modul {\tt Booleans} in keinster Weise, so da"s der
Identifikation der Sorte {\tt BOOL} mit dem Orginal (welches m"oglicherweise mittels
weiterer Befehle importiert wird) nichts entgegen steht. 
Andererseits k"onnen Manipulationen, die beim kopierenden Import vorgenommen
werden auch indirekt importierte Teilsignaturen betreffen.
In diesem Fall gen"ugt es nicht, nur die Namensr"aume der direkt betroffenen
Signaturnamen zu instanziieren. Vielmehr m"ussen ebenfalls alle Namensr"aume,
die von den instanziierten Namensr"aumen abh"angen bis hin zum Namensraum des
direkt importierten Moduls instanziiert werden.

Allgemein f"uhrt das Manipulieren von indirekt importierten
Modulsignaturen zur Instanziierung mehrerer Namensr"aume.
Zur Illustration betrachten wir ein \ASFm-Beispiel:

\begin{minipage}{11cm}
\begin{verbatim}
module exA
{
   add signature{  public: sorts A  }
}

module exAB
{
   import exA   {  public: A  }
   add signature{  public: sorts B  }
}

module exABC
{
   import exAB  {  public: A, B  }
   add signature{  public: sorts C  }
}

module CopyDemo
{
   import exABC[Copy]{  public: A,
                                B renamed to Bnew,
                                C  }
   import exABC      {  public: A  }
   import exABC      {  public: C  }
}
\end{verbatim}
\end{minipage}
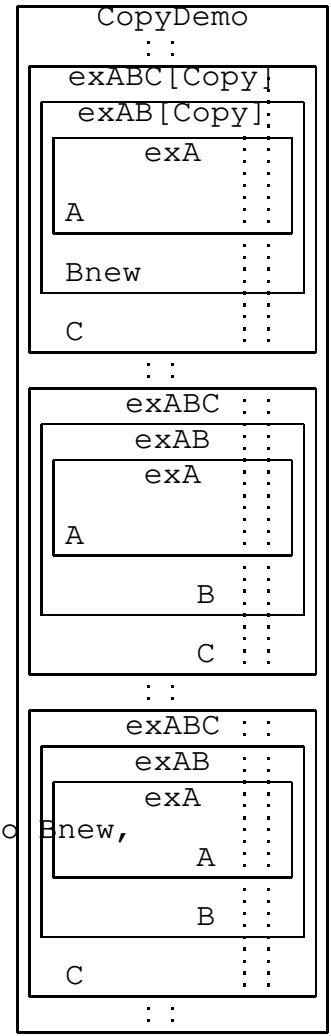
\begin{minipage}{4cm}
\begin{picture}(130,447)(0,-10)
\drawline(115,65)(115,105)(15,105)
        (15,65)(115,65)
\drawline(120,40)(120,120)(10,120)
        (10,40)(120,40)
\drawline(125,15)(125,135)(5,135)
        (5,15)(125,15)
\put(65,125){\makebox(0,0)[b]{\tt exABC}}
\put(65,110){\makebox(0,0)[b]{\tt exAB}}
\put(65,95){\makebox(0,0)[b]{\tt exA}}
\put(20,20){\makebox(0,0)[lb]{\tt C}}
\put(75,70){\makebox(0,0)[lb]{\tt A}}
\put(75,45){\makebox(0,0)[lb]{\tt B}}
\drawline(115,200)(115,240)(15,240)
        (15,200)(115,200)
\drawline(120,175)(120,255)(10,255)
        (10,175)(120,175)
\drawline(125,150)(125,270)(5,270)
        (5,150)(125,150)
\put(65,260){\makebox(0,0)[b]{\tt exABC}}
\put(65,245){\makebox(0,0)[b]{\tt exAB}}
\put(65,230){\makebox(0,0)[b]{\tt exA}}
\put(20,205){\makebox(0,0)[lb]{\tt A}}
\put(75,180){\makebox(0,0)[lb]{\tt B}}
\put(75,155){\makebox(0,0)[lb]{\tt C}}
\drawline(115,335)(115,375)(15,375)
        (15,335)(115,335)
\drawline(120,310)(120,390)(10,390)
        (10,310)(120,310)
\drawline(125,285)(125,405)(5,405)
        (5,285)(125,285)
\put(65,365){\makebox(0,0)[b]{\tt exA}}
\put(65,380){\makebox(0,0)[b]{\tt exAB[Copy]}}
\put(65,395){\makebox(0,0)[b]{\tt exABC[Copy]}}
\put(20,315){\makebox(0,0)[lb]{\tt Bnew}}
\put(20,340){\makebox(0,0)[lb]{\tt A}}
\put(20,290){\makebox(0,0)[lb]{\tt C}}
\drawline(130,0)(130,430)(0,430)
        (0,0)(130,0)
\thicklines
\dottedline{5}(105,135)(105,15)
\dottedline{5}(95,135)(95,15)
\dottedline{5}(105,270)(105,150)
\dottedline{5}(95,270)(95,150)
\dottedline{5}(105,405)(105,285)
\dottedline{5}(95,375)(95,285)
\dottedline{5}(55,415)(55,405)
\dottedline{5}(55,285)(55,270)
\dottedline{5}(55,150)(55,135)
\dottedline{5}(55,15)(55,0)
\dottedline{5}(65,415)(65,405)
\dottedline{5}(65,285)(65,270)
\dottedline{5}(65,150)(65,135)
\dottedline{5}(65,15)(65,0)
\put(65,420){\makebox(0,0)[b]{\tt CopyDemo}}
\end{picture}

\end{minipage}

Der erste Importbefehl von {\tt CopyDemo} manipuliert die Signatur des (indirekt)
importierten Moduls {\tt exAB}. Dies f"uhrt zu einer Instanziierung des 
zugeordneten Namensraumes --- {\tt Bnew} geh"ort nun dem neuen Namensraum {\tt exAB[Copy]} an.
Auf die Signatur des ebenfalls (indirekt) importieren Moduls {\tt exA} hat das keinen
Einflu"s, daher k"onnen die Sorten {\tt A} aus den ersten beiden Importbefehlen
identifiziert werden und nach wie vor dem Namensraum {\tt exA} angeh"oren.
Eine Manipulation in der Signatur von {\tt exAB} hat Einflu"s auf die Signatur
des {\tt exAB} importierenden Moduls {\tt exABC}, weil hier die ver"anderten Namen
sichtbar sind und im allgemeinen auch in den Funktionsdeklarationen auftreten werden.
\ASFm\ ordnet dem Sortennamen {\tt C} im ersten Importbefehl den instanziierten Namensraum
{\tt exABC[Copy]} zu. Im dritten Importbefehl geh"ort {\tt C}
dagegen dem (nicht modifizierten) Namensraum {\tt exABC} an. Die Identifikationsregel
sieht darin einen Namenskonflikt und wird die vorliegende Spezifikation
nicht akzeptieren.
ASF hingegen w"urde die beiden Sorten {\tt C} identifizieren, was im allgemeinen die
weiter oben bereits aufgezeigten Probleme bereitet.

In \ASFm\ bleibt der Namenskonflikt auch dann bestehen, wenn
der erste Importbefehl durch
\begin{verbatim}
   import exABC[Copy]{  public: A, B renamed to B, C  }
\end{verbatim}
ersetzt wird.
Der Ausdruck $name_1$ {\tt renamed to} $name_2$ hat im Kontext eines
\ASFm-Imports also zwei verschiedene Auswirkungen. Neben der Zugriffs"anderung
bewirkt er auch eine Instanziierung eines oder mehrerer Namensr"aume. Ist man nur
an letzterer Wirkung interessiert, kann ein Ausdruck $name$ {\tt renamed
to} $name$ sinnvoll sein, was in \ASFm\ auch als {\tt copy of} $name$
geschrieben werden kann.

In beiden F"allen l"ost sich der Namenskonflikt auf, wenn der dritte
Importbefehl in {\tt CopyDemo} entfernt wird.

Die Realisation der hier vorgestellten Semantik erfordert zwei "Anderungen in der
modifizierten Vereinigungssemantik. Zun"achst mu"s der Wertebereich der Originfunktionen
auf Namensraumbezeichnungen ausgedehnt werden. Sie setzen sich aus Modulnamen und
Instanzbezeichnungen zusammen.
Die Originfunktion des normalisierten Moduls {\tt exABC} kann beispielsweise
folgenderma"sen dargestellt werden:
\{({\tt A},{\tt exA}), ({\tt B},{\tt exAB}), ({\tt C},{\tt exABC})\}.
Neben der Umbenennung von {\tt B} zu {\tt Bnew} ver"andert das explizite
Renaming auch den Wertebereich der Originfunktion:
\{({\tt A},{\tt exA}), ({\tt Bnew},{\tt exAB[Copy]}), ({\tt C},{\tt exABC[Copy]})\}.
Um ermitteln zu k"onnen, welche Namensr"aume instanziiert werden m"ussen, wird
au"serdem Information "uber den hierarchischen Aufbau der Spezifikation ben"otigt.
Aus diesem Grund f"uhren wir zur Erkl"arung der Semantik von \ASFm\
eine zus"atzliche Funktion namens Dependenzfunktion ein, die
jedem innerhalb eines Moduls auftretenden Namensraum die Menge aller
Namensr"aume zuordnet, die von ihm abh"angen. Sie wird im folgenden
Abschnitt \ref{Abhaengigkeiten zwischen Namensraeumen} diskutiert.

Die hier vorgestellte Sorte von Importen bezeichnen wir als kopierende
Importe. Ihre Verwendung ist immer dann sinnvoll, wenn ``wesentliche''
Eigenschaften der importierten Struktur ge"andert werden sollen.
Was aber sind ``wesentliche'' Eigenschaften? Neben den bereits diskutierten
Signaturmanipulationen (explizites Renaming und Parameterbindung) k"onnen
im importierenden Modul auch neue Konstruktoren
und Funktionen zu einer importierten Sorte bereitgestellt und
im Gleichungsblock neue Beziehungen zwischen Elementen der importierten
Struktur definiert werden (z.~B.\ zwecks Erweiterung partiell definierter
Funktionen). All dies hat Einflu"s auf die G"ultigkeit von Klauseln.
W"urde man in allen F"allen kopierenden Import verlangen, h"atte das
zur Folge, da"s bewiesene Beweisziele eines Moduls beim benutzenden
Import des Moduls in ein anderes ihre G"ultigkeit behalten w"urden --- ein
denkbar einfacher Beweismodularisierungsansatz. Formal erscheint diese
``seiteneffektfreie'' Semantik des benutzenden Imports optimal. Auch
kann die Einhaltung der Restriktionen vom Normalformalgorithmus syntaktisch
gepr"uft werden. Andererseits scheinen
die Bedingungen f"ur praktischen Gebrauch zu restriktiv, weil unn"otig
viele Namen und Instanzbezeichnungen den Blick auf das Wesentliche versperren.
\ASFm\ schreibt den kopierenden Import nur bei Manipulationen
durch Renaming und Parameterbindung vor und "uberl"a"st in allen anderen F"allen
dem Spezifizierer die Wahl des Importtyps.

Die folgende Spezifikation einer zyklischen Gruppe mit drei Elementen
{\tt Nat3} kann als Anschauungsbeispiel daf"ur dienen, wie der kopierende
Import auch "uber Renaming und
Parameterbindung hinaus als Spezifikationshilfsmittel sinnvoll
eingesetzt werden kann:

\pagebreak %-----------------------------------------------------------------

\begin{verbatim}
module Nat3
{
    import Naturals[Nat3]
    {  public: copy of NAT,
               0, s, +  }
    
    variables  {  x: -> NAT  }
    equations  {  [e1] s(s(s(x))) = x  } 
}
\end{verbatim}

Die Gleichung {\tt e1} nimmt destruktiven Einflu"s auf die importierte
Datenstruktur.
W"urden hier die Namen {\tt NAT} und {\tt s} mit den Originalen aus
{\tt Naturals} identifiziert, so st"ande das unverf"alschte Original f"ur
die gesamte Spezifikation nicht mehr zur Verf"ugung.
Mit Einf"uhrung zus"atzlicher Restriktionen (z.~B.\ ``Verbot des Auftretens
von Termen aus ausschlie"slich benutzend importierten Funktionssymbolen
als linke Seite einer Gleichung im {\tt equations}--Block.'') k"onnte der
kopierende Import von {\tt Naturals} erzwungen werden.

\subsection{Abh"angigkeiten zwischen Namensr"aumen}
\label{Abhaengigkeiten zwischen Namensraeumen}

Die hierarchische Struktur der Spezifikation bedingt Abh"angigkeiten
zwischen den erzeugten Namensr"aumen. Die Semantik von \ASFm\ wird ihnen
durch Einf"uhrung einer sogenannten Dependenzfunktion gerecht, welche
der Bezeichnung jedes Namensraumes die Menge der Bezeichnungen aller von
ihm abh"angigen Namensr"aume zuweist. Diese Dependenzfunktion soll
hier diskutiert werden.

Ein importfreies Modul $M$ namens $\modn$ enth"alt nur einen Namensraum,
n"amlich den moduleigenen. Seine Bezeichnung stimmt mit dem Modulnamen "uberein,
Abh"angigkeiten zu anderen Namensr"aumen bestehen nicht. Die zugeh"orige
Dependenzfunktion lautet also

$\depf\quad :=\quad \{(\modn, \emptyset)\}$.

Der moduleigene Namensraum eines nicht importfreien Moduls $M$ namens $\modn$ ist von
allen importierten Namensr"aumen abh"angig. Die zugeh"orige Dependenzfunktion $\depf$
kann aus den Dependenzfunktionen, die sich aus den einzelnen Importbefehlen ergeben,
berechnet werden. Zu diesem Zweck definieren wir eine Hilfsfunktion
\CombineDependencies, die eine Menge von Dependenzfunktionen zu einer Funktion
zusammenfa"st.

\label{CombineDependencies}
\begin{tabbing}
\indent $\CombineDependencies(\{\depf_i\quad|\quad i \in A\})\quad :=$
\\ \indent\indent
   $\{\ (\modinst, \displaystyle\bigcup_{i\in B} \depf_i(\modinst))\quad|\quad$\=
            $\modinst \in \displaystyle\bigcup_{i \in A}\dom(\depf_i)\quad \wedge$
\\       \> $B = \{i \in A \ | \ \modinst \in \dom(\depf_i)\}\ \}$
\end{tabbing}

Mit Hilfe von \CombineDependencies\ kann nun die Dependenzfunktion $\depf$ f"ur
beliebige, nicht notwendigerweise importfreie Module definiert
werden:

\begin{tabbing}
\indent $\depf\ :=\ $
    \=$\{\ (\modn,  \emptyset)\ \}\quad\cup$
\\  \> \{\ \=$(\modinst,\modinstances\cup\{\modn\})\quad |\quad(\modinst, \modinstances)$
\\  \>     \>$%(\modinst, \modinstances)
               \in \CombineDependencies(\{\depfimpconst_i\ |\ 1 \leq i \leq l\})\ \}$
\end{tabbing}

\noindent wobei $\depfimpconst_i$ die zum $i$-ten Importbefehl des Moduls $M$
zugeh"orige Dependenzfunktion beinhaltet ($1 \leq i \leq l$). Wie die zu einem Importbefehl
zugeh"orige Dependenzfunktion \depfimpconst\ aus der Dependenzfunktion
\depfimpmod\ des importierten Moduls zu berechnen ist h"angt vom Importtyp
ab und wird im folgenden erkl"art.

Handelt es sich um einen benutzenden
Import des Moduls {\em M-imp} und ist \depfimpmod\ die Dependenzfunktion des
Moduls, so gilt $\depfimpconst := \depfimpmod$.

Handelt es sich dagegen um einen kopierenden Import von {\em M-imp}, in dem explizites
Renaming durchgef"uhrt wird, dann geht \depfimpconst\ aus \depfimpmod\ dadurch
hervor, da"s jedes Auftreten von Bezeichnungen der vom Renaming direkt betroffenen
Namensr"aume sowie der von diesen bez"uglich \depfimpmod\ abh"angigen
Namensr"aume durch die mit der Instanzbezeichnung des Importbefehls
instanziierten Namensraumbezeichnung ersetzt wird.

Werden formale Parameter an Namen aus $k$ aktuellen Modulen {\em M-act}$_j$
($1\leq j\leq k$) gebunden, so sind zus"atzlich die Namensraumbezeichnungen
aller formalen
Parameter, an die aktuelle Parameter gebunden werden, sowie alle bez"uglich
\depfimpmod\ von ihnen abh"angigen Namensr"aume in \depfimpmod\ zu
instanziieren. Die resultierende Funktion nennen wir $\depfimpmod'$. Des weiteren
sind die Namensraumabh"angigkeiten der implizit importierten aktuellen Module
$\depfactmod_j$ zu ber"ucksichtigen. Sei analog zum expliziten Import

$\depfimpconst\ :=\ \CombineDependencies(%
\begin{array}[t]{@{}l}
   \{\depfimpmod'\}\ \cup \\
   \{\depfactmod'_j\ |\ 1\leq j\leq k\}).
\end{array}$

\noindent W"urde man hier $\depfactmod'_j$\ mit $\depfactmod_j$\ gleichsetzen,
so entspr"achen die
aus implizitem Import resultierenden Abh"angigkeiten innerhalb des bindenden Moduls
denen eines benutzenden Imports. Unber"ucksichtigt blieben dabei jedoch die Beziehungen
zwischen dem Modul der formalen Parameter (hier {\em M-imp}) und den Modulen der
aktuellen Parameter (hier {\em M-act}$_j$). Dies ist jedoch erforderlich: Werden
beispielsweise bei einem sp"ateren Import des bindenden Moduls (hier $M$) aktuelle Parameter
aus der Bindung umbenannt, so hat dies auch Einflu"s auf die Namen des Moduls der formalen
Parameter. Allgemein definieren wir daher:

\begin{tabbing}
\indent$\depfactmod'_j\quad :=\quad
   \{\ $\=$(\modinst, \modinstances\ \cup\ 
                   $\=$\{\paradefmodinst_j\}\ \cup$
\\           \>     \>$\depfimpmod'(\paradefmodinst_j)\quad|$
\\           \>$(\modinst, \modinstances) \in \depfactmod_j\ \}$,
\end{tabbing}

\noindent wobei $\paradefmodinst_j$ der Namensraum der an Namen des Moduls
{\em M-act}$_j$ zu bindenden formalen Parameter des Moduls {\em M-imp} ist.

Im folgenden Beispiel werden die Abh"angigkeiten zwischen den Namensr"aumen aus
{\tt OrdNatSequences} durch die zugeh"orige Dependenzfunktion dargestellt. Sie kann
auch aus dem in Kapitel 3 vorgestellten Strukturdiagramm gewonnen werden.

\begin{tabbing}
\indent
$\{\ $\=$({\tt Booleans},\quad \{\ {\tt Naturals}, {\tt OrdNaturals},
                              {\tt OrdSequences[ONSeq]}, {\tt OrdNatSequences}\ \}),$
\\    \>$({\tt Naturals},\quad \{\ {\tt  OrdNaturals}, {\tt OrdSequences[ONSeq]},
                              {\tt OrdNatSequences}\ \}),$
\\    \>$({\tt OrdSequences[ONSeq]},\quad \{\ {\tt OrdNatSequences}\ \})\ \}$
\end{tabbing}

\vfill\pagebreak

\subsection{Verdeckte Namen}
Im Prinzip k"onnten alle Namenskonflikte, die beim Import von Modulen
auftreten, durch ``explizite'' Umbenennungen (s.~o.)\ aufgel"ost werden.
Allerdings erfordert dies vom Spezifizierer einen "Uberblick "uber alle
eingef"uhrten Namen, was mit zunehmender
Spezifikationskomplexit"at immer schwieriger wird. Um den Spezifizierer vom
Umbenennen ``unwichtiger'' Namen zu entlasten, unterscheidet ASF sichtbare
und verdeckte Namen. W"ahrend die Konfliktl"osung zwischen sichtbaren Namen
weiterhin in der Verantwortung des Spezifizierers liegt, werden Konflikte
zwischen verdeckten Namen vom Normalformalgorithmus durch automatisches
Umbenennen (implizites Renaming) aufgel"ost.

 %Verdeckte Signaturnamen dienen in ASF als Hilfsfunktionen und -sorten, die
 %im spezifizierten Modell nur innerhalb der Kongruenzrelation ber"ucksichtigt
 %werden.
ASF beschr"ankt die Referenzierbarkeit verdeckter Namen jeweils auf das
definierende Modul, was f"ur Variablen (sie k"onnen in diesem Sinne als
verdeckt betrachtet werden) ausreichend ist.
Um die Zahl der Konflikte zwischen Sorten- und Funktionsnamen wirksam
zu reduzieren, erscheint diese Einschr"ankung jedoch zu restriktiv.
W"unschenswert w"are ein Mechanismus, der es erlaubt, Namen, die in der
jeweiligen Spezifikationsebene nicht mehr gebraucht werden, ``auszublenden''.
Modulare Programmiersprachen stellen zu diesem Zweck Ex- und Importlisten
zur Verf"ugung.

\ASFm\ "ubernimmt die Importlisten (alle weiterhin sichtbaren Namen
m"ussen im Importkonstrukt aufgef"uhrt werden). Das Exportverhalten von Namen
wird dagegen direkt in der Definition bzw.\ beim Import durch die
Schl"usselworte \private\ und \public\ festgelegt. Dies reduziert
den Code und dient der "Ubersicht.
Werden Namen beim Import verdeckt, so ersetzt der Normalformalgorithmus alle
diese Namen durch neue, innerhalb der gesamten Spezifikation eindeutige
Namen. Zu diesem Zweck wird dem alten vom Spezifizierer vereinbarten Namen
der (abgek"urzte) Name des entsprechenden Namensraumes gefolgt von einem
Bindestrich vorangestellt. Beispielsweise werden die Namen {\tt and}, {\tt or}
und {\tt not} aus {\tt Booleans} beim verdeckten Import in das Modul
{\tt Naturals} durch {\tt Bo-and}, {\tt Bo-or} und {\tt Bo-not} ersetzt.
Neben der Trennungsfunktion zwischen Namensraum und urspr"unglichem Namen
garantiert der Bindestrich die Konfliktfreiheit zwischen verdeckten und
sichtbaren Namen, da er in letztgenannten nicht zugelassen ist.

Besonders n"utzlich erweist sich das Instrument der Namensverdeckung in Verbindung
mit dem kopierenden Import, bei dem zahlreiche Namensumbenennungen erforderlich
werden, da Signaturnamen aus verschiedenen Instanzen eines Moduls nicht miteinander
identifiziert werden d"urfen.
\ASFm\ erledigt das f"ur die verdeckten Namen automatisch, der Spezifizierer
mu"s sich lediglich um die sichtbaren, ihn interessierenden Namen k"ummern.

\subsection{Overloading}
Bisher gingen wir davon aus, da"s jeder Signaturname genau ein Signaturobjekt
(Sorte oder Funktion) spezifiziert. In der Praxis ist es jedoch sehr n"utzlich,
wenn verschiedene Objekte mit dem gleichen Namen referenziert werden k"onnen.
Beispielsweise schreibt man gew"ohnlich die Summe zweier Zahlen $x$ und
$y$ als $(x + y)$, egal, ob es sich bei $x$ und $y$ um nat"urliche, ganze oder
rationale Zahlen handelt. Die tats"achliche Bedeutung des Namens ``$+$''
ergibt sich aus dem Kontext.

Das aus ASF "ubernommene Overloading gestattet es, Funktionsnamen zu
"uberladen, wenn diese sich in ihren Argumentsorten unterscheiden.
Die Restriktion erlaubt es, durch Bottom-Up-Sortenpr"ufung jedem Funktionsnamen
innerhalb eines Terms eine eindeutige Funktion zuzuordnen.
Um "uberladene Funktionen behandeln zu k"onnen, m"ussen wir im Definitionsbereich
der Originfunktion zu {\em disambiguierten} Namen "ubergehen. Dabei handelt es
sich um Tupel (\specname, \sortvector) bestehend aus dem Signaturnamen
und einem (f"ur n-stellige Funktionen n-dimensionalen) Sortenvektor. Jede
Funktionsdeklaration im {\tt add signature}-Konstrukt definiert genau einen neuen
disambiguierten Namen. Der Import eines Funktionsnamens zieht im allgemeinen
den Import mehrerer disambiguierter Namen nach sich.
Ist als Ergebnis der Normalisierung eine "uberladungsfreie Spezifikation
gew"unscht, kann dies erreicht werden, indem alle Funktionsnamen des
Normalformmoduls durch eine geeignete Repr"asentation ihrer disambiguierten
Namen (z.~B.\ {\tt +[NAT,NAT]}) ersetzt werden.

\vfill\pagebreak

\section{Syntax}

Die Syntax von \ASFm\ ist gegeben durch folgende kontextfreie Grammatik\footnote%
{Wir kennzeichenen Terminale durch Anf"uhrungszeichen und Typewriterfont und
 Nichtterminale durch spitze Klammern (\nts{$\ldots$}). \rep{$x$} bedeutet null, eine oder
 mehrere und \reps{$x$} eine oder mehrere Wiederholungen von $x$, \repwt{$x$}{$ts$}
 und \repswt{$x$}{$ts$} stehen f"ur Wiederholungen von $x$, getrennt durch das
 Terminalsymbol $ts$. Optionale Zeichenketten sind in eckige Klammern ([$\ldots$])
 eingefa"st.}:

\noindent
\begin{tabular}{@{}r@{\hspace{0.5em}}r@{\hspace{0.5em}}l@{}}
   \Gspezification      &::=& \reps{\Gmodule}
\\ \Gmodule             &::=& \ts{module} \Gmodulename
                              \opt{\ts{<} \reps{\Gparameterblock} \ts{>}}
\\                      &   & \opt{\ts{short}\ \Gshortmodulename}
\\                      &   & \ts{\{}\hspace*{1ex}
\begin{tabular}[t]{l}
   \rep{\Gimport}
\\ \opt{\Gaddsignature}
\\ \opt{\Gvariables}
\\ \opt{\Gequations}
\\ \opt{\Ggoals} \hspace{3ex} \ts{\}}
\end{tabular}\

\\ \Gparameterblock     &::=& \tpar{\ \repswt{\Gsortorfunctionname}{,}\ }
\\ \Gsortorfunctionname &::=& \Gsortname\ $|$ \Gfunctionname

\\ \Gimport             &::=& \ts{import}\ \Gmodulename \opt{\ts{[}\Ginstancename \ts{]}}
\\                      &   & \opt{\ts{<} \reps{\Gextendedparameterblock}\ts{>}}
\\                      &   & \opt{\Gimportblock}
\\ \Gextendedparameterblock &::=& \ts{(}\repswt{\Gnamewithrenaming}{,} \ts{)}
\\                      &$|$& \ts{(}\repswt{\Gsortorfunctionname\ 
                              \ts{bound to}\ \Gsortorfunctionname}{,} 
\\                      &   & \ts{)} \ts{of}\ \Gmodulename \opt{\ts{<} \reps{\Gparameterblock} \ts{>}}
\\ \Gnamewithrenaming   &::=& \Gsortorfunctionname\ \opt{\ts{renamed to}\ \Gsortorfunctionname}
\\                      &$|$& \ts{copy of} \Gsortorfunctionname
\\ \Gimportblock        &::=& \ts{\{}\hspace*{1ex}
\begin{tabular}[t]{ll}
   {[}\ \ts{public:}   & \repswt{\Gnamewithrenaming}{,}\ ]
\\ {[}\ \ts{private:}  & \repswt{\Gnamewithrenaming}{,}\ ] \hspace{3ex}\ts{\}}
\end{tabular}

\\ \Gaddsignature       &::=& \ts{add signature}
\\                      &   & \ts{\{}\hspace*{1ex}
\begin{tabular}[t]{ll}
   {[}\ \ts{parameters:} & \reps{\Gparameterblocksignature}\ {]}
\\ {[}\ \ts{public:}     &       \Gsignature\ {]}
\\ {[}\ \ts{private:}    &       \Gsignature\ {]} \hspace{3ex}\ts{\}}
\end{tabular}
\\ \Gparameterblocksignature &::=& \ts{(}\hspace*{1ex}
\begin{tabular}[t]{l} 
   \Gsignature
\\ {[}\ \ts{conditions}\ \reps{\Gclause}\ ] \hspace{3ex}\ts{)}
\end{tabular}

\\ \Gsignature         &::=& [\ \ts{sorts} \repswt{\Gsortname}{,}\ ]
\\                     &   & 
\begin{tabular}[t]{@{}ll}
   {[}\ \ts{constructors}     & \reps{\Gfunctiondec}\ ]
\\ {[}\ \ts{non-constructors} & \reps{\Gfunctiondec}\ ]
\end{tabular}
\\ \Gfunctiondec       &::=& \repswt{\Gextfunctionname}{,} \ts{:}\ 
                             \repwt{\Gsortname}{\#}
\\                     &   & \ts{->}\ \Gsortname
\\  \Gextfunctionname   &::=& \Gfunctionname\ \opt{\tsym{\_}}
\\                     &$|$& \tsym{\_} \Gfunctionname\ \tsym{\_}

\\ \Gclause            &::=& \ts{[} \Glabel \ts{]}
                             \repwt{\Geq}{,} \ts{-->}\ \repwt{\Geq}{,}
\\ \Geq                &::=& \Gterm\ \opt{\ts{=}\ \Gterm}
\end{tabular}

\begin{tabular}{@{}r@{\hspace{0.5em}}r@{\hspace{0.5em}}l@{}}
\\ \Gterm              &::=& \opt{\Gterm\ \Gfunctionname} \Gprimary
\\ \Gprimary           &::=& \Gfunctionname \opt{\ts{(} \repswt{\Gterm}{,} \ts{)}}
\\                     &$|$& \Gvariablename
\\                     &$|$& \ts{(}\Gterm \ts{)}
\\                     &$|$& \Gfunctionname\ \Gprimary
\\ \Gvariables         &::=& \ts{variables}
\\                     &   & \ts{\{}\hspace*{1ex}
\begin{tabular}[t]{ll}
   {[}\ \opt{\ts{constructors}} & \reps{\Gvariabledec}\ ]
\\ {[}\ \ts{non-constructors}   & \reps{\Gvariabledec}\ ]\hspace{3ex}\ts{\}}
\end{tabular}
\\ \Gvariabledec      &::=& \repswt{\Gvariablename}{,}\ \ts{:}\ \ts{->}\ \Gsortname

\\ \Gequations         &::=& \ts{equations}
\\                     &   & \ts{\{}\ \reps{\Gequation}\ \ts{\}}
\\ \Gequation          &::=& \ts{[} \Glabel \ts{]}
                             \Geq\ \opt{\ts{if}\ \repswt{\Geq}{,}}
\\                     &$|$& \Gmacroequation
\\ \Ggoals             &::=& \ts{goals}
\\                     &   & \ts{\{}\ \reps{\Gclause}\ \ts{\}}
\end{tabular}
\vspace{1cm}

Lexikalisch gelten in der Syntax von \ASFm\ folgende Konventionen:
\begin{itemize}
\item Als Trennzeichen zwischen den einzelnen lexikalischen Token sind
   erlaubt: Leerzeichen, horizontaler Tabulator, carriage return, Zeilen-
   und Seitenvorschub sowie jede Kombination dieser Zeichen.
\item Modulnamen, -k"urzel, Instanzbezeichnungen, Marken- und Sortennamen
   (also \Gmodulename, \Gshortmodulename, \Ginstancename, \Glabel\ und \Gsortname)
   bestehen aus einer beliebigen Folge von Zahlen, Buchstaben, Apostroph \ts{'})
   und Unterstrich (\ts{\_}). Jedoch darf der Unterstrich weder am Anfang,
   noch am Ende eines Namens stehen.
\item In Funktonsnamen (\Gfunctionname), die hier auch die Operatoren aus
   ASF beinhalten, sind zus"atzlich folgende ASCI-Zeichen zul"assig:
   \ts{!}, \ts{\$}, \ts{\%}, \ts{\&}, \ts{$+$}, \ts{$*$}, 
   \ts{;}, \ts{?}, \ts{$\sim$}, \ts{$\backslash$}, \ts{$|$}, \ts{/}, \ts{.}.
\item Die Schl"usselworte \ts{if}, \ts{equation}, \ts{else}, \ts{case},
   \ts{renamed}, \ts{bound}, \ts{sorts} und
   \ts{constructors} stehen als Namen nicht zur
   Verf"ugung.
\end{itemize}
   
Man beachte, da"s in benutzerdefinierten Modulen Sorten-, Funktions- und Markennamen
keinen Bindestrich (\ts{-}) enthalten d"urfen. Andernfalls w"aren Namenskonflikte
zwischen benutzerdefinierten und verdeckten, vom Normalformalgorithmus
erzeugten Namen nicht auszuschlie"sen.

\vfill\pagebreak

\section{Die Normalform-Prozedur}
\label{normalisierung}

Im Mittelpunkt dieses Kapitels steht der Algorithmus, mit dessen Hilfe
beliebige \ASFm -Spezifikationen, bestehend aus einem Topmodul und einer Folge
von direkt, indirekt und implizit importierten Modulen, in flache, importfreie
Spezifikationen umgewandelt werden k"onnen. Besonderen Wert wurde auf die
m"oglichst konsequente Verwendung disambiguierter Namen gelegt. Die
Formalisierung des ASF zugrunde liegenden Algorithmus in \Bergstra, Kapitel
1.3.2, l"a"st hier einige Fragen offen\footnote{
    Beispielsweise ist der zweite Wert eines RENAMING-Tupels ($x$,$y$) im
    allgemeinen kein Element aus SFV. Trotzdem wird ihm in der Beschreibung von
    {\it rename\_visibles} ein Origin zugeordnet.}.
Schwerwiegender ist dagegen das (nicht dokomentierte) Fehlverhalten des
ASF-Normalformalgorithmus bei mehrfachem Import namensgleicher Sorten und
Funktionen mit unterschiedlicher Sichtbarkeit: 
\begin{verbatim}
module exhiddenA
begin
   sorts A
end exhiddenA

module exA
begin
   exports begin sorts A end
end exA

module Certain-Clash
begin
   imports exhiddenA, exA
end Certain-Clash
\end{verbatim}
Da"s die Normalisierung von ASF hier einen Namenskonflikt ausgibt, erscheint
genauso unverst"andlich wie die Tatsache, da"s er sich durch "Anderung
der Importreihenfolge in {\tt Certain-Clash} beheben l"a"st. Zwar wird in der
Beschreibung der Hilfsfunktion {\em combine}\footnote{
   Siehe \Bergstra, Absatz 1.3.2.2.3}
darauf hingewiesen, da"s
verdeckte Namen des ersten Arguments mit sichtbaren Namen des zweiten
Arguments kollidieren k"onnen, ein Hinweis auf die kaum akzeptablen
Auswirkungen auf die Kombination mehrerer zu importierender Module (im
Beispiel {\tt exhiddenA} und {\tt exA}) fehlt jedoch v"ollig.

Der gleiche Fehler f"uhrt zusammen mit dem nur unpr"azise formalisierten
impliziten Renaming sogar dazu, da"s Namenskonflikte zwischen Namen,
die durch die Normalisierung "uberhaupt erst erzeugt wurden,
nicht auszuschlie"sen sind: 
\begin{verbatim}
module exAhiddenA
begin
   exports begin sorts A end
   imports exhiddenA
end exAhiddenA
\end{verbatim}

\vfill\pagebreak

\begin{verbatim}
module exB
begin
   sorts B
end exB

module Possible-Clash
begin
   imports exAhiddenA, exB
end Possible-Clash
\end{verbatim}

Im Zuge der Normalisierung wird zun"achst {\tt exAhiddenA} in Normalform
gebracht. Die dabei notwendige Umbenennung der verdeckt importierten Sorte
{\tt A} erledigt die Funktion {\it rename\_hiddens}. Da sie keine Kenntnis
"uber das Modul {\tt exB} hat, steht einer Ersetzung des Namens
{\tt A} durch {\tt B} aus Sicht des Algorithmus nichts im Wege.
In diesem Fall aber liefert die Normalisierung von {\tt Possible-Clash}
wieder einen Namenskonflikt (gleiche Situation wie oben).

Grund f"ur die Namenskonflikte beider Beispiele ist die Asymetrie der
Hilfsfunktion
{\it combine}, die beim kombinieren zweier Module zwecks
Konfliktvermeidung nur Umbenennungen innerhalb eines Modules vornehmen
darf und sowohl bei der Kombination von Importen untereinander, als auch
mit dem importierenden Modul selbst Verwendung findet.\footnote{
   Siehe \Bergstra, Absatz 1.3.2.3, 4.\ Schritt des Algorithmus}\ \ 
Wir ersetzten {\it combine} durch zwei verschiedene Varianten:
\CombineImports\ kombiniert zwei importierte Module untereinander. Ihre
Argumente (zwei Module in Normalform) werden gleich behandelt, somit ist
die Reihenfolge der Importanweisungen belanglos. \CombineWithImports\
entspricht in etwa {\it combine} aus ASF --- sie kombiniert das importierende
Modul mit der Kombination aller Importe.

\subsection{Datenstrukturen}
Bevor der Normalformalgorithmus vorgestellt werden kann, m"ussen zun"achst
die Daten erl"autert werden, auf denen er operiert. Als Basistyp
beschr"anken wir uns auf Zeichenketten. Sie werden
in Mengen- und Strukturtypen, die wir als Tupel mit unterschiedlichen
Komponententypen darstellen werden, zu komplexeren Datenstrukturen zusammen
gesetzt. Funktionen werden als Mengen repr"asentiert:
$f = \{(x,y) \quad | \quad y = f(x)\}$. \pot($X$) bezeichnet die Potenzmenge
von $X$, also die Menge aller Teilmengen.

Ziel der Normalisierung ist die Transformation einer \ASFm-Spezifikation,
bestehend aus einzelnen \ASFm-Modulen, in eine neue importfreie \ASFm-Spezifikation.
Neben den Typen \ASFMODULE\ und \ASFSPEC\ werden f"ur die Eingabeschnittstelle
der Normalisierungsprozedur auch Informationen "uber bereits gef"uhrte Beweise
ben"otigt. Sie werden im Typ \PROVEDB\ zusammengefa"st.
\begin{itemize}
\item \ASFMODULE\ ist die Menge aller Zeichenfolgen, die syntaktisch korrekte
   \ASFm-Module darstellen.
\item \ASFSPEC\ := \ASFMODULE\ $\times$ \pot(\ASFMODULE) \\
   \ASFm-Spezifikationen bestehen aus einem Topmodul und einer Menge
   von Modulen, die mindestens alle vom Topmodul direkt, indirekt
   und implizit importierten Module enthalten mu"s.
\item \PROVEDB\ ist eine nicht n"aher konkretisierte Wissensbasis f"ur gelungene
   Beweise. Mit ihrer Hilfe wird die G"ultigkeit von semantischen Bedingungen
   f"ur Parameterbindungen gepr"uft.
\end{itemize}

Allerdings eignet sich die Repr"asentation eines \ASFm-Moduls als unstrukturierte
Zeichenkette kaum zur ad"aquaten Beschreibung der f"ur die Transformation
notwendigen Operationen (z.~B.\ Kombination mehrerer Module). Wir
f"uhren daher einen strukturierten Datentyp \MODULE\ ein, der es erm"oglicht,
auf einzelne Teile eines repr"asentierten Moduls (z.~B.\ auf die Importbefehle)
direkt zuzugreifen. Die kleinsten logischen Einheiten eines Moduls bestehen
aus Namen, die in Abh"angigkeit vom Kontext ihres Auftretens als Modulnamen
oder -k"urzel, als Marken, Instanzbezeichnungen, Variablen-, Sorten- oder
als Funktionsnamen dienen. Unter den Sorten- und Funktionsnamen besitzen
wiederum in einer Parametersignatur definierten Namen einen Sonderstatus,
sie hei"sen Sorten- und Funktionsparameter.
Einige Namen werden w"ahrend der Normalisierung ver"andert oder zur Ver"anderung
anderer Namen gebraucht.
Um die Zahl der Namenstypen m"oglichst "uberschaubar zu halten, fassen wir
Namen, auf denen die gleichen Operationen ausgef"uhrt werden, gruppenweise
zusammen:
\begin{itemize}
\item \MODULENAME\ ist Menge aller Modulnamen.
\item \SHORTMODULENAME\ ist Menge aller abgek"urzten Modulnamen. Sie enth"alt
   alle Modulk"urzel sowie die Namen der Module, f"ur die kein
   K"urzel angegeben worden ist. Unter dem  ``abgek"urzten Namen eines Moduls''
   verstehen wir das im Modul vereinbarte K"urzel, oder (falls nicht
   vorhanden) den Modulnamen selbst.
\item \INSTNAME\ ist Menge aller Instanzbezeichnungen.
\item \USERNAME\ ist die Menge aller dem Spezifizierer zur Verf"ugung
   stehenden Namen f"ur Parameter, Sorten, Funktionen, Variablen und Marken.
\end{itemize}

Neben den vom Spezifizierer erzeugten Namen generiert der Normalformalgorithmus
auch selbstst"andig Namen, die sich (im Gegensatz zu ASF) aus den vom 
Spezifizierer vorgegebenen Namen und K"urzeln zusammensetzen.
Das modulare Konzept aus \ASFm\ basiert wesentlich auf der Zuordnung von
Namen zu Namensr"aumen. Eine Namensraumbezeichnung
besteht aus einem Modulnamen und einer gegebenenfalls leeren Liste von
Instanzbezeichnungen, welche Auskunft dar"uber gibt, um welche Version des
Namensraumes es sich handelt. Auch Namensraumbezeichnungen k"onnen mit Hilfe der
Modulk"urzel abgek"urzt werden.
\begin{itemize}
\item \MODULEINSTNAME\ enth"alt alle Namensraumbezeichnungen. Syntaktisch
   kann \MODULEINSTNAME\ durch eine Grammatik-Produktionsregel wie folgt
   beschrieben werden:

   \begin{tabular}{lrl}
   \MODULEINSTNAME &::=& \MODULENAME \\ 
                   &$|$& \MODULENAME \ts{[}\repswt{\INSTNAME}{,}\ts{]}
   \end{tabular}

\pagebreak

\item \SHORTMODULEINSTNAME\ enth"alt alle abgek"urzten Namensraumbezeichnungen.

   \begin{tabular}{l@{\ }r@{\ }l}
   \SHORTMODULEINSTNAME &::=& \SHORTMODULENAME \\
                         &$|$& \SHORTMODULENAME \ts{[}\repswt{\INSTNAME}{,}\ts{]}
   \end{tabular}
\end{itemize}

Wird beim Import ein Name verdeckt, so ersetzt der \ASFm-Normalforalgorithmus
den Namen durch einen neuen, innerhalb der gesamten Spezifikation eindeutigen Namen,
indem er dem alten Namen eine abgek"urzte Namensraumbezeichnung gefolgt
von einem Bindestrich voranstellt. Der so erzeugte Name ist kein \USERNAME,
kann also mit keinem vom Spezifizierer eingef"uhrten Namen in Konflikt geraten.
\begin{itemize}
\item \SPECNAME\ umfa"st alle Parameter-, Sorten-, Funktions-, Variablen- und
   Markennamen, die nach der Normalisierung in der Spezifikation auftreten
   k"onnen. Zur Charakterisierung der Syntax geben wir wieder eine Produktionsregel
   an:
   
   \SPECNAME\quad ::=\quad \USERNAME\ $\quad |\quad$ \SHORTMODULEINSTNAME\ts{-}\USERNAME
\end{itemize}

\ASFm\ gestattet es, Funktionsnamen zu "uberladen. Um eine spezielle Funktion
identifizieren zu k"onnen ist deshalb die Kenntnis der Argumentsorten
erforderlich. Dies f"uhrt uns zu disambiguierten Namen:
\begin{itemize}
\item \SORTVECTOR\ ist eine Menge von Listen, deren
   Komponenten Sortennamen ($\in$ \SPECNAME) sind.
\item \DISAMBSPECNAME\ := \SPECNAME\ $\times$ \SORTVECTOR \\
   umfa"st die Menge der disamiguierten Namen. Disambiguierte Namen
   sind Tupel (\name, \sortv). Falls \name\ ein Sorten-, Marken-,
   Variablen oder Konstantenname ist, ist der Sortenvektor \sortv\ leer.
   Handelt es sich dagegen um einen Funktionsnamen (bzw. Funktionsparameter)
   enth"alt er die Namen der Argumentsorten.
\end{itemize}

Die Datenstruktur f"ur die Importe nimmt alle, aus den Importkonstrukten
hervorgehenden Informationen auf, gruppiert sie nach den Erfordernissen
der sequenziellen Auswertung jedoch neu:
\begin{itemize}
\item \VISIBILITYFUNC\ := \USERNAME\ $\longrightarrow$ \{\tsym{public}, \tsym{private}\} \\
   Sie bestimmt die Sichtbarkeit von Signaturnamen beim Import eines
   Moduls. Da \ASFm\ verlangt, da"s alle nach dem Import sichtbaren Namen
   im Importkonstrukt aufgef"uhrt werden m"ussen, kann sie direkt aus der
   Importanweisung bestimmt werden. Namen aus dem importierten Modul, die
   keine Parameter sind und denen
   keine Sichtbarkeitsstufe $\in$ \{\tsym{public}, \tsym{private}\} zugewiesen
   wird, werden beim Import verdeckt.
\item \RENAMINGFUNC\ := \USERNAME\ $\longrightarrow$ \SPECNAME \\
   Werden Sorten- und Funktionsnamen durch eine Funktion aus \RENAMINGFUNC\
   auf andere Sorten- und Funktionsnamen
   abgebildet, so beschreibt diese Funktion explizites Renaming. Handelt es sich
   dagegen im Definitionsbereich ausschlie"slich um Parameter, so kann mit ihrer
   Hilfe eine Parametertupelbindung beschrieben werden:
\item \BINDINGBLOCK\ := \RENAMINGFUNC\ $\times$ \MODULENAME\\
   Tupel (\binding, \modn) dieses Typs repr"asentieren einen Block des
   Importbefehls, 
\pagebreak
   der "uber \binding\ das Binden von Parametern
   eines Tupels an Signaturnamen eines Moduls namens \modn\ beschreibt.
\item \IMPORT\ := \MODULENAME\ $\times$ \INSTNAME\ $\times$ \VISIBILITYFUNC\ $\times$
   \RENAMINGFUNC\ $\times$ \pot(\BINDINGBLOCK)\\
   Elemente dieses Typs repr"asentieren Importbefehle. Es handelt sich hier
   also um eine strukturierte Repr"asentation einer Zeichenkette, die den
   Import in \ASFm\ beschreibt. Wird ein benutzender Import dargestellt, ist
   die zweite Komponente leer.

   Zur Veranschaulichung sei als Beispiel folgender Importbefehl gegeben:
\begin{verbatim}
   import Sequences[NSeq] <(ITEMpar bound to NAT) of Naturals>
   {  public:  SEQ renamed to NSEQ
      private: nil renamed to nnil, cons  }
\end{verbatim}
   Wir erhalten folgende Tupeldarstellung:
   \begin{tabbing}
   \hspace*{4ex}(\ \= \tsym{Sequences}, \tsym{NSeq},\\
             \> \{(\tsym{SEQ}, \tsym{public}), (\tsym{nil}, \tsym{private}),
                  (\tsym{cons}, \tsym{private})\},\\
             \> \{(\tsym{SEQ}, \tsym{NSEQ}), (\tsym{nil}, \tsym{nnil})\},\\
             \> \{(\{(\tsym{ITEMpar}, \tsym{NAT})\}, \tsym{Naturals})\}\ )
   \end{tabbing}
\end{itemize}

W"ahrend Importkonstrukte nur Teil einer nicht normalisierten Spezifikation
sind, ist das Auftreten von Signaturen, Variablenvereinbarungen, Klauseln
und Gleichungen unabh"angig vom Grad der Normalisierung:
\begin{itemize}
\item \SIG\ := \pot(\SPECNAME) $\times$
              \pot(\DISAMBSPECNAME\ $\times$ \SPECNAME)$^2$ \\
   Dieser Datentyp repr"asentiert eine Teilsignatur eines Moduls.
   Teilsignaturen bestehen aus einer Menge von Sortennamen und je einer
   Mengen von Deklarationen f"ur Konstruktoren und Non-Konstruktoren.
   Jede Deklaration wird durch einen disambiguierten Namen
   (Funktionsname + Argumentsorten) und die zugeh"origen Zielsorte
   repr"asentiert.
\item \VARSORTFUNC\ := \SPECNAME\ $\longrightarrow$ \SPECNAME \\
   Die Variablenvereinbarung eines Moduls beschreibt eine Funktion, die jedem
   Variablennamen eine Sorte zuweist.
\item \CLAUSE
\item \EQUATION
\end{itemize}

Mit Hilfe der so definierten Strukturen kann nun ein \ASFm-Modul wie folgt
als 9-Tupel repr"asentiert werden:
\begin{itemize}
\item \MODULE\ := \MODULENAME\
        $\times$ \pot(\IMPORT)
        $\times$ \pot(SIG $\times$ \pot(\CLAUSE))
        $\times$ \SIG$^2$
        $\times$ \VARSORTFUNC$^2$
        $\times$ \pot(\EQUATION)
        $\times$ \pot(\CLAUSE) \\
   Dem Modulnamen
   folgen die Importe, eine Menge von Parametersignaturen (mit Bedingungsklauseln),
   die exportierte (\public) und die nur innerhalb des Moduls sichtbare
   ({\tt private}) Signatur, zwei Funktionen, die den Konstruktor- bzw.\
   Non-Konstruktor-Variablen ihre jeweilige Sorte zuweisen, eine Menge von
   spezifizierenden Gleichungen und schlie"slich eine Menge von Beweiszielen.
\end{itemize}

Wir k"onnen nun pr"azisieren, was wir im Folgenden unter der komponentenweisen
Vereinigung einer Menge von Modulen verstehen werden:

\noindent Sei $\{\module_i\quad |\quad i \in A\}$ eine Menge von Modulen
und es gelte f"ur $i \in A$
\begin{tabbing}
\module$_i$ = \=(\ \= \modiname$_i$, \imports$_i$, \parameters$_i$,
\\            \> \> (\sorts$_{\mbox{\tiny pub},i}$, \const$_{\mbox{\tiny pub},i}$,
                     \nonconst$_{\mbox{\tiny pub},i}$),
\\            \> \> (\sorts$_{\mbox{\tiny pri},i}$, \const$_{\mbox{\tiny pri},i}$,
                     \nonconst$_{\mbox{\tiny pri},i}$),
\\            \> \> \varsortfunc$_{\mbox{\tiny     const},i}$,
                    \varsortfunc$_{\mbox{\tiny non-const},i}$,
\\            \> \> \equations$_i$, \goals$_i$\ ).
\\ \\
$\Cunion_{i \in A} \module_i := $
\\         \>(\>$\emptyset, \Union_{i \in A} \imports_i, \Union_{i \in A} \parameters_i,$
\\         \> \>$(\Union_{i \in A} \sorts_{\mbox{\tiny pub},i},
                  \Union_{i \in A} \const_{\mbox{\tiny pub},i},
                  \Union_{i \in A} \nonconst_{\mbox{\tiny pub},i}),$
\\         \> \>$(\Union_{i \in A} \sorts_{\mbox{\tiny pri},i},
                  \Union_{i \in A} \const_{\mbox{\tiny pri},i},
                  \Union_{i \in A} \nonconst_{\mbox{\tiny pri},i}),$
\\         \> \>$\Union_{i \in A} \varsortfunc_{\mbox{\tiny     const},i}$,
                $\Union_{i \in A} \varsortfunc_{\mbox{\tiny non-const},i}$,

\\         \> \>$\Union_{i \in A} \equations_i,
                 \Union_{i \in A} \goals_i\ )$
\end{tabbing}

Im Zuge der Importelemination geht Information "uber den hierachischen
Aufbau der Spezifikation und die Herkunft der Signaturnamen verloren.
Um dieses Wissen der Normalisierungsprozedur zug"anglich zu machen, werden
eine Origin- und eine Dependenzfunktion eingef"uhrt:
\label{origin}
\begin{itemize}

\item \ORIGIN\ := \USERNAME\
         $\times$ \MODULEINSTNAME\
         $\times$ \{\lvs, \tsym{function}\}
         $\times$ \{\tsym{parameter}, \tsym{public}, \tsym{private}, \tsym{hidden}\}
\\ Im Prinzip w"urden Origins, die Auskunft "uber den Namensraum eines
   Bezeichners geben, ausreichen, um die angestrebte Semantik zu realisieren.
   Zur Formulierung des Normalformalgorithmus erweist es sich jedoch als
   zweckm"a"sig, weitere redundante Informationen, beispielsweise aus der
   Signatur aufzunehmen. In \ASFm\ werden Origins als Viertupel erkl"art.
   Die vier Komponenten eines Origins (\uname, {\it defmodiname}, \symboltype, \visibility)
   sind wie folgt definiert:
   \begin{itemize}
   \item \uname\ enth"alt den Namen ($\in$ \USERNAME), der vom Spezifizierer f"ur
      die spezifizierte Sorte, Funktion, Variable oder Marke (im folgenden als das
      spezifizierte Objekt bezeichnet) eingef"uhrt wurde. Explizites Renaming
      ver"andert nicht nur den Namen selbst sondern auch den Eintrag \uname\
      des zugeordneten Origins.
   \item \defmodiname\ gibt Auskunft "uber den Namensraum, dem der Name
      angeh"ort.
   \item \symboltype: Namens"anderungen (sowohl implizites als auch explizites
      Renaming) werden im Normalformalgorithmus von \ASFm\ in zwei Stufen
      durchgef"uhrt. Zun"achst werden alle Sorten, Variablen und Marken
      umbenannt (f"ur zugeh"orige Origins gilt \symboltype\ $\in$ \{\lvs\},
      danach folgt die Umbenennung der Funk\-tionen. Diese Reihenfolge
      operationalisiert die durch Overloading bedingte rekursive Struktur
      der Identifikationsregel aus \ASFm.
   \item \visibility: F"ur Sorten und Funktionen gibt es drei Sichtbarkeitsstufen:
      \begin{itemize}
      \item \tsym{public}: innerhalb des Moduls sichtbar und exportf"ahig.
      \item \tsym{private}: innerhalb des Moduls sichtbar, jedoch nicht
         exportf"ahig.
      \item \tsym{hidden}: innerhalb des Moduls verdeckt und nat"urlich auch
         nicht exportf"ahig.
      \end{itemize}
      Marken und Variablen gelten innerhalb des Moduls, in dem sie definiert
      werden als \tsym{private}, beim Import des Moduls werden sie verdeckt.
      Parameter geh"oren dem Sonderstatus \tsym{parameter} an und k"onnen
      nicht verdeckt werden.
   \end{itemize}
\item \ORIGINFUNC\ := \DISAMBSPECNAME\ $\longrightarrow$ \ORIGIN \\
   Funktionen dieses Typs weisen (disambiguierten) Namen aus der Spezifikation
   Origins zu. 
\item \DEPENDENCYFUNC\ := \MODULEINSTNAME\ $\longrightarrow$
                         \pot(\MODULEINSTNAME) \\
   Dependenzfunktionen beschreiben die Abh"angigkeiten zwischen Namensr"aumen
   einer Spezifikation. Jeder Namensraumbezeichnung aus dem Definitonsbereich
   wird die Menge der Bezeichnungen aller abh"angigen Namensr"aume
   zugewiesen.
\end{itemize}

Der rekursive Normalformalgorithmus operiert auf einer Datenstruktur, die
wir ``general forms'' (\GFORM) nennen:
\begin{itemize}
\item \GFORM\ := \MODULE\ $\times$ \ORIGINFUNC\ $\times$ \DEPENDENCYFUNC \\
   General forms bestehen aus der (internen) Repr"asentation eines Moduls,
   einer Originfunktion und einer Dependenzfunktion. Origin- und 
   Dependenzfunktion k"onnen partiell sein in dem Sinn, da"s Namen aus zu
   importierenden Modulen zun"achst unber"ucksichtigt bleiben.
\item \NFORM\ $\subseteq$ \GFORM \\
   Als  Normalformen bezeichenen wir alle general forms, die ein
   importfreies Modul, eine auf den im Modul vorkommenden Namen
   ($\in$ \DISAMBSPECNAME) totale Originfunktion und eine auf den
   Bezeichnungen aller im Modul enthaltenen Namensr"aume totale
   Dependenzfunktion beinhalten. Eine Normalform repr"asentiert also nicht nur
   ein normalisiertes Modul sondern auch den Bauplan der Spezifikation, der
   das Modul seine Erzeugung verdankt.
\end{itemize}

Schlie"slich wird f"ur die Normalisierungsprozedur noch eine Funktion
ben"otigt, die Namensumbenennungen einzelner ggf.\ "uberladener Signaturnamen
eindeutig beschreibt.
Da sich das Exportverhalten "uberladener Funktionsnamen unterscheiden
kann, ist eine Differenzierung nach den Argumentsorten erforderlich:
\begin{itemize}
\item \DISAMBRENAMINGFUNC\ := \DISAMBSPECNAME\ $\longrightarrow$ \SPECNAME \\
   Dieser Datentyp beschreibt Umbenennungen, die aufgrund von "Anderungen der
   Sichtbarkeit einzelner Namen beim Import erforderlich werden. Es ist zu
   beachten, da"s die Durchf"uhrung von Funktionsumbenennungen dieser Art
   in Gleichungen
   das Disambiguieren der Funktionssymbole jedes einzelnen Terms
   erfordert. Hierzu wird die Signatur des Moduls gebraucht.
\end{itemize}

\subsection{Der Algorithmus}

\subsubsection{Globale Hilfsfunktionen f"ur Sichtbarkeits"anderungen}
Das dynamische Verdeckungsprinzip von \ASFm\ erfordert bei der Kombination
verschiedener Module zahlreiche Signaturnamensumbenennungen.
Jede Namens"anderung zieht im allgemeinen Ver"anderungen in fast allen
Teilen des Moduls und der Originfunktion nach sich. 
Der Normalformalgorithmus erledigt dies in zwei Schritten.
Zuerst wird die 4.\ Komponente \visibility\ der Origins aller Namen auf
die Sichtbarkeitsstufe gesetzt, die die jeweiligen Namen zuk"unftig
haben sollen.
Die sich daraus ergebenden Umbenennungen im Modul und dem Defi\-ni\-tions\-bereich
der Originfunktion, sowie die Neuordnung der Signatur werden dann von der
Funktion \MakeConsistent\ erledigt.

Die Vorgehensweise des hier vorgestellten Algorithmus nutzt die in der
Originfunktion enthaltene Redundanz aus: Jedes Origin \origin\ beschreibt
den ihm zugeordneten (nicht disambiguierten) Namen 
\GetSpecName(\origin) eindeutig.

\begin{algorithmus}{\GetSpecName}%
{\ORIGIN}{\SPECNAME}%
{  \GetSpecName((\uname, \defmodiname, *, \visibility)) berechnet aus der
   ersten, zweiten und vierten Komponente eines Origins den zugeordneten 
   (nicht disambiguierten) Namen ($\in$ \SPECNAME).  }%
{R"uckgabewert: \specname}
   falls \visibility\ $\in$ \{\tsym{parameter}, \tsym{public}, \tsym{private}\}
   \begin{block}
      Setze \specname\ := \uname.
   \end{block}

   falls \visibility\ = \tsym{hidden}
   \begin{block}
      Setze {\em shortmodiname} gleich dem abgek"urzten Modulinstanznamen
      von \defmodiname.
      
      \specname\ := {\em shortmodiname}\tsym{-}\uname
   \end{block}
\end{algorithmus}

\pagebreak

Manipulationen der Sichtbarkeitskomponente im Wertebereich einer
Originfunktion f"uhren im allgemeinen dazu,
da"s die Namen des Defi\-ni\-tions\-bereichs nicht mehr zu den zugeordneten
Origins passen. Sei ((\name$_i$, \sortv$_i$), \origin$_i$) Element einer
Originfunktion, dann entspricht \GetSpecName(\origin$_i$) dem ``Sollwert''
von \name$_i$. Zur Namensaktualisierung im Modul und im Defi\-ni\-tions\-bereich
der Originfunktion dient eine ``Istwert-Sollwert''-Liste, die von
\GetRenaming\ erzeugt wird:

\begin{algorithmus}{\GetRenaming}%
{\ORIGINFUNC\ \times\ \pot(\{\lvs, \tsym{function}\})}%
{\DISAMBRENAMINGFUNC}%
{  \GetRenaming(\originf, \symboltypes) errechnet ein \renaming\
   f"ur disambiguierte Namen.
   Mit dessen Hilfe kann innerhalb der Originfunktion sowie eines
   Normal\-form\-moduls ein konsistenter Zustand hergestellt werden.
   \symboltypes\ bestimmt, welche Namenstypen in das \renaming\ aufgenommen
   werden sollen.  }%
{R"uckgabewert: \renaming}

   $\renaming\ :=\ \{\ 
   \begin{array}[t]{l}
      ((\name, \sortv), \name')\quad| \\
      (u, n, \symboltype, v)=\originf((\name, \sortv))\quad\wedge \\
      \symboltype \in \symboltypes\quad \wedge \\
      \name' = \GetSpecName((u, n, \symboltype, v))\quad \wedge \\
      \name' \neq \name\ \}
   \end{array}$

 %%%     \end{block}
\end{algorithmus}

Bei der Beschreibung von Umbenennungen "uberladbarer Signaturnamen ist zu
ber"ucksichtigen, da"s Sortenumbenennungen auch die Sortenvektoren der
umzubenennenden (dis\-ambiguierten) Funktionsnamen beeinflussen. Aus Gr"unden
der "Ubersichtlichkeit verzichten wir auf ``simultanes'' Umbenennen von
Sorten und Funktionen, \MakeConsistent\ behandelt Sorten und Funktionen
nacheinander:

\begin{algorithmus}{\MakeConsistent}%
{\MODULE\ \times\ \ORIGINFUNC}{\MODULE\ \times\ \ORIGINFUNC}%
{  \MakeConsistent(\module, \originfunc) erh"alt ein (normalisiertes)
   Modul \module\ und eine Originfunktion \originfunc\, deren Wertebereich
   zwecks Durchf"uhrung von Verdeckung oder "Anderung des
   Exportverhaltens von Namen
   manipuliert wurde. Unter Zuhilfename der Funktionen \GetSpecName\ und
   \GetRenaming\ berechnet sie ein konsistentes Tupel
   (\module$'''$, \originfunc$''$).
   Durchgef"uhrt werden Umbenennung von Namen ($\in$ \SPECNAME) in \module\
   und im Defi\-ni\-tions\-bereich von \originfunc, sowie der Austausch von
   Sortennamen und Funktionsdeklarationen zwischen der \public- und
   \private-Signatur.  }%
{R"uckgabewert: (\module$'''$, \originfunc$''$)}
   \pagebreak

   \renaming\ := \GetRenaming(\originfunc, \{\lvs\})

   Berechne \module$'$ durch Ersetzen der Sorten-, Variablen- und
   Markennamen in \module\ nach Ma"sgabe von \renaming.
   
   Berechne \originfunc$'$ durch Ersetzen der Sorten-, Variablen- und
   Markennamen im Defi\-ni\-tions\-bereich von \originfunc\ nach Ma"sgabe von \renaming.
   Betroffen sind insbesondere auch die Sortenvektoren der disambiguierten
   Funktionsnamen.
  
   \renaming$'$ := \GetRenaming(\originfunc$'$, \{\tsym{function}\})

   Berechne \module$''$ und \originfunc$''$ durch Ersetzen der Funktionsnamen
   in \module$'$ und im Defi\-ni\-tions\-bereich der Originfunktion \originfunc$'$
   nach Ma"sgabe von \renaming$'$.
   
   \module$'''$ entsteht aus \module$''$ durch Aktualisierung der \public-
   und \private-Signatur. Namen mit Sichtbarkeitsstufe \tsym{private} oder
   \tsym{hidden} sind nicht exportf"ahig und geh"oren in die \private-Signatur.
   Solche mit Sichtbarkeitsstufe \tsym{public} hingegen geh"oren in die
   \public-Signatur. Parameter bleiben wo sie sind, n"amlich in der 
   {\tt parameter}-Signatur.
\end{algorithmus}
\vspace{3ex}

\subsubsection{Kombination von Modulen}

Der Import sowie das Binden von Parametern an Signaturnamen eines
Moduls f"uhrt bei der Normalisierung dazu, da"s mehrere general forms
zu einer neuen general form zusammengefa"st werden m"ussen.
Diese Aufgabe erledigen die drei Funktionen  \CombineImports,
\CombineWithImports\ und \CombineWithActModule. 
Als Hilfsfunktion greifen sie auf \CombineDependencies\
und \AdaptVisibility, welche die notwendigen Sichtbarkeitsanpassungen
vornimmt, zu.

\AdaptVisibility\ kann als Identifikationsregel gelesen werden, die beim
Kombinieren mehrerer Module festlegt,
wann (disambiguierte) Namen miteinander identifiziert werden d"urfen
und unter welchen Umst"anden es zu Namenskonflikten kommt. Wegen der
Overloading-F"ahigkeit ist hierbei von Wichtigkeit,
da"s zuerst alle Sortenidentifikationen vorgenommen werden 
(Aufruf von \AdaptVisibility\ mit \symboltypes\ := \{\tsym{sort}\}). Erst
danach k"onnen die Funktionsidentifikationen korrekt durchgef"uhrt werden
(\symboltypes\ := \{\tsym{function}\}). Die Zweistufigkeit reduziert
den Sortenvektortest auf syntaktische Gleichheit. Andernfalls m"u"sten beim
Test auf Identifizierbareit die Argumentsorten der (disambiguierten)
Funk\-tions\-namen komponentenweise (rekursiv) auf Identifizierbarkeit gepr"uft
werden. Analog zur sogenannten ``Originrule'' aus ASF
k"onnen wir die \ASFm\ zugrundeliegende Identifikationsregel folgenderma"sen
beschreiben:

\pagebreak

\paragraph{Identifikationsregel:}
Die (disambiguierten) Namen (\name$_1$, \sortv$_1$) und (\name$_2$, \sortv$_2$)
aus zwei zu kombinierenden Modulen sind genau dann zu identifizieren, wenn
\begin{itemize}
\item die ihnen zugeordneten Origins in den ersten drei Komponenten 
   "ubereinstimmen,
\item die 4. Komponenten der Origins "ubereinstimmen oder eine 4. Komponente
   den Wert \tsym{hidden}, die andere 4. Komponente dagegen \tsym{private}
   oder \tsym{public} enth"alt und
\item die in \sortv$_1$ und \sortv$_2$ enthaltenen Argumentsorten
   (nur bei Funktionsnamen relevant) miteinander identifiziert werden k"onnen.
\end{itemize}
Man beachte, da"s diese Definition nicht eigentlich rekursiv ist, da der R"uckbezug nicht 
wiederum selbst r"uckbez"uglich ist.

Zwischen den disambiguierten Namen $(\name_1, \sortv_1)$ und $(\name_2, \sortv_2)$
aus zwei zu kombinierenden Modulen kommt es genau dann zum Konflikt, wenn
\begin{itemize}
\item sie nicht miteinander identifiziert werden k"onnen, obwohl
\item $\name_1$ mit $\name_2$ "ubereinstimmt und
\item die in \sortv$_1$ und \sortv$_2$ enthaltenen Argumentsorten
   miteinander identifiziert werden k"onnen.
\end{itemize}

Wir f"uhren noch eine Sprechweise ein, die sich bei der Behandlung von Parameterbindungen
als n"utzlich erweisen wird.

\label{approx}
Seien 
$\originf_1$ und $\originf_2$ zwei Originfunktionen.
Sei $(\name_1, \sortv_1)$ aus dem Defi\-ni\-tions\-bereich von $\originf_1$
und $(\name_2, \sortv_2)$ aus dem Defi\-ni\-tions\-bereich von $\originf_2$.
Wir definieren:
{\em
$(\name_1, \sortv_1)$
referenziert bez"uglich $\originf_1$ dasselbe Objekt wie
$(\name_2, \sortv_2)$
bez"uglich $\originf_2$}\/
(im Zeichen: $(\name_1, \sortv_1)/\originf_1\approx(\name_2, \sortv_2)/\originf_2$)
genau dann, wenn
\begin{itemize}
\item die ihnen zugeordneten Origins in den ersten drei Komponenten 
   "ubereinstimmen und
\item die in den Komponenten von \sortv$_1$ und \sortv$_2$ enthaltenen
   Argumentsorten (nur bei Funktionsnamen relevant)
   jeweils dasselbe Signaturobjekt referenzieren.
\end{itemize}

Die Identifikationsregel aus \ASFm\ identifiziert also Namen, die das gleiche
Signaturobjekt referenzieren und deren Exportverhalten in den Importbefehlen
nicht widerspr"uchlich festgelegt wird.

\vspace{3ex}
\begin{algorithmus}{\AdaptVisibility}%
{\pot(\NFORM)\ \times\ \pot(\{\lvs, \tsym{function}\})}%
{\pot(\NFORM)}%
{  \AdaptVisibility(\normalforms, \symboltypes)
   sorgt f"ur die Angleichung der Sichtbarkeit von Sorten-
   (\tsym{sort} $\in$ \symboltypes)
   und Funktionsnamen (\tsym{function} $\in$ \symboltypes)
   aus verschiedenen (NF-) Modulen.
   Gleichzeitige \public- und \private-Importe eines Namens weisen auf einen
   Spezifikationsfehler hin, weil ein Name entweder exportierbar
   oder nicht-exportierbar sein kann aber nicht beides gleichzeitig.  }
{R"uckgabewert: $\{(\mod'_i, \originf'_i, \depf'_i)\quad |\quad 1\leq i\leq p\}$}
 
   Sei $\{(\mod_i, \originf_i, \depf_i)\ |\ 1\leq i\leq p\}\ =\ \normalforms$

   F"ur $i:=1$ bis $p$ wiederhole
   \begin{block}
      F"ur $j:=i+1$ bis $p$ wiederhole
      \begin{block}
         F"ur alle ((\name$_i$, \sortv$_i$),
                    (\uname$_i$, \modiname$_i$, \symboltype$_i$, \visibility$_i$))
                   $\in$ \originf$_i$ wiederhole
         \begin{block}
            F"ur alle ((\name$_j$, \sortv$_j$),
                       (\uname$_j$, \modiname$_j$, \symboltype$_j$, \visibility$_j$))
                      $\in$ \originf$_j$ wiederhole
         \end{block}
      \end{block}
   \end{block}

            \begin{einruecken}{12ex}
               /* Alle Origins aller "ubergebenen Originfunktionen \originf$_i$
                  werden mit allen Origins aller anderen "ubergebenen
                  Originfunktionen \originf$_j$ verglichen. */
     
               Falls (\symboltype$_i$ $\in$ \symboltypes) und \\
               \sortv$_i$ = \sortv$_j$ und \uname$_i$ = \uname$_j$
         
               \begin{block}
                  Falls \modiname$_i$   = \modiname$_j$
                  \begin{block}
                     Falls $\symboltype_i \neq \symboltype_j$
                     \begin{block}
                        SPEZIFIKATIONSFEHLER
                     \end{block}
                        
                     /* Beide disambiguierten Namen verdanken ihre Existenz
                     derselben Definition */

                     Falls (\visibility$_i$ = \tsym{hidden} und
                            \visibility$_j$ $\in$ \{\tsym{public}, \tsym{private}\})
                     \begin{block}
                        Setze (mit "Anderung von \originf$_i$)
                        \visibility$_i$ := \visibility$_j$ \\
                     \end{block}
      
                     Sonst falls (\visibility$_j$ = \tsym{hidden} und
                           \visibility$_i$ $\in$ \{\tsym{public}, \tsym{private}\})
                     \begin{block}
                        Setze (mit "Anderung von \originf$_j$)
                        \visibility$_j$ := \visibility$_i$ \\
                     \end{block}
                     Sonst falls \visibility$_i$ $\neq$ \visibility$_j$
                     \begin{block}
                        EXPORTIERBARKEITS-KONFLIKT
                     \end{block}
                  \end{block}

                  Sonst falls \name$_i$ = \name$_j$
                  \begin{block}
                     /* Der disambiguierte Name $(\name_i, sortv_i)$ tritt
                        in beiden Normalformen mit unterschiedlicher
                        Bedeutung auf. */

                     NAMENSKONFLIKT
                  \end{block}
               \end{block}
            \end{einruecken}

   F"ur alle $i \in \{1,\ldots, p\}$
   \begin{block}
      $(\mod'_i, \originf'_i, \depf'_i) := \MakeConsistent(\mod_i, \originf_i, \depf_i)$
   \end{block}
\end{algorithmus}

\pagebreak

\vspace{3ex}

\begin{algorithmus}{\CombineDependencies}%
{\pot(\DEPENDENCYFUNC)}{\DEPENDENCYFUNC}%
{  $\CombineDependencies(\{\depf_j\ |\ j\in A\})$
   erzeugt aus den Dependenzfunktionen mehrerer zu kombinierender
   general forms eine neue Dependenzfunktion $\depf'$.  }%
{R"uckgabewert: $\depf'$}
  /* Siehe Seite \pageref{CombineDependencies}. */
\end{algorithmus}

Importe werden in \ASFm\ eleminiert, indem zun"achst die Normalformen der
importierten Module berechnet werden. Diese werden nach Ma"sgabe der
Importbefehle modifiziert und instanziiert (siehe dazu den folgenden Abschnitt
\ref{Modifikationen in Importbefehlen})
und anschlie"send untereinander kombiniert. Daf"ur zust"andig ist die Funktion
\CombineImports:

\begin{algorithmus}{\CombineImports}%
{\pot(\NFORM)}{\NFORM}%
{  \CombineImports(\normalforms) kombiniert mehrere Normalformen.  }%
{R"uckgabewert: $(\mod', \originf', \depf')$}

   $\normalforms'\  := \AdaptVisibility(\normalforms, \{\lvs\})$

   $\normalforms''\ := \AdaptVisibility(\normalforms', \{\tsym{function}\})$

   Sei $\{(\mod_i, \originf_i, depf_i)\ |\ i\in A\} = \normalforms''$.

   $\mod'\ := \Cunion_{i\in A} \mod_i$

   $\originf'\ := \Union_{i\in A} \originf_i$

   Falls $\originf'$ keine Funktion
   \begin{block}
      NAMECLASH
   \end{block}

   $\depf'\ := \CombineDependencies(\{\depf_i\ |\ i\in A\})$
\end{algorithmus}

Mit Hilfe von \CombineImports\ wird eine Normalform erzeugt, die
alle importierten Module in sich vereint. Sie wird anschlie"send
durch Anwendung der Funktion \CombineWithImports\
mit der general form des importierenden Moduls kombiniert. Hier sind
keine Sichtbarkeitsanpassungen mehr notwendig:

\begin{algorithmus}{\CombineWithImports}%
{\GFORM\ \times\ \NFORM}{\NFORM}%
{  $\CombineWithImports((\mod, \originf, \depf), (\mod\subimp, \originf\subimp, \depf\subimp))$
   kombiniert die general form einer Modulinstanz mit einer Normalform,
   die aus allen von ihr importierten Modulen errechnet worden ist.  }%
{R"uckgabewert: (\mod$''$, \originf$'$, \depf$'$)}
   \mod$'$ geht aus \mod\ durch L"oschen aller Importkonstrukte hervor.
   
   $\mod''\ :=\ \mod' \cunion \mod\subimp$ \\
   Der Modulname von \mod$''$ (erste Komponente) wird auf den
   f"ur die Normalform von \mod\ vorgesehenen Namen gesetzt.
   Dieser kann beispielsweise
   aus dem Modulnamen von \mod\ durch Anh"angen der Extension
   \tsym{.nf} gewonnen werden.

   $\originf'\ := \originf\ \cup \originf\subimp$ \\
   Falls \originf$'$ keine Funktion: NAMECLASH

   Sei nun \modname\ der Modulname (1.\ Komponente) von \mod.
   \begin{tabbing}
   $\depf'\ :=\ $\=$ \{\ $\=$(\modname, \emptyset)\ \}\ \cup$
   \\            \> $\{\ (\modiname, \modinames \cup \{\modname\})\ |$
   \\            \>       \>$(\modiname, \modinames)\in \depf\subimp\ \}$
   \end{tabbing}
\end{algorithmus}
\vspace{3ex}

Wird in einem Importbefehl die Bindung eines Parametertupels aus dem
importierten Modul \mod$_{FORM}$  an Namen eines Moduls \mod$_{ACT}$
vorgenommen, so erfordert die Auswertung das Kombinieren der zugeh"origen
Normalformen. Dieser implizite Import des Moduls \mod$_{ACT}$ in das
Modul \mod$_{FORM}$ unterscheidet sich von gew"ohnlichen Importen, weil
hierdurch ein Modul ``nachtr"aglich'' in eine bereits bestehende
Modulhierarchie eingepflanzt wird.

\begin{algorithmus}{\CombineWithActModule}%
{\NFORM\ \times\ \MODULEINSTNAME\ \times\ \NFORM}{\NFORM}%
{  $\CombineWithActModule((\mod\subform, \originf\subform, \depf\subform),$
   \paradefmodiname, (\mod$\subact$,  \originf$\subact$,  \depf$\subact$))
   ``implantiert'' die Normalform (\mod$\subact$, \originf$\subact$, \depf$\subact$)
   in die Normalform (\mod$\subform$, \originf$\subform$, \depf$\subform$).
   Dabei wird eine Abh"angigkeit zwischen den Namensr"aumen
   des Moduls \mod$\subact$ und dem Namensraum der formalen
   Parameter \paradefmodiname\ aus \mod$\subform$ hergestellt.
   Es wird davon ausgegangen, da"s bereits alle Renamings in der
   Normalform des formalen Moduls
   und die Sichtbarkeitsanpassungen zwischen den Namen
   beider Normalformen durchgef"uhrt worden sind.
}%  
{R"uckgabewert: (\mod, \originf, \depf)}
   $\mod\ :=\ \mod\subact \cunion \mod\subform$ \\
   Der Modulname \modname\ von \mod$\subform$\ (erste Komponente)
   wird in \mod\ "ubernommen.

   $\originf\ :=\ \originf\subform \cup \originf\subact$ \\
   Falls \originf\ keine Funktion: NAMECLASH

   $\depf'\subact\ :=\ \{\ 
   \begin{array}[t]{@{}l}
      (\modiname, \modinames\ \cup \{\paradefmodiname\} \cup
                  \depf\subform(\paradefmodiname)\ | \\
      (\modiname, \modinames)\in\depf\subact\ \}
   \end{array}$

   $\depf\ :=\ \CombineDependencies(\depf\subform, \depf'\subact)$
\end{algorithmus}
\vspace{3ex}

\subsubsection{Modulmodifikationen in Importbefehlen}
\label{Modifikationen in Importbefehlen}

Werden in einem Importbefehl Namen umbenannt, Parameter gebunden oder die
Sichtbarkeit von Signaturnamen ver"andert, so f"uhrt das semantisch dazu,
da"s die Normalformen der zu importierenden Module modifiziert
werden m"ussen, bevor sie zu einer einzigen Normalform zusammengefa"st
werden k"onnen.
Diese Aufgabe "ubernehmen die Funktionen \Hide, \Rename\ und \Bind\ mit den
Hilfsfunktionen \InstanciateModInstName, \Instanciate, \SeperateParaBlock,
\GetParameterRenamings\ und \CheckSemanticConditions.

Ein wesentlicher Teil eines jeden Importbefehls sind die den Schl"usselworten
\ts{private:} und \ts{public:} folgenden Listen von Signaturnamen. Sie
geben Auskunft "uber die Sichtbarkeit der vom importierten Modul exportierten
Signaturnamen. Mit Hilfe der Funktion \Hide\ werden alle nicht exportierten
Namen verdeckt und die Sichtbarkeit der exportierten Signaturnamen den 
Vorgaben des Importbefehls angepa"st.

\begin{algorithmus}{\Hide}%
{\NFORM\ \times\ \VISIBILITYFUNC}%
{\NFORM}%
{  \Hide((\mod, \originf, \depf), \visibilityfunc) verdeckt alle Namen mit
   Sichtbarkeitsstufe \tsym{private}. Namen mit Sichtbarkeitsstufe \tsym{public}
   werden auf die in \visibilityfunc\ angegebene Sichtbarkeitsstufe gesetzt;
   ist keine Angabe vorhanden, erhalten sie die Sichtbarkeitsstufe
   \tsym{hidden}.
}%
{R"uckgabewert: $(\mod', \originf'', \depf)$}

\begin{tabbing}
$\originf'\ :=\ \{\ $\=$(\disambname, (\uname, \modinst, \symboltype, \visibility'))\quad |$
\\  \> $(\disambname, (\uname, \modinst, \symboltype, \visibility)) \in \originf$
\\  \> $\wedge\
        ($\=$(\visibility \in \{\tsym{hidden}, \tsym{parameter}\}\ \wedge$
             $\visibility = \visibility')\ \vee$
\\  \>    \>$(\visibility = \tsym{private} \wedge \visibility' = \tsym{hidden})\ \vee$
\\  \>    \>$(\visibility = \tsym{public}$\ 
\\  \>    \>\mbox{}\ $\wedge\
            ($\=$(\uname \notin \dom(\visibilityfunc)$
                 $\wedge\ \visibility' = \tsym{hidden})\ \vee$
\\  \>    \> \> $(\uname \in    \dom(\visibilityfunc)$
           $\wedge\ \visibility' = \visibilityfunc(\uname)))))\ \}$
\end{tabbing}

   $(\mod', \originf'') := \MakeConsistent(\mod, \originf')$
\end{algorithmus}

Der kopierende Import aus \ASFm\ basiert auf der Zuordnung der zu kopierenden
(Signa\-tur-) Namen zu neuen Namensr"aumen. Zu diesem Zweck werden neue
Namensraumbezeichnungen generiert, die sich aus den alten Bezeichnungen und
der Instanzbezeichnung des Importbefehls zusammensetzen.

\begin{algorithmus}{\InstanciateModInstName}%
{\MODULEINSTNAME\ \times\ \INSTNAME}{\MODULEINSTNAME}%
{  \InstanciateModInstName(\modiname, \iname) instanziiert die Namensraumbezeichnung
   \modiname\ mit der Instanzbezeichnung \iname. Wurde \modiname\ bereits
   mit \iname\ instanziiert, so liegt ein Spezifikationsfehler vor.  }%
{R"uckgabewert: {\em imodiname}}
   Falls \modiname\ $\in$ \MODULENAME \hspace{1cm} /* erste Instanziierung */
   \begin{block}
      {\em imodiname} := \modiname \tsym{[}\iname\tsym{]}
   \end{block}
   
   Sonst
   \begin{block}
      Sei \modname\tsym{[}\oldinames\tsym{]} =  \modiname \\
      Falls \iname\ in \oldinames\ enthalten ist
      \begin{block}
         SPEZIFIKATIONSFEHLER!
      \end{block}
      {\em imodiname} := \modname\tsym{[}\oldinames\tsym{,}\iname\tsym{]}
   \end{block}
\end{algorithmus}

Eine Normalform repr"asentiert nicht nur ein normalisiertes Modul; Origin-
und Dependenzfunktion erlauben die Rekonstruktion der gesamten zugrundeliegenden
Modulhierarchie. Werden Teile einer Normalform durch
explizites Renaming oder Parameterbindung modifiziert, k"onnen die
erforderlichen Instanziierungen auf die direkt betroffenen und die davon
abh"angigen Namensr"aume begrenzt werden.

\begin{algorithmus}{\Instanciate}%
{\NFORM\ \times\ \RENAMINGFUNC\ \times\ \pot(\BINDINGBLOCK)\ \times\ \INSTNAME}%
{\NFORM}%
{  \Instanciate((\mod, \originf, \depf), \renaming, \bindingblocks, \iname)
   instanziiert Namensraumbezeichnungen in der Normalform (\mod, \originf, \depf)
   mit der Instanzbezeichnung \iname. Instanziiert werden die Bezeichnungen
   aller vom expliziten Renaming \renaming\ und von den Parameterbindungen
   \bindingblocks\ direkt betroffenen
   Namensr"aume, sowie alle bez"uglich \depf\ von ihnen abh"angigen Namensr"aume.}%
{R"uckgabewert: (\mod$'$, \originf$'$, \depf$'$)}

   Sei $\{(\binding_i, \modname_i)\ |\ i\in A\} = \bindingblocks$

   $\toinstanciate\ :=\ 
       \{\ \modiname\ |\ 
       \begin{array}[t]{l}
          ((\name, *), (*, \modiname, *, *))\in\originf\ \wedge
       \\ \name \in \dom(\renaming) \cup \Union_{i\in A} \dom(\binding_i)\ \}
       \end{array}
$

   $\toinstanciate'\ :=\ \toinstanciate\ \cup\ 
                          \{\depf(\modinst)\ |\ \modinst\in\toinstanciate\}$

   Berechne $(\mod', \originf', \depf')$ durch
   Ersetzen jedes Auftretens einer Namensraumbezeichnung \modiname\ $\in$
   \toinstanciate$'$
   \begin{itemize}
      \item in \mod\ ("uberall dort, wo sie Teil eines verdeckten Namens ist),
      \item in \originf\ (Im Defi\-ni\-tions\-bereich "uberall dort, wo sie Teil eines
         verdeckten Namens ist und in der 2. Komponente der Origins des 
         Wertebereichs) und
      \item in \depf\ (wo immer sie auftritt)
   \end{itemize}
   durch \InstanciateModInstName(\modiname, \iname).
\end{algorithmus}

\begin{algorithmus}{\Rename}%
{\NFORM\ \times\ \RENAMINGFUNC}%
{\NFORM}%
{  \Rename((\mod, \originf, \depf), \renaming)
   f"uhrt explizites Renaming durch. \renaming\ enth"alt die Umbenennungen
   aller Renaminganweisungen des Importbefehls. }%
{R"uckgabewert: $(\mod', \originf', \depf)$}
   
   Sei $\ren(x) :=\ \left\{%
   \begin{array}{ll}
     y & $falls$\ (x,y)\in\renaming
\\   x & $sonst$
   \end{array}\right.$

   und \ren$'$ die Erweiterung von \ren\ auf Sortenvektoren:

   \ind $\ren'((\sortn_1,\ldots,\sortn_n)) :=
        (\ren(\sortn_1),\ldots,\ren(\sortn_n))$

   $\mod'$ wird aus $\mod$ durch syntaktisches Ersetzen aller Signaturnamen $\name$
   durch $\ren(\name)$ erzeugt. Man beachte da"s \ren\ nur Einflu"s auf sichtbare
   Namen ($\in$ \USERNAME) hat.

   SPEZIFIKATIONSFEHLER falls $\mod'$ keine korrekte Signatur enth"alt.
\\ /* Ursache kann hier ein fehlerhafter Renamingbefehl sein, der dazu f"uhrt,
   da"s urspr"unglich verschiedene Namen des gleichen Namensraumes nach Durchf"uhrung
   des Renamings zusammenfallen. Renamings dieser Art k"onnen Funktionen mit
   gleichen disambiguierten Namen aber unterschiedlichen Zielsorten erzeugen. */

   $\originf'\ := \{\ %
   \begin{array}[t]{l}
      (\ren(\name),\ren'(\sortv)),(\uname', \modiname, \symboltype, \visibility))\quad|
\\    ((\name,    \sortv),(\uname , \modiname, \symboltype, \visibility))\in\originf\ \wedge
\\    (\begin{array}[t]{@{}l}%
(\visibility   =    \tsym{hidden} \wedge\ \uname' = \uname)\ \vee
\\(\visibility \neq \tsym{hidden} \wedge\ \uname' = \ren(\uname)))\ \}
       \end{array}
   \end{array}$

\end{algorithmus}

Alle folgenden Funktionen dieses Abschnitts behandeln die Auswertung einer
Parametertupelbindung. Der Trivialfunktion \SeperateParaBlock\ und der (etwas
technischen) Hilfsfunktion \GetParameterRenamings\ folgen die Hauptfunktionen
\CheckSemanticConditions\ und \Bind. Die Komplexit"at der Funktionen folgt
aus der Tatsache, da"s es sich bei jeder Parametertupelbindung um einen
impliziten Import (also einen Import im Import) handelt und neben den schon
betrachteten Operationen (z.~B.\ Verdecken von Namen, Instanziieren von
Namensr"aumen) im Zuge des Testens semantischer Bedingungen und des
Implantierens eines aktuellen Moduls in die bereits bestehende Modulhierarchie
eines formalen Moduls eine Vielzahl neuer Rechenschritte erforderlich sind.

\begin{algorithmus}{\SeperateParaBlock}%
{\NFORM \times\ \pot(\SPECNAME)}%
{\NFORM \times\ (\SIG \times\ \pot(\CLAUSE)) \times\ \MODULEINSTNAME)}%
{  \SeperateParaBlock((\mod, \originf, \depf), \parameters) extrahiert aus der
   internen Moduldarstellung \mod\ die Parametersignatur und -bedingungen der in \parameters\
   enthaltenen Parameter eines Tupels. Die Parameter werden aus dem Defi\-ni\-tions\-bereich
   der Originfunktion entfernt und \paradefmodiname\ der Namensraum zugewiesen,
   dem die Parameter angeh"oren.
   \SeperateParaBlock\ ist Hilfsfunktion von \Bind.  }%
{R"uckgabewert: $((\mod', \originf', \depf), (\sig_p, \conditions),
                  \paradefmodiname)$}

   Seien \begin{tabular}[t]{l@{\hspace{1ex}}ll}
      $\sig_p$      & = &  die zu extrahierende Parametersignatur, \\
      \conditions   & = &  die zu $\sig_p$ geh"orenden Bedingungsklauseln und\\
      $\mod'$       & = &  das Modul, das nach Entfernen von ($\sig_p$, \conditions)
                           aus \mod\ entsteht. 
   \end{tabular}

   SPEZIFIKATIONSFEHLER, wenn keine Parametersignatur
   in \mod\ enthalten ist, die genau alle Namen aus \parameters\ enth"alt.

   Sei {\em parameter} $\in$ {\parameters}

   (*, \paradefmodiname, *, *) := \originf({\em parameter})

   /* Welcher Parameter genommen wird, hat keinen Einflu"s auf \paradefmodiname\ */

   $\originf'\ :=\ \{\ 
   \begin{array}[t]{l}
      ((\name, \sortv), \origin)\quad |\\
      ((\name, \sortv), \origin) \in \originf\ \wedge\ \name \notin \parameters\ \}
   \end{array}$
\end{algorithmus}

\yestop
\yestop

\begin{algorithmus}{\GetParameterRenamings}%
{\SIG\ \times\ \RENAMINGFUNC\ \times\ \ORIGINFUNC^2}%
{\RENAMINGFUNC}%
{  $\GetParameterRenamings((\sorts_p, \cons_p, \ncons_p), \binding,
    \originf\subact,$ $\originf\subactav)$ 
   berechnet die jenigen Namen, durch welche die nach der Vorschrift
   \binding\ an Namen eines
   aktuellen Moduls zu bindenden formalen Parameter syntaktisch ersetzt werden
   m"ussen. Die errechneten Namen sind im allgemeinen nicht mit denen aus 
   \range(\binding) identisch, weil alle Namen aus dem aktuellen Modul
   beim impliziten Import verdeckt werden, sofern sie nicht bereits im formalen
   Modul sichtbar sind. Als Argumente werden die Signatur
   des zu bindenden Parametertupels $(\sorts_p, \cons_p, \ncons_p)$,
   die Bindungsvorschrift \binding, die Originfunktion des aktuellen Moduls
   $\originf\subact$ und eine weitere 
   Originfunktion $\originf\subactav$, die aus $\originf\subact$ durch
   Setzen aller Namen auf die beim impliziten Import angestrebte Sichtbarkeitsstufe
   hervorgeht, "ubergeben.  }%
{R"uckgabewert: \parrenaming}

   Falls $\{(\binding(sortpar), \emptyset)\ |\ \sortpar\in\sorts_p\}\ 
   \not\subseteq\ \dom(\originf\subact)$
   \begin{block}
      SPEZIFIKATIONSFEHLER \ 
/* \parbox[t]{8.5cm}{Sortenparameterbindung fehlerhaft. Aktuelle Sorten
           existieren nicht im aktuellen Modul */}
   \end{block}

   $\sortparrenaming\ :=
\\ \ind \{\ (\sortpar, \name)\ |\ 
\begin{array}[t]{@{}l}
       \sortpar\in\sorts_p\ \wedge
\\     (\name, \emptyset)\in \dom(\originf\subactav)\ \wedge
\\     (\name, \emptyset)/\originf\subactav \approx
       (\binding(\sortpar), \emptyset)/\originf\subact\footnote%
{Zur Definition von $\approx$ siehe Seite \pageref{approx}}\ 
 \}
\end{array}$

   \hspace{2ex}

   Sei $\sorts\subnp$ die Menge aller Sortennamen, die in den Deklarationen\\
   $\cons_p \cup \ncons_p$ auftreten, aber nicht in $\sorts_p$ enthalten sind.

   Falls $\{(\sort, \emptyset)\ |\ \sort\in\sorts\subnp\}\
         \not\subseteq \dom(\originf\subactav)$
   \begin{block}
      SPEZIFIKATIONSFEHLER \ 
/* \parbox[t]{8.5cm}{Da Funktionsparameter nur an aktuelle Funktionsnamen gleicher
   Deklaration gebunden werden k"onnen, m"ussen die Nicht-Pa\-ra\-meter-Sorten\-namen
   der Funktionsparameterdeklarationen nicht nur im formalen, sondern auch im
   aktuellen Modul auftreten. */}
   \end{block}

   $\renaming\ :=
\\ \ind \{\ (\sortpar, \binding(\sortpar))\ |\ \sortpar\in \sorts_p\ \}\ \cup\
\\ \ind \{\ (\sort, \name)\ |\ 
\begin{array}[t]{@{}l}
   \sort \in \sorts\subnp\ \wedge
\\ (\name, \emptyset)\in \dom(\originf\subact)\ \wedge
\\ (\name, \emptyset)/\originf\subact \approx (\sort, \emptyset)/\originf\subactav\ \}
\end{array}$
   \hspace{2ex}

   Berechne $\cons'_p$ und $\ncons'_p$ aus $\cons_p$ und $\ncons_p$ durch 
   Umbenennen aller Sorten nach Ma"sgabe von $\renaming$.

   Sei $\disfuncs_p\ =\ \{\ (\funcpar, \sortv)\ |\ 
   ((\funcpar, \sortv), \sort) \in \cons'_p \cup \ncons'_p\ \}$

   Falls $\{(\binding(\funcpar), \sortv)\ |\ (\funcpar, \sortv)\in\disfuncs_p\}\ 
   \not\subseteq\ \dom(\originf\subact)$
   \begin{block}
      SPEZIFIKATIONSFEHLER \ 
/* \parbox[t]{8.5cm}{Bindung der Funktionsparameter fehlerhaft,
                kein ``wohlsortierter'' aktueller Parameter vorhanden */}
   \end{block}

   $\funcparrenaming\ :=
\\ \ind \{\ 
\begin{array}[t]{@{}l}
   (\funcpar, \name)\ |\
\\ (\funcpar, \sortv)\in\disfuncs_p\ \wedge
\\ (\name, \sortv') \in  \dom(\originf\subactav)\ \wedge
\\ (\name, \sortv')/\originf\subactav \approx 
       (\binding(\funcpar), \sortv)/\originf\subact\ \}
\end{array}$

$\parrenaming\ :=\ \sortparrenaming \cup \funcparrenaming$
\end{algorithmus}

\pagebreak

\begin{algorithmus}{\CheckSemanticConditions}%
{\pot(\CLAUSE)\ \times\ \MODULE^2\ \times\ \NFORM\ \times\ \PROVEDB}{-}%
{  $\CheckSemanticConditions(\conditions, \mod\subform, \mod\subactav, 
                             \nform\subact, \provedb)$
    pr"uft, ob die semantischen Bedingungen, die an die Bindung von Parametern aus
    $\mod\subform$ an Namen des in $\nform\subact$ enthaltenen aktuellen Moduls
    gekn"upft wurden, erf"ullt sind.
    \conditions\ ist eine Menge von Gentzen-Klauseln, die aus den semantischen Bedingungen
    nach Ersetzen der formalen durch die aktuellen Parameter hervorgegangen ist.
    $\mod\subactav$ ist eine Variante des aktuellen Moduls, in der die Sichtbarkeit
    der Signaturnamen an die Sichtbarkeit innerhalb des formalen Moduls  angepa"st
    worden ist. }%
{R"uckgabewert: -}

   Setze $(*, *, *, *, *, \varsortfunc\subconstform, \varsortfunc\subnonconstform, *, *)
   := \mod\subform$

   Setze $(*, *, *, *, *, \varsortfunc\subconstactav, \varsortfunc\subnonconstactav, *, \goals\subactav)
   := \mod\subactav$

$\begin{array}{@{}l@{}l@{}l}
   \varsortfunc'\subform  &:= \varsortfunc\subconstform  &\cup\ \varsortfunc\subnonconstform

\\ \varsortfunc'\subactav &:= \varsortfunc\subconstactav &\cup\ \varsortfunc\subnonconstactav
\end{array}$
      
   F"ur alle Gentzenklauseln $\condition \in \conditions$
   \begin{block}
      Falls es nicht eine Gentzenklausel $\goal \in \goals\subactav$ und eine
      Variablensubstitution \subst\ gibt mit:
      \begin{itemize}
      \item $\subst(\goal) = \condition$\ (Die Marken werden hier nicht
         ber"ucksichtigt),
      \item \subst\ ist ``sortenrein'', d~h.\ f"ur alle $(x, y) \in \subst$ gilt \\
         $\varsortfunc'\subactav(x) = \varsortfunc'\subform(y)$,
      \item \subst\ substituiert Konstruktor-Variablen mit Konstruktor-Variablen und
         Non-Konstruktor-Variablen mit Non-Konstruktor-Variablen,
         d.~h.\ f"ur alle $(x, y)\in\subst$ gilt
         $x \in \dom(\varsortfunc\subconstactav) \Longleftrightarrow
          y \in \dom(\varsortfunc\subconstform)$ und
      \item das zu \goal\ korrespondierende Beweisziel in $\nform\subact$ 
         (kann mit Hilfe des Markennamens bestimmt werden) gilt
         dort als bewiesen, 
         dh. es gibt einen entsprechenden Beweis in \provedb.
      \end{itemize}

         SEMANTIC ERROR: Bevor die Spezifikation akzeptiert werden kann
         mu"s (falls noch nicht vorhanden) ein entsprechendes Beweisziel
         in das aktuelle Modul eingef"ugt und
         dessen G"ultigkeit bewiesen werden.

   \end{block}
\end{algorithmus}

\pagebreak %------------------------------------------------------------------------

\begin{algorithmus}{\Bind}%
{\NFORM\ \times\ \RENAMINGFUNC\ \times\ \NFORM\ \times\ \PROVEDB}%
{\NFORM}%
{  $\Bind((\nform\subform, \binding,
          (\mod\subact,  \originf\subact,  \depf\subact)$,  \provedb)
   f"uhrt die Bindung eines Parametertupels durch. $\nform\subform$ enth"alt das
   normalisierte, parametrisierte Modul, dessen Parameter nach der Vorschrift
   \binding\ an Namen des normalisierten Moduls $\mod\subact$ gebunden werden sollen. }%
{R"uckgabewert: $\nform\subresult$}

   /* Zun"achst werden alle Namen aus
      $(\mod\subact,  \originf\subact,  \depf\subact)$ verdeckt. */

   $\nform'\subact\ :=\ 
    \Hide((\mod\subact, \originf\subact, \depf\subact), \emptyset)$

\begin{block}
   /* \AdaptVisibility\ "andert die Sichtbarkeit von Namen aus verschiedenen Modulen
      nach dem Prinzip der ``maximalen'' Sichtbarkeit. Angewandt auf $\nform\subform$
      und $\nform'\subact$
      bleibt $\nform\subform$ unver"andert, weil es dort keinen
      Signaturnamen gibt, der in $\nform'\subact$ sichtbar ist. */
\end{block}

   $\nforms\ :=\ 
    \AdaptVisibility(\{\nform'\subact, \nform\subform\}, \{\lvs\})$

   $\{(\mod\subactav,  \originf\subactav,  \depf\subactav)\}\ :=$
\\ \ind
   $\AdaptVisibility(\nforms, \{\tsym{function}\})
   \quad \backslash\quad \{\nform\subform\}$

   $((\mod\subform, \originf\subform, \depf\subform),
     (sig_p, \conditions), \paradefmodiname)\ :=$
\\ \ind
   $ \SeperateParaBlock(\nform\subform, \dom(\binding))$

   $\parrenaming :=
   \GetParameterRenamings(\sig_p, \binding, \originf\subact, \originf\subactav)$

   Berechne $\mod'\subform, \originf'\subform$ und $\conditions'$
   aus $\mod\subform, \originf\subform$ und $\conditions$ durch Ersetzen
   der formalen Parameter nach Ma"sgabe von \parrenaming.

   SPEZIFIKATIONSFEHLER falls $\mod'\subform$ keine korrekte Signatur enth"alt.
   \begin{block}
   /* Eine fehlerhafter Parameterbindung kann dazu f"uhren,
      da"s Funktionen mit gleichen disambiguierten Namen und unterschiedlichen
      Zielsorten erzeugt werden. */
   \end{block}

   $\CheckSemanticConditions(%
   \begin{array}[t]{@{}l}
      \conditions', mod'\subform, \mod\subactav, \\
      (\mod\subact, \originf\subact, \depf\subact), \provedb)
   \end{array}$

   $\nform\subresult\ :=\ $
   $\CombineWithActModule(%
    \begin{array}[t]{@{}l}
       (\mod'\subform, \originf'\subform, \depf\subform), 
       \paradefmodiname, \\
       (\mod\subactav,  \originf\subactav,  \depf\subactav))
    \end{array}$

\end{algorithmus}

\pagebreak %---------------------------------------------------------------------

\vspace{3ex}
\subsubsection{Die Normalisierungsfunktionen \NForm\ und \NormalForm}
Ziel dieses Abschnitts ist die Vorstellung einer Funktion \NormalForm,
die eine gegebene hierarchische \ASFm-Spezifikation in eine flache, nur
aus einem Topmodul bestehende \ASFm-Spezifikation transformiert. \NormalForm\
besteht im wesentlichen aus einem Aufruf der rekursiven Funktion \NForm.
\NForm\ ist die f"ur das Verst"andnis des Algorithmus grundlegende Funktion.
Weiterhin werden die Trivialfunktionen \ModuleText, \MakeGF\
und \ExternModRep\ ben"otigt.

\begin{algorithmus}{\ModuleText}%
{\MODULENAME\ \times\ \ASFSPEC}{\ASFMODULE}%
{  \ModuleText(\modname, \spec) sucht ein Modul {\em asf-module} namens
   \modname\ in \spec. Falls kein solches Modul existiert:
   SPEZIFIKATIONSFEHLER! \ModuleText\ ist Hilfsfunktion von \NForm.  }%
{R"uckgabewert: {\em asf-module}}
   Weitere Formalisierung entf"allt.
\end{algorithmus}

\begin{algorithmus}{\MakeGF}%
{\ASFMODULE}{\GFORM}%
{  \MakeGF({\em asf-module}) berechnet aus einem isolierten nicht
   notwendig importfreien \ASFm-Modul einer Spezifikation eine general form
   (\mod, \originf, \depf). \MakeGF\ ist Hilfsfunktion von \NForm. }%
{R"uckgabewert: (\mod, \originf, \depf)}
   Der Wert von \mod\ wird direkt aus dem ASF-Modul ermittelt, es handelt
   sich hier lediglich um eine andere Repr"asentationsform.
   
   Jedem (disambiguierten) Sorten- und Funktionsnamen aus der Signatur, jedem
   (disambiguierten) Parameternamen aus einer der Parametersignaturen und jedem,
   innerhalb des Moduls auftretenen (disambiguierten) Variablen- und
   Markennamen wird vermittels \originf\ ein Origin zugeordnet.%
\footnote{Siehe dazu Seite \pageref{origin}}
   \originf\
   ist zun"achst partiell in dem Sinn, da"s importierte (Teil-) Signaturen
   noch nicht in \dom(\originf) enthalten sind.

   $\depf\ := \emptyset$
\end{algorithmus}

\pagebreak %-------------------------------------------------------------------------

\begin{algorithmus}{\NForm}%
{\MODULENAME\ \times\ \ASFSPEC\ \times\ \PROVEDB}{\NFORM}%
{  \NForm(\modname, \spec, \provedb) berechnet rekursiv die Normalform $\nform\subresult$
   der zum Modul namens \modname\ zugeh"origen general form.  }%
{R"uckgabewert: $\nform\subresult$}

   $\importinggf\ :=\ \MakeGF(\ModuleText(\modname, \spec))$

   Setze $((*, \imports, \ldots), *, *) := \importinggf$

   Sei $\{(\modname_i, \iname_i, \visibilityfunc_i, \renaming_i, \bindingblocks_i)
       \quad |\quad i\in A\}\ =\ \imports$

   F"ur alle $i \in A$
   \begin{block}
      $\begin{array}{@{}ll}
      \nform_i     & :=\ \NForm(\modname_i, \spec, \provedb)
\\    \nform'_i    & :=\ \Hide(\nform_i, \visibilityfunc_i)
      \end{array}$

      Falls $\iname_i = \emptyset\ \wedge\ (\renaming_i\neq\emptyset\ \vee\
                                            \bindingblocks_i\neq\emptyset)$
      \begin{block}
         SPEZIFIKATIONSFEHLER
      \end{block}

      Falls $\iname_i \neq \emptyset$%
      \begin{block}$%
         \begin{array}{@{}ll}
         \nform''_i   & :=\ \Instanciate(\nform'_i, \renaming_i, \bindingblocks_i, \iname_i)
\\       \nform'''_i  & :=\ \Rename(\nform''_i, \renaming_i)
         \end{array}$

         F"ur alle $(\binding, \modname\subact) \in \bindingblocks_i$ wiederhole
         \begin{block}$
            \begin{array}{@{}ll}
            \nform\subact\ & :=\ \NForm(\modname\subact, \spec, \provedb)
\\          \nform'''_i\   & :=\ \Bind(\nform'''_i, \binding, \nform\subact, \provedb)
            \end{array}$
         \end{block}
      \end{block}
   \end{block}
   $\nform\subresult := \CombineWithImports(\importinggf,
                        \CombineImports(\{\nform'''_i\ |\ i\in A\}))$

   %%%{\em imp} := \NForm({\em impmodname}, \spec, \provedb) \\
   %%%\justinstanciated := \math\emptyset
      
   %%%{\em allimp} := \CombineImports({\em allimp}, {\em imp})
   
\end{algorithmus}

\begin{algorithmus}{\ExternModRep}%
{\MODULE}{\ASFMODULE}%
{  \ExternModRep(\module) berechnet die \ASFm-Darstellung {\em asf-module} des
   Moduls \module. Diese Funktion kann mit einer Option ausgestattet
   werden, die es erlaubt "uberladene Funktionsnamen durch eindeutige
   Repr"asentationen ihrer disambiguierten Namen zu ersetzen.
   \ExternModRep\ ist Hilfsfunktion von \NormalForm.  }%
{R"uckgabewert: {\em asf-module}}
   Weitere Formalisierung entf"allt!
\end{algorithmus}

\pagebreak %----------------------------------------------------------

\begin{algorithmus}{\NormalForm}%
{\ASFSPEC\ \times\ \PROVEDB}{\ASFSPEC}%
{  \NormalForm(\spec, \provedb) berechnet aus einer modularen \ASFm-Spezifikation
   eine Spezifikation, bestehend aus einem einzigen (importfreien)
   Modul {\em asf-module}. Die Wissensbasis \provedb\ beinhaltet Informationen
   "uber gelungene Beweise und wird f"ur die "Uberpr"ufung von semantischen
   Bedingungen gebraucht.}%
{R"uckgabewert: ({\em asf-module}, \math\emptyset)}
   Sei \modname\ der Name des Topmoduls aus \spec.\\
   (\mod, \originf, \depf) := \NForm(\modname, \spec, \provedb)
   
   {\em asf-module} := \ExternModRep(\mod)
\end{algorithmus}

\vfill\pagebreak

\subsection{Ein Beispiel f"ur ein normalisiertes Modul}

Um die Arbeitsweise des Normalformalgorithmus zu veranschaulichen
geben wir schlie"slich noch das importfreie, durch Normalisierung erzeugte Modul {\tt OrdNatSequences.nf} an.

\begin{verbatim}
module OrdNatSequences.nf
{      
   add signature
   {
      public:
         sorts
            BOOL, NAT, NSEQ
         constructors
            true, false   :             -> BOOL
            0             :             -> NAT
            s             : NAT         -> NAT
            Nnil          :             -> NSEQ
            cons          : NAT # NSEQ  -> NSEQ
         non-constructors
            greater       : NAT  # NAT  -> BOOL
            greater       : NSEQ # NSEQ -> BOOL

      private:
         non-constructors
            Bo-and, Bo-or : BOOL # BOOL -> BOOL
            Bo-not        : BOOL        -> BOOL
            _ Nat-+ _     : NAT # NAT   -> NAT
            Nat-eq        : NAT # NAT   -> BOOL  
            ONat-geq      : NAT # NAT   -> BOOL
  }
   
   variables
   {  constructors
         Nat-x, Nat-y, Nat-u,
         ONat-x, ONat-y, ONat-u, ONat-v,
         OSeq-i1, OSeq-i2, OSeq-i3              : -> NAT
         OSeq-seq1, OSeq-seq2, OSeq-s1, OSeq-s2 : -> NSEQ
      non-constructors
         Bo-x, Bo-y                             : -> BOOL  }
      
   equations
   {
      macro-equation Bo-and(Bo-x,Bo-y)
      {
         case
         {  ( Bo-x @ true ) : Bo-y
            ( Bo-x @ false ): false  }
      }
      
      macro-equation Bo-not(Bo-x)
      {
         case
         {  ( Bo-x @ true ) : false
            ( Bo-x @ false ): true  }
      }

      [Bo-e1] Bo-or(Bo-x, Bo-y) =
              Bo-not(Bo-and(Bo-not(Bo-x), Bo-not(Bo-y)))

      macro-equation (Nat-x Nat-+ Nat-y)
      {
         case
         {  ( Nat-y @ 0 )        : Nat-x
            ( Nat-y @ s(Nat-u) ) : s(Nat-x Nat-+ Nat-u)  }

      macro-equation Nat-eq(Nat-x, Nat-y)
      {  if ( Nat-x = Nat-y ) true
         else                 false  }
      }

      macro-equation greater(ONat-x, ONat-y)
      {
         case
         {  ( ONat-x @ 0 )                     : false
            ( ONat-x @ s(ONat-u), ONat-y @ 0 ) : true
            ( ONat-x @ s(ONat-u), ONat-y @ s(ONat-v) ):
              greater(ONat-u,ONat-v)  }
      }
      
      [ONat-e1] ONat-geq(ONat-x,ONat-y) = 
                Bo-or(greater(ONat-x,ONat-y), eq(ONat-x,ONat-y))


      macro-equation greater(OSeq-seq1, OSeq-seq2)
      {              /* lex-order of sequences */
         case
         {
            ( OSeq-seq1 @ Nnil )                  : false
            ( OSeq-seq1 @ cons(OSeq-i1, OSeq-s1),
              OSeq-seq2 @ Nnil                   ): true
            ( OSeq-seq1 @ cons(OSeq-i1, OSeq-s1),
              OSeq-seq2 @ cons(OSeq-i2, OSeq-s2) ):
              if ( greater(OSeq-i1, OSeq-i2) )
                   true
              else if ( OSeq-i1 = OSeq-i2 ) 
                        greater(OSeq-s1, OSeq-s2)
                   else false
         }
      }

   }
   
   goals
   {
      [ONat-irref] greater(ONat-x, ONat-x)
                   -->
      [ONat-trans] greater(ONat-x, ONat-u), greater(ONat-u, ONat-y)
                   --> greater(ONat-x, ONat-y)
      [ONat-total] 
                   --> greater(ONat-x, ONat-y), greater(ONat-y, ONat-x),
                       ONat-x = ONat-y
   }
} /* OrdNatSequences.nf */
\end{verbatim}

\vfill\pagebreak

\section{Abschlie"sende Zusammenfassung}

Mit \ASFm\ ist es gelungen, eine algebraische Spezifikationssprache zu
entwickeln, die neue Konzepte wie beispielsweise das differenzierte
Verdecken von Signaturnamen, semantische Bedingungen
an Parameter und die Angabe von Beweiszielen in sich vereint, ohne
dabei auf wesentliche Elemente der bereits existierenden Sprache ASF
verzichten zu m"ussen. Hierbei konnte die Syntax von ASF sogar noch
vereinfacht werden.

\ASFm\ ist jedoch mehr als eine nur um zus"atzliche Konstruke erweiterte
Version von ASF. Grunds"atzliche Untersuchungen (wie in Kapitel
\ref{hierachische Konzepte} dargestellt) deckten Fehler in der
Semantik von ASF auf und f"uhrten zu den Begriffsbildungen
``benutzender'' und ``kopierender Import''. W"ahrend der benutzende
Import aus ASF "ubernommen wurde, verhindern in \ASFm\ von den
kopierenden Importbefehlen zur Verf"ugung gestellte Instanzbezeichnungen
Namensverwechselungen zwischen dem manipulierten Modul und seinem Original.

Wesentlicher Bestandteil von \ASFm\ ist das Namensraumkonzept, welches jedem
Signaturnamen bei seiner Definition den Modulnamen zuordnet.
W"ahrend beim benutzenden Import der Namensraum unver"andert bleibt,
f"uhrt der kopierende Import eines Namens zur Instanziierung des
zugeordneten Namensraumes. Bei der Kombination mehrerer Module zu einem
Normalformmodul werden nur solche Namen identifiziert, die dem gleichen
Namensraum angeh"oren.

Das Namensraumkonzept spielt auch in der Semantik verdeckter Namen eine
wichtige Rolle. Jedem zu verdeckenden Namen wird im Zuge der
Normalisierung die (abgek"urzte) Namensraumbezeichnung vorangestellt.
Dies erh"oht die Verst"andlichkeit des erzeugten Normalformmoduls und
macht den modularen Aufbau der Spezifikation sichtbar.

Der Preis f"ur die Verbesserungen ist jedoch eine gewisse
Verkomplizierung der Normalisierungsprozedur, was beim Vergleich des
im Kapitel \ref{normalisierung} vorgestellten Algorithmus mit dem
aus \Bergstra\ (Seite 23-28) deutlich wird.

Schlie"slich erlauben die von uns entwickelten Strukturdiagramme eine
ebenso informative wie leicht verst"andliche Darstellung von
\ASFm-Spezifikationen. Diese Strukturdiagramme eignen sich dar"uberhinaus
auch dazu, ein korrektes intuitives Verst"andnis f"ur die
wesentlichen Konzepte der Normalisierungsprozedur --- wie
Originfunktion, Dependenzfunktion, Sichtbarkeitsanpassung, Renaming,
Parameterbindung, Namensrauminstanziierung, etc.\ ---
zu vermitteln.

\vfill\pagebreak

\noindent
\begin{center}
{\Large\bf Literatur} \vspace{5mm}

\label{bibliography}

\begin{tabular}{ll}
\Bergstra
  & J.\ A.\ Bergstra, J.\ Heering, P.\ Klint (1989). \\
  & {\it Algebraic Specification}. \\
  & ACM Press. \\ \\

\Eschbach
  & Robert Eschbach (1994). \\
  & {\it ART --- Modularisierung von} \\ 
  & {\it Induktionsbeweisen "uber Gleichungsspezifikationen.}\\
  & SEKI-WORKING-PAPER SWP--94--03 (SFB),\\
  & Fachbereich Informatik, Universit\"at Kaiserslautern, \\
  & D--67663 Kaiserslautern.\\ \\

[Hendriks91]
  & P.~R.~H.~Hendriks (1991). \\
  & {\it Implementation of Modular Algebraic Specifications}. \\
  & PhD.\ Thesis, \\
  & CWI (Centrum voor Wiskunde en Informatica), Amsterdam.\\ \\

\WirthA
  & Claus-Peter Wirth, Bernhard Gramlich (1993). \\
  & {\it A Constructor-Based Approach for} \\
  & {\it Positive/Negative-Conditional Equational Specifications}.\\
  & 3$^{\mbox{\tiny rd}}$\ CTRS 1992, LNCS 656, Seiten 198-212, Springer-Verlag.\\
  & "Uberarbeitete und erweiterte Version in: \\
  & J.\ Symbolic Computation (1994) 17, Seiten 51-90, \\
  & Academic Press.\\ \\

\WirthB
  & Claus-Peter Wirth, Bernhard Gramlich (1994). \\
  & {\it On Notions of Inductive Validity }\\
  & {\it for First-Order Equational Clauses.} \\
  & 12$^{\mbox{\tiny th}}$\ CADE 1994, LNAI 814, Seiten 162-176, Springer-Verlag.
\\ \\

\Lunde
  & Claus-Peter Wirth, R"udiger Lunde (1994). \\
  & {\it Writing Positive/Negative-Conditional Equations} \\
  & {\it Conveniently.} \\
  & SEKI-WORKING-PAPER SWP--94--04 (SFB),\\
  & Fachbereich Informatik, Universit\"at Kaiserslautern, \\
  & D--67663 Kaiserslautern.\\

\end{tabular}
\end{center}
\end{document}